\PassOptionsToPackage{dvipsnames}{xcolor}
\documentclass[acmsmall, screen, nonacm]{acmart}
\setcopyright{none}

\usepackage[frozencache, cachedir=.]{minted}
\usepackage{graphicx} 
\usepackage{amsmath}  
\usepackage{amsthm}   
\usepackage{bbold}
\usepackage{bbm}
\usepackage{bussproofs}
\usepackage{lipsum}
\usepackage{mathtools}
\usepackage{cancel}
\usepackage{enumitem}
\usepackage{pifont}
\newcommand{\xmark}{\ding{55}}%
\newcommand{\cmark}{\ding{51}}%

\usepackage{epigraph}  
\makeatletter
\renewenvironment{proof}[1][\proofname]{%
  \par\pushQED{\qed}%
  \normalfont
  \topsep6\p@\@plus6\p@\relax
  \trivlist
  \item[\hskip\labelsep\sffamily \ul{#1}.]\ignorespaces
}{%
  \popQED\endtrivlist\@endpefalse
}
\makeatother

\usepackage{adjustbox}

\usepackage{quiver}
\usepackage{array}
\usepackage{tikz}
\usepackage{tikz-cd}
\usetikzlibrary{shadows}

\usepackage[most]{tcolorbox}
\usepackage[T1]{fontenc}
\usepackage[utf8]{inputenc}

\newcommand{\tul}[1]{%
  \leavevmode
  \sbox0{#1}
  \dimen0=\wd0
  {#1}%
  \llap{\rule[-3.8pt]{\dimen0}{.8pt}}
}

\tcbset{
    bthmbox/.style={
        enhanced,
        unbreakable,
        sharp corners=all,
        fonttitle=\sffamily\normalsize,
        fontupper=\normalsize,
        top=0mm,
        bottom=0mm,
        right=0mm,
        borderline west={3.35mm}{0pt}{black},
        borderline west={3.0mm}{0.35mm}{bmidnight},
        colback=white,
        colframe=white,
        colbacktitle=white,
        coltitle=black,
        attach boxed title to top left,
        boxed title style={empty, size=minimal, bottom=1.5mm},
        overlay unbroken ={
            \draw[line width=.8pt] (title.south west)--(title.south east);
        },
        overlay first={
            \draw[line width=.8pt] (title.south west)--(title.south east); 
            \draw[line width=.8pt] ([xshift=3.5mm]frame.north west)--([xshift=3.5mm]frame.south west);
        },
        overlay middle={
            \draw[line width=.8pt] ([xshift=3.5mm]frame.north west)--([xshift=3.5mm]frame.south west);
        },
        overlay last={
        },
    },
}

\tcbset{
    rthmbox/.style={
        enhanced,
        unbreakable,
        sharp corners=all,
        fonttitle=\sffamily\normalsize,
        fontupper=\normalsize,
        top=0mm,
        bottom=0mm,
        right=0mm,
        borderline west={3.35mm}{0pt}{black},
        borderline west={3.0mm}{0.35mm}{cayenne},
        colback=white,
        colframe=white,
        colbacktitle=white,
        coltitle=black,
        attach boxed title to top left,
        boxed title style={empty, size=minimal, bottom=1.5mm},
        overlay unbroken ={
            \draw[line width=.8pt] (title.south west)--(title.south east);
        },
        overlay first={
            \draw[line width=.8pt] (title.south west)--(title.south east); 
            \draw[line width=.8pt] ([xshift=3.5mm]frame.north west)--([xshift=3.5mm]frame.south west);
        },
        overlay middle={
            \draw[line width=.8pt] ([xshift=3.5mm]frame.north west)--([xshift=3.5mm]frame.south west);
        },
        overlay last={
        },
    },
}

\newtcbtheorem{bthm}{\textbf{\tul{Theorem}}}{bthmbox}{thm}
\newtcbtheorem{blem}{\textbf{\tul{Lemma}}}{bthmbox}{lemm}
\newtcbtheorem{rlem}{\textbf{\tul{Lemma}}}{rthmbox}{lemm}
\newtcbtheorem{rthm}{Theorem}{rthmbox}{thm}
\newtcbtheorem{bex}{Example}{bthmbox}{ex}
\newtcbtheorem{rex}{Example}{rthmbox}{ex}
\newtcbtheorem{bdef}{Definition}{bthmbox}{ex}
\newtcbtheorem{rdef}{Definition}{rthmbox}{ex}

\usepackage{tabularray}
\usepackage{mdframed}
\usepackage{subcaption}
\usepackage{wrapfig}

\newcommand{\myhrulefill}{\leavevmode\leaders\hrule height .8pt\hfill\kern-.8pt}
\newcommand{\rulediv}[2]{\raisebox{.5ex}{\makebox[\linewidth]{\rule{\dimexpr50pt-#1\relax}{.8pt}\ \raisebox{-.412ex}{\makebox[2\dimexpr#1\relax]{\textsf{\small{#2}}}}\hspace{.7ex}\myhrulefill}}}

\usepackage{tcolorbox}
\tcbuselibrary{minted,skins}

\newtcblisting{hscode1}[1][]{
    listing engine=minted,
    colback=hscodebg,
    colframe=black!100,
    listing only,
    boxrule=0.8pt,
    minted style=friendly,
    minted language=haskell,
    minted options={linenos=false,numbersep=3.5mm,texcl=true,#1},
    left=5mm,enhanced,
    sharp corners,
    overlay={\begin{tcbclipinterior}\fill[cayenne] (frame.south west)
            rectangle ([xshift=3mm]frame.north west);\end{tcbclipinterior}}
}
\newtcblisting{hscode2}[1][]{
    listing engine=minted,
    colback=hscodebg,
    colframe=black!100,
    listing only,
    boxrule=0.8pt,
    minted style=friendly,
    minted language=haskell,
    minted options={linenos=false,numbersep=3.5mm,texcl=true,#1},
    left=5mm,enhanced,
    sharp corners, 
    overlay={\begin{tcbclipinterior}\fill[bmidnight] (frame.south west)
            rectangle ([xshift=3mm]frame.north west);\end{tcbclipinterior}}
}
\definecolor{hscodebg}{rgb}{0.98,0.98,0.98}
\definecolor{cayenne}{rgb}{0.58,0.067,0}
\definecolor{midnight}{rgb}{0.008, 0.08, 0.47}

\usepackage{minted}
\setminted{fontsize=\fontsize{8.25pt}{11pt}\selectfont}
\definecolor{offwhite}{rgb}{0.98,0.98,0.98}

\usepackage[skip=2pt]{caption} 

\usepackage{preamble}
\newcommand{\Cj}{$\cj${\scriptsize $(\nb{\Bool},\nb{\lto})$}}
\newcommand{\Cjn}{$\cj${\scriptsize $(\nb{\mathbb{N}},\nb{\lto})$}}

\begin{document}

\title{Compiling to linear neurons}

\author{Joey Velez-Ginorio}
\email{joeyv@seas.upenn.edu}
\orcid{0009-0004-6451-5107}
\affiliation{
  \institution{University of Pennsylvania}
  \country{USA}
}

\author{Nada Amin}
\email{namin@seas.harvard.edu}
\orcid{0000-0002-0830-7248}
\affiliation{
  \institution{Harvard University}
  \country{USA}
}

\author{Konrad Paul Kording}
\email{kording@upenn.edu}
\orcid{0000-0001-8408-4499}
\affiliation{
  \institution{University of Pennsylvania}
  \country{USA}
}

\author{Steve Zdancewic}
\email{stevez@seas.upenn.edu}
\orcid{0000-0002-3516-1512}
\affiliation{
  \institution{University of Pennsylvania}
  \country{USA}
}


\newif\ifcomments\commentstrue   
\newif\ifaftersubmission \aftersubmissionfalse 
\newif\ifplentyofspace \plentyofspacefalse 
\ifcomments
  \newcommand{\proposecut}[1]{\ifcomments{\color{gray}\sout{#1}}\fi}
  \newcommand{\sz}[1]{\textcolor{brown}{{[SZ:~#1]}}}
  \newcommand{\kk}[1]{\textcolor{blue}{{[KK:~#1]}}}
  \newcommand{\na}[1]{\textcolor{orange}{{[NA:~#1]}}}
  \newcommand{\jvg}[1]{\textcolor{ForestGreen}{{[JVG:~#1]}}}
  
\else
  \newcommand{\proposecut}[1]{}
  \newcommand{\todo}[1]{}
\fi

\def\Cj{$\cj${\scriptsize $(\nb{\Bool},\nb{\lto})$}}

\begin{abstract}
We don't program neural networks directly. Instead, we rely on an indirect style where learning algorithms, like gradient descent, determine a neural network’s function by learning from data. This indirect style is often a virtue; it empowers us to solve problems that were previously impossible. But it lacks discrete structure. We can’t compile most algorithms into a neural network—even if these algorithms could help the network learn. This limitation occurs because discrete algorithms are not obviously differentiable, making them incompatible with the gradient-based learning algorithms that determine a neural network’s function. To address this, we introduce \cj: a typed, higher-order and linear programming language intended to be a minimal vehicle for exploring a direct style of programming neural networks. We prove \cj\ programs compile to linear neurons, allowing discrete algorithms to be expressed in a differentiable form compatible with gradient-based learning. With our implementation of \cj, we conduct several experiments where we link these linear neurons against other neural networks to determine part of their function prior to learning. Linking with these neurons allows networks to learn faster, with greater data-efficiency, and in a way that's easier to debug. A key lesson is that linear programming languages provide a path towards directly programming neural networks, enabling a rich interplay between learning and the discrete structures of ordinary programming.
\end{abstract}

\maketitle

\section{Introduction}




\begin{wrapfigure}{h}{0.33\textwidth}
    \centering
    \vspace{-1em}
    \includegraphics[width=.33\textwidth]{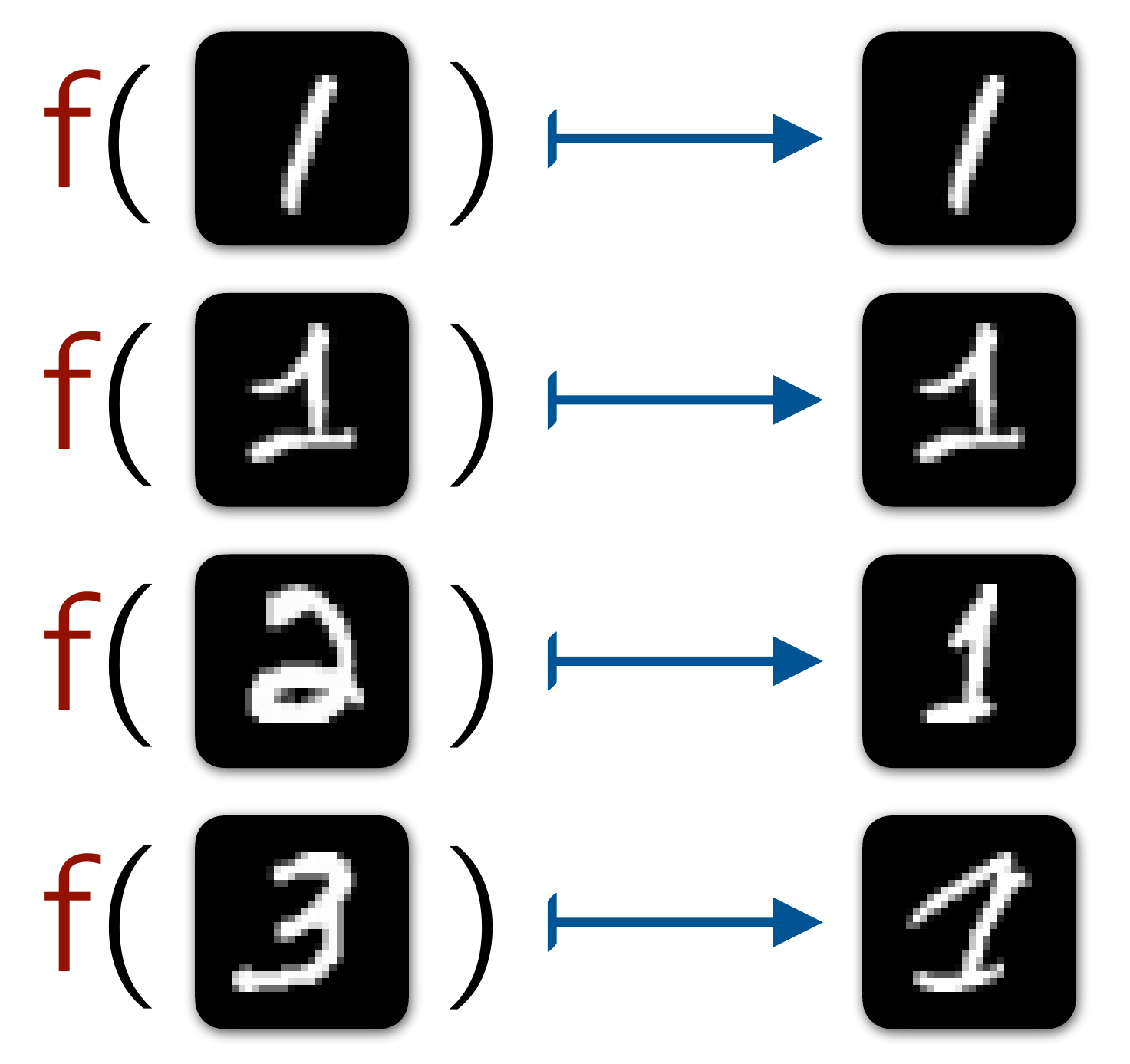}
    \vspace{-.5em}
  \caption{Transforming MNIST digits}
  \label{fig:task}
  \vspace{-1.2em}
\end{wrapfigure}

This paper reduces the problem of directly programming neural networks to a question of compiling: when do neural networks implement discrete algorithms? As a case study, we focus on a restricted but expressive class of discrete algorithms, those defined over boolean data. To illustrate the challenge, we present a simple example that exposes the gap between direct and indirect programming---a gap that ultimately stems from difficulties with compiling.


Consider how we could train a neural network to solve the task in Fig. \ref{fig:task}. We need a function which conditionally transforms an input image $x$: 
\setlist[itemize]{topsep=4pt, itemsep=1pt, parsep=2pt, leftmargin=1.5em, rightmargin=1.5em}
\begin{itemize}[label=$\triangleright$]
\item If $x$ is a 1, return $x$
\item If $x$ isn't a 1, transform $x$ to a 1
\end{itemize}
A simple approach is to define a 3-layer neural network, using data\footnote{These data are from the Modified National Institute of Standards and Technology (MNIST) database \cite{lecun2010mnist}.} like those shown in Fig. \ref{fig:task} to train the network. An advantage here is ease of specification. Fig. \ref{fig:indirect-program} shows how we could program this network; it's just a composition of functions between vectors. Our efforts lie instead on heuristic adjustments to learning, so that the network adequately learns our task. This indirect style of programming is typical in machine learning, where we defer to learning algorithms to determine a neural network's function \cite{lecun2015deep, bengio2017deep}.  And it continues to perform remarkable feats. Neural networks can learn to identify objects in an image, write code, or even respond sensibly\footnote{Sensible but not necessarily correct.} to most questions \cite{krizhevsky2012imagenet, achiam2023gpt}. 
But this deferral to learning comes at a cost. We lack control over the algorithms neural networks will learn, how long it can take to learn them, or how to debug them when they behave inadequately \cite{sculley2015hidden, bereska_mechanistic_2024}.

\begin{figure}[t]
  \centering
  \begin{minipage}[b]{0.36\textwidth}
    \centering
    \begin{hscode1}
f :: Real(784) -> Real(784)
f = w2 . relu . w1
    where w1 x = _
          w2 x = _
    \end{hscode1}
    \vspace{-.5em}
    \subcaption{Indirectly programmed neurons}
    \label{fig:indirect-program}
  \end{minipage}
  \hfill
  \begin{minipage}[b]{0.59\textwidth}
      \centering
    \begin{hscode2}
f :: Real(784) -> Real(784)
f image = if one image then image else toOne image
    where one   x = _ 
          toOne x = _ 
    \end{hscode2}
      \vspace{-0.5em}
      \subcaption{Directly programmed neurons}
      \label{fig:direct-program}
    \end{minipage}
  \vspace{1.5em}
  \caption{Programming style and its impact on program structure}
  \label{fig:program-style}
\end{figure}

We could instead define a neural network where we specify a part of the algorithm it will learn, like in Fig. \ref{fig:direct-program}. The program first checks if the input image is a 1. If so, it returns the image. If the input isn't a 1, it transforms the input into a 1. This programming style directly specifies part of the algorithm: a conditional. It then defers to learning where we lack algorithmic intuition, for detecting whether an image is a 1, or transforming an image into a 1. An important point about the direct style is its retreat to ordinary programming. We don't just program with vectors, but also discrete structures---booleans.
Yet no one programs neural networks this way. We lack programming languages where you can.\footnote{Later we discuss related works which are trying to change this (Section \ref{sec:related}).}

\begin{figure}[h]
  \centering
  \begin{minipage}[b]{0.43\textwidth}
    \centering
    \includegraphics[width=\textwidth]{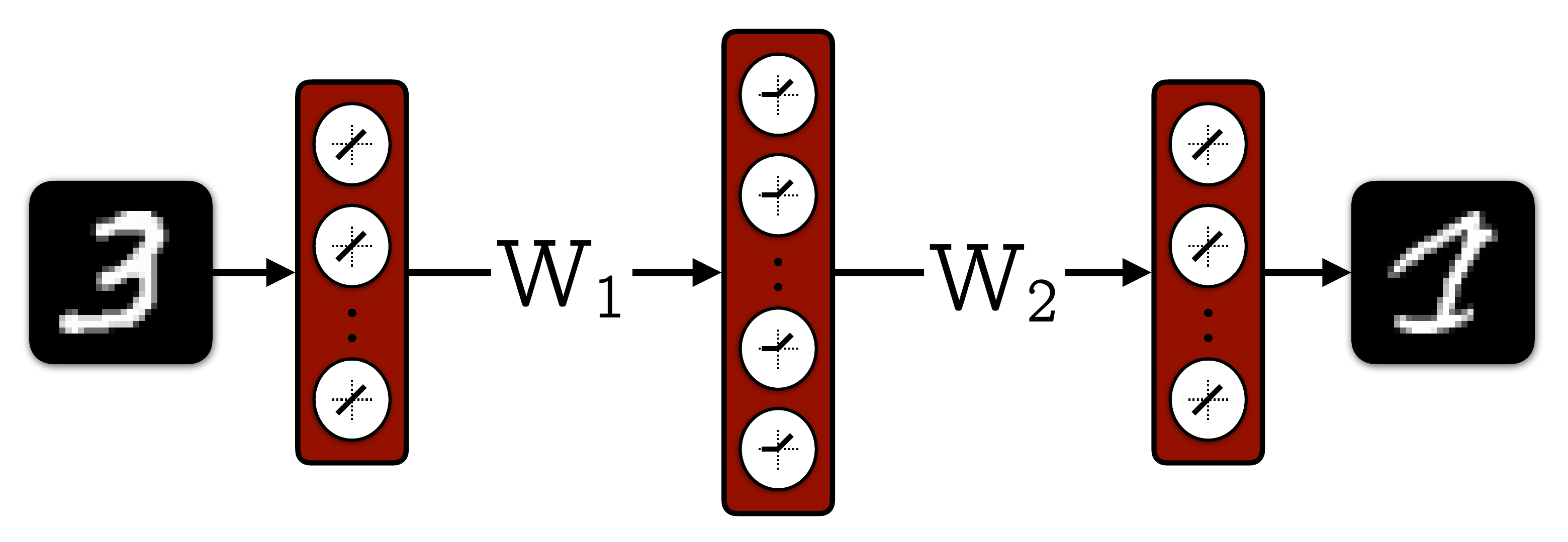} 
    \subcaption{Indirectly programmed neurons}
    \label{fig:indirect-circuit}
    \vspace{3.04em}
  \end{minipage}
  \hfill
  \begin{minipage}[b]{0.52\textwidth}
    \centering
    \includegraphics[width=\textwidth]{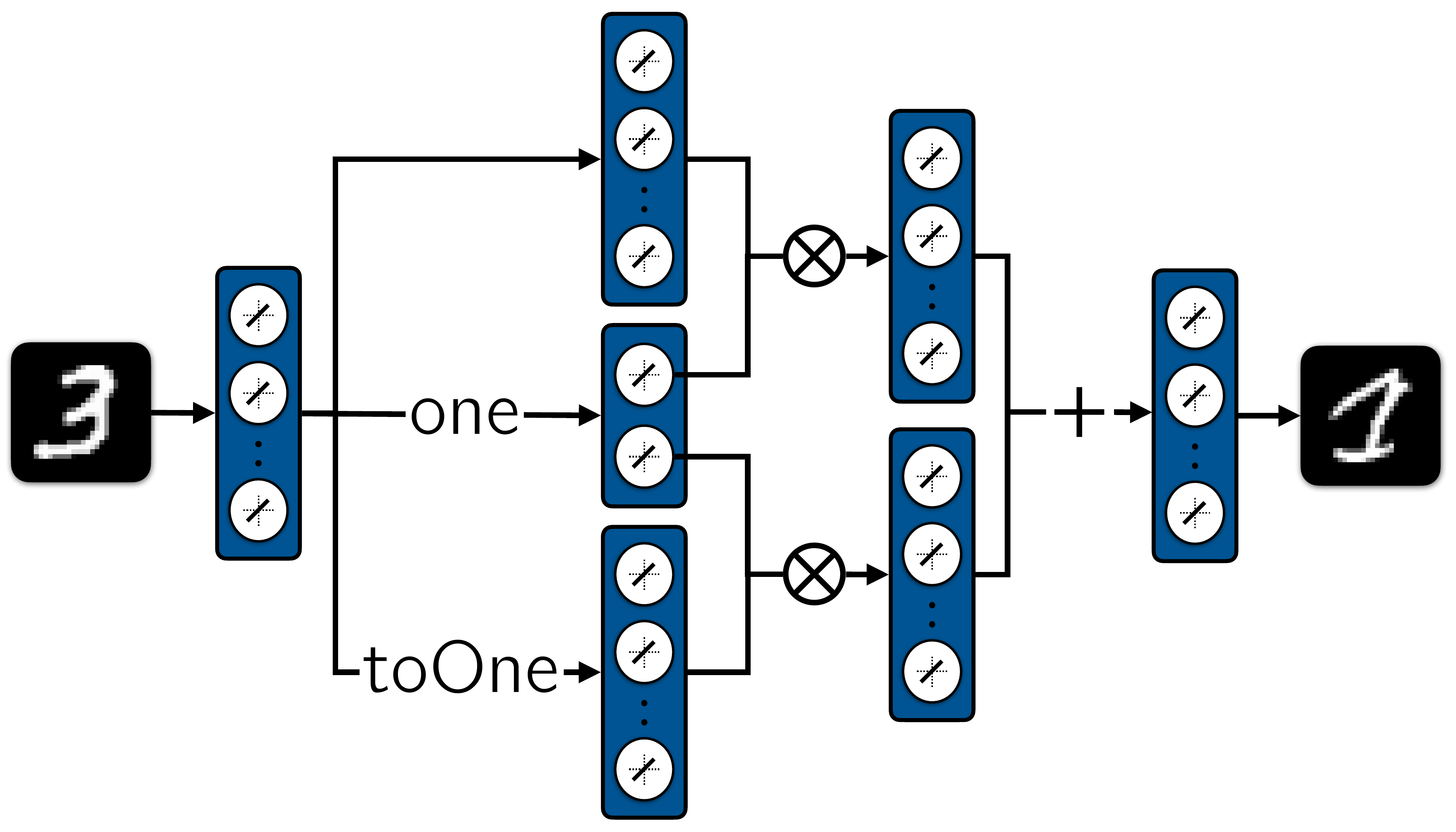} 
    \subcaption{Directly programmed neurons}
    \label{fig:direct-circuit}

    \end{minipage}
  \vspace{1.5em}
  \caption{Programming style and its impact on neural network structure}
  \label{fig:circuits}
\end{figure}

The issue is our compilers.
To define a neural network, your program must, at the very least, compile into a function between vectors. For the indirect style, compiling is simpler. A program like Fig. \ref{fig:indirect-program} is already a function between vectors: essentially no translation occurs when compiling it to Fig. \ref{fig:indirect-circuit}. But for the direct style, compiling is tricky. A program like Fig. \ref{fig:direct-program} uses both vectors and booleans: these booleans and conditionals must translate to vectors and functions over vectors when compiling to Fig. \ref{fig:direct-circuit}.
Of course, the compiler could be wrong. Yet no compiler correctness theory clarifies what it means for a compiler like this to be wrong, or correct.\footnote{There are no compiler correctness theorems or proofs concerning these kinds of compilers.} Beyond correctness, this theory could change how we build and understand neural networks, allowing a rich interplay between learning and the discrete structures of ordinary programming.

We think this theory is hiding in plain sight---the theory of \textit{linear} programming languages \cite{walker2005substructural}.
These languages restrict how a program can use its variables. For example, the Rust programming language restricts the way its programs can reference memory, which eliminates entire classes of memory-related bugs \cite{matsakis2014rust}. But a lesser known property of linear programming languages is that their programs can compile to linear maps: a function between vectors also subject to certain restrictions \cite{valiron2014finite}. In other words, you can define linear maps using programs, even if they contain discrete structures. This is especially powerful when you recognize that neural networks are full of linear maps---in the form of linear neurons \cite{saxe2015deep}. It leads to the most important idea of this paper:
\vspace{.4em}
\begin{quote}
\textit{We can use linear programming languages to directly program neural networks.}
\end{quote}
\vspace{.4em}

Exploring this idea led us to \cj, a minimal, typed, higher-order and linear programming language whose programs compile to linear neurons. This paper explains its design and implementation. 
Specifically, our contributions are:
\setlist[itemize]{topsep=4pt, itemsep=2pt, parsep=2pt, leftmargin=1.5em, rightmargin=1.5em}
\begin{itemize}[label=$\triangleright$]
\item We specify the syntax, evaluation, and typing of \cj\ programs (Section \ref{sec:defining}). 
\item We prove that \cj\ programs compile correctly to linear neurons (Section \ref{sec:compiling}), also discussing the limits of what we can expect of a compiler correctness theorem (Section \ref{sec:linking}).
\item We implement \cj, directly programming a neural network to solve the conditional transform task from Fig. \ref{fig:task} and a conditional classification task (Section \ref{sec:experiments}). 
\end{itemize}

\cj\ is intended as a minimal vehicle for exploring a \textit{theory} of directly programming neural networks. Such a theory would help us address which types may be difficult, if impossible to add, or to understand which kinds of neurons can even be programmed. By contrast, prior work on direct programming has been primarily empirical, focusing on experiments that test whether directly programmed neural networks learn more effectively (Section \ref{sec:related}). These experiments are essential. But they do not address the central question of our work: when do neural networks implement discrete algorithms?





\section{Linear neurons}
\label{sec:lin-neuron}
Before we introduce $\cj$, we want to make clear what linear neurons are. They are linear maps, a special kind of function between vector spaces\footnote{A vector space is a set $V$ with a vector addition $+$ and scalar multiplication $\.$ on $V$, more details in  \cite{axler2015linear}.}. For example, consider the function $f : \R^3 \to \R^2$. 

\[
f\Bigl(\begin{bmatrix}x_1\\x_2\\x_3\end{bmatrix}\Bigr) = \begin{bmatrix}2\.x_1 + 4\.x_2\\ x_3\end{bmatrix}
\]

\vspace{.6em}

It is linear because it satisfies the following property. 

\[
f\Bigl(\alpha \. \begin{bmatrix}y_1\\y_2\\y_3\end{bmatrix} + \beta \. \begin{bmatrix}z_1\\z_2\\z_3\end{bmatrix}\Bigr) = 
\alpha \. f\Bigl(\begin{bmatrix}y_1\\y_2\\y_3\end{bmatrix}\Bigr) + 
\beta \. f\Bigl(\begin{bmatrix}z_1\\z_2\\z_3\end{bmatrix}\Bigr)
\]

\vspace{.5em}

A powerful consequence of this property is that we can understand $f$ in terms of simpler parts.

\vspace{-.3em}
\[
\begin{aligned}
f\Bigl(\begin{bmatrix}1\\2\\3\end{bmatrix}\Bigr) &= 1\.f\Bigl(\begin{bmatrix}1\\0\\0\end{bmatrix}\Bigr) + 2\.f\Bigl(\begin{bmatrix}0\\1\\0\end{bmatrix}\Bigr) + 3\.f\Bigl(\begin{bmatrix}0\\0\\1\end{bmatrix}\Bigr)\\
 \end{aligned}
\]
\vspace{.5em}

Notice how we can decompose the output of $f$ in terms of three observations---its \textit{basis}. Any basis of a vector space allows you to do this decomposition, and for any input.

$$f\Bigl(\begin{bmatrix}1\\0\\0\end{bmatrix}\Bigr)=\begin{bmatrix}2\\0\end{bmatrix} \quad\quad\quad f\Bigl(\begin{bmatrix}0\\1\\0\end{bmatrix}\Bigr)=\begin{bmatrix}4\\0\end{bmatrix} \quad\quad\quad f\Bigl(\begin{bmatrix}0\\0\\1\end{bmatrix}\Bigr)=\begin{bmatrix}0\\1\end{bmatrix}$$
\vspace{.4em}

\noindent For any linear map, it's enough to know how it behaves on the basis. This leads to a powerful alternative for defining linear maps: a matrix. You can define the matrix $W$ of $f$ by stacking column-wise the output of $f$ on each basis vector, as we do in Fig. \ref{fig:correspondence}. To compute $f(x)$ you can instead compute a matrix multiplication $Wx$. This is also how linear neurons compute their output. The entries of a matrix denote the strength of connections between neurons, and their output on an input is exactly $Wx$. Fig. \ref{fig:correspondence} summarizes the correspondence: linear neurons are linear maps.

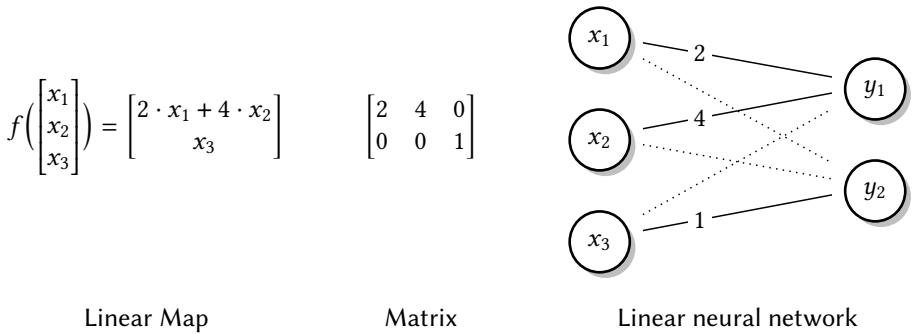
\begin{figure}[h]
\begin{center}
\begin{tabular}{ccc}
$f\Bigl(\begin{bmatrix}x_1\\x_2\\x_3\end{bmatrix}\Bigr) = \begin{bmatrix}2\.x_1 + 4\.x_2\\ x_3\end{bmatrix}$ & 
$\hspace{2em}\begin{bmatrix}
2 & 4 & 0 \\
0 & 0 & 1 
\end{bmatrix}\hspace{2em}$ & 
\begin{tikzcd}[row sep=-1.2em, every label/.append style = {font=\normalsize}]
	{\tikz[baseline=(x.base)] \node[drop shadow, fill=white, circle, draw, line width=1.pt, minimum size=.8cm, inner sep=0pt] (x) {$x_1$};} \\
	&&& {\tikz[baseline=(x.base)] \node[drop shadow, fill=white, circle, draw, line width=1.pt, minimum size=.8cm, inner sep=0pt] (x) {$y_1$};} \\
	{\tikz[baseline=(x.base)] \node[drop shadow, fill=white, circle, draw, line width=1.pt, minimum size=.8cm, inner sep=0pt] (x) {$x_2$};} \\
	&&& {\tikz[baseline=(x.base)] \node[drop shadow, fill=white, circle, draw, line width=1.pt, minimum size=.8cm, inner sep=0pt] (x) {$y_2$};} \\
	{\tikz[baseline=(x.base)] \node[drop shadow, fill=white, circle, draw, line width=1.pt, minimum size=.8cm, inner sep=0pt] (x) {$x_3$};} \\
	{}
	\arrow["2"{description, pos=0.3}, no head, from=1-1, to=2-4]
	\arrow[dotted, no head, from=1-1, to=4-4]
	\arrow["4"{description, pos=0.3}, no head, from=3-1, to=2-4]
	\arrow[dotted, no head, from=3-1, to=4-4]
	\arrow[dotted, no head, from=5-1, to=2-4]
	\arrow["1"{description, pos=0.3}, no head, from=5-1, to=4-4]
\end{tikzcd}\\[6em]
{\normalsize {\textsf{Linear Map}}} & {\normalsize {\textsf{Matrix}}} & {\normalsize {\textsf{Linear neural network}}}
\end{tabular}
\end{center}
\vspace{1em}
\caption{Correspondence between a linear map, matrix, and linear neural network}
\label{fig:correspondence}
\end{figure}

If we consider linear maps between matrices, we can extend this correspondence to \textit{hypernetworks}, neural networks whose outputs are neural networks \cite{ha2016hypernetworks}. For example, consider the following linear map $g$ which maps a real number $r \in \R$ to a matrix $W \in \R^{2 \times 2}$. Later, we will see that these hypernetworks can implement higher-order programs (Section \ref{sec:formal-approach}).

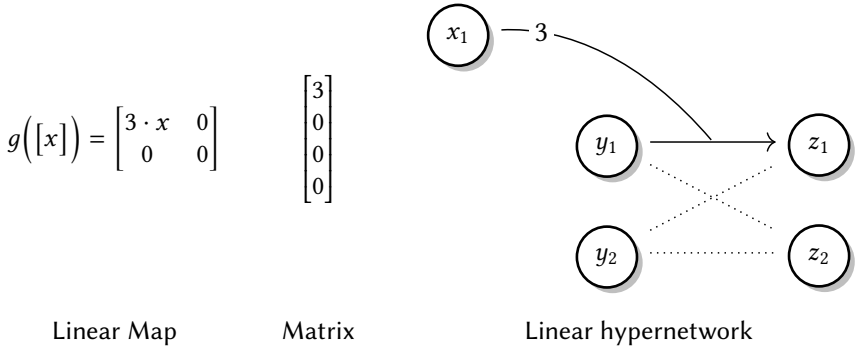
\begin{figure}[h]
\begin{center}
\begin{tabular}{ccc}
$g\Bigl(\begin{bmatrix}x\end{bmatrix}\Bigr) = \begin{bmatrix}3\.x  & 0\\ 0 & 0 \end{bmatrix}$ & 
$\hspace{2em}\begin{bmatrix}
3 \\ 0 \\ 0 \\ 0
\end{bmatrix}\hspace{2em}$ & 
\begin{tikzcd}[row sep=1.em]
	{\tikz[baseline=(x.base)] \node[drop shadow, fill=white, circle, draw, line width=1.pt, minimum size=.8cm, inner sep=0pt] (x1) {$x_1$};} \\
	& {\tikz[baseline=(x.base)] \node[drop shadow, fill=white, circle, draw, line width=1.pt, minimum size=.8cm, inner sep=0pt] (h1) {$y_1$};} && {\tikz[baseline=(x.base)] \node[drop shadow, fill=white, circle, draw, line width=1.pt, minimum size=.8cm, inner sep=0pt] (h2) {$z_1$};} \\
	& {\tikz[baseline=(x.base)] \node[drop shadow, fill=white, circle, draw, line width=1.pt, minimum size=.8cm, inner sep=0pt] (y1) {$y_2$};} && {\tikz[baseline=(x.base)] \node[drop shadow, fill=white, circle, draw, line width=1.pt, minimum size=.8cm, inner sep=0pt] (y2) {$z_2$};}
	\arrow[""{name=0, anchor=center, inner sep=0}, from=2-2, to=2-4]
	\arrow[dotted, no head, from=2-2, to=3-4]
	\arrow[dotted, no head, from=3-2, to=2-4]
	\arrow[dotted, no head, from=3-2, to=3-4]
	\arrow["\text{\large 3}"{description, pos=0.2}, curve={height=-18pt}, no head, from=1-1, to=0]
\end{tikzcd}\\[6em]
{\normalsize {\textsf{Linear Map}}} & {\normalsize {\textsf{Matrix}}} & {\normalsize {\textsf{Linear hypernetwork}}}
\end{tabular}
\end{center}
\vspace{1em}
\caption{Correspondence between a linear map, matrix, and linear hypernetwork}
\vspace{.5em}
\label{fig:correspondence}
\end{figure}

However, these neat correspondences break for nonlinear maps and nonlinear neural networks. It's not enough to know how a nonlinear map behaves on a basis. We cannot reduce its behavior to simpler observations as before. There is no analog of a matrix for nonlinear maps. As a result, directly constructing a nonlinear neural network from a nonlinear map is difficult. Universal approximation theorems\footnote{These theorems identify the class of functions a neural network can learn to approximate.} for neural networks sometimes rely on constructive techniques for directly building neural networks which approximate nonlinear maps \cite{cybenko1989approximation}. But they are difficult to exploit in practice, due to their asymptotics \cite{chong2020closer}. This is why nonlinear neural networks are typically built through \textit{learning}. Nevertheless, nonlinear neural networks contain lots of linear structure, and the overarching theme of this work is that the linear structure can be directly programmed.

\section{Linear programs}
\label{sec:lin-prog}
\cj\ is a linear programming language---a language with \textit{linear} types \cite{walker2005substructural}. These types ensure that programs use a variable exactly once when they evaluate. \footnote{There are some exceptions, but the intuition is sound for this paper.} This means linear programs can't duplicate or discard their variables. It's what links \textit{linear} programs and \textit{linear} maps.

\begin{figure}[h]
  \centering
  \begin{minipage}[b]{0.42\textwidth}
    \centering
    \begin{hscode1}
fls :: Bool -> Bool
fls x = if x then False else x
    \end{hscode1}
    \vspace{-.5em}
    \subcaption{Programming nonlinearly}
    \label{fig:nonlinear-dup}
  \end{minipage}
  \hfill
  \begin{minipage}[b]{0.45\textwidth}
      \centering
    \begin{hscode2}
fls :: Bool -> Bool
fls x = if x then False else False
    \end{hscode2}
      \vspace{-0.5em}
      \subcaption{Programming linearly}
      \label{fig:linear-dup}
    \end{minipage}
  \vspace{1.0em}
  \caption{Programming with and without duplicating/discarding variables}
  \label{fig:prog-duplication}
\end{figure}

Consider the linear and nonlinear version of \nb{\texttt{fls}} in Fig. \ref{fig:prog-duplication}. The nonlinear version in Fig. \ref{fig:nonlinear-dup} duplicates \texttt{x} for each branch. If applied to an argument, the function will use \texttt{x} when checking the condition and when returning a value. But the linear version in Fig. \ref{fig:linear-dup} doesn't duplicate \texttt{x}. If applied to an argument, the function will only use \texttt{x} when checking the condition. Additionally, the nonlinear version in Fig. \ref{fig:nonlinear-dup} discards the duplicated \texttt{x} (from its available context) when returning \t{ff} in the then branch of the conditional. But the linear version in Fig. \ref{fig:linear-dup} doesn't discard anything.
 
\begin{figure}[h]
  \centering
  \begin{minipage}[b]{0.45\textwidth}
    \centering

        \scalebox{.9}{
        \def\extraVskip{4pt}
        \def\defaultHypSeparation{\hskip .35in}
        \bottomAlignProof
        \AxiomC{}
        \UnaryInfC{$\p{\textcolor{red}{\bind{x}{\Bool}}} {\bl{x}} {\bl{\Bool}}$}
        \AxiomC{}
        \UnaryInfC{$\p{\textcolor{red}{\bind{x}{\Bool}}} {\bl{\t{ff}}} {\bl{\Bool}}$}
        \AxiomC{}
        \UnaryInfC{$\p{\textcolor{red}{\bind{x}{\Bool}}} {\bl{x}} {\bl{\Bool}}$}
        \LeftLabel{\textcolor{red}{\huge \xmark}}
        \TrinaryInfC{$\p{\textcolor{red}{\bind{x}{\Bool}}} {\bl{\ite{\ul{x}}{\ul{\t{ff}}}{\ul{x}}}} {\bl{\Bool}}$}
        \UnaryInfC{$\p{\bl{\emp}} {\bl{\lam{x}{\ite{\ul{x}}{\ul{\t{ff}}}{\ul{x}}}}} {\bl{\Bool \to \Bool}}$}
        \DisplayProof}
    \vspace{.5em}
    \subcaption{Typing nonlinearly}
    \label{fig:typ-nonlinear-dup}
  \end{minipage}
  \hfill
  \begin{minipage}[b]{0.45\textwidth}
      \centering

        \scalebox{.9}{
        \def\extraVskip{4pt}
        \def\defaultHypSeparation{\hskip .35in}
        \bottomAlignProof
        \AxiomC{}
        \UnaryInfC{$\p{\textcolor{ForestGreen}{\bind{x}{\Bool}}} {\bl{x}} {\bl{\Bool}}$}
        \AxiomC{}
        \UnaryInfC{$\p{\bl{\emp}} {\bl{\t{ff}}} {\bl{\Bool}}$}
        \AxiomC{}
        \UnaryInfC{$\p{\bl{\emp}} {\bl{\t{ff}}} {\bl{\Bool}}$}
        \LeftLabel{\textcolor{ForestGreen}{\huge \cmark}}
        \TrinaryInfC{$\p{\textcolor{ForestGreen}{\bind{x}{\Bool}}} {\bl{\ite{\ul{x}}{\ul{\t{ff}}}{\ul{\t{ff}}}}}{\bl{\Bool}}$}
        \UnaryInfC{$\p{\bl{\emp}} {\bl{\lam{x}{\ite{\ul{x}}{\ul{\t{ff}}}{\ul{\t{ff}}}}}} {\bl{\Bool \lto \Bool}}$}
        \DisplayProof}

      \vspace{.5em}
      \subcaption{Typing linearly}
      \label{fig:typ-linear-dup}
    \end{minipage}
  \vspace{1.0em}
  \caption{Typing with and without duplicating/discarding variables}
  \label{fig:typ-duplication}
\end{figure}

Typing can detect variable duplication and discarding. The nonlinear typing in Fig. \ref{fig:typ-nonlinear-dup} duplicates the variable $\textcolor{red}{\bind{x}{\Bool}}$ when typing the conditional. It shares a context across each subprogram.
But the linear typing in Fig. \ref{fig:typ-linear-dup} doesn't duplicate $\textcolor{ForestGreen}{\bind{x}{\Bool}}$ when typing the conditional. It splits a context across each subprogram. Additionally, the nonlinear typing in Fig. \ref{fig:typ-nonlinear-dup} discards the variable $\textcolor{red}{\bind{x}{\Bool}}$ when typing $\t{ff}$ in the then branch of the conditional. But the linear typing in Fig. \ref{fig:typ-linear-dup} doesn't discard anything across each subprogram. 

To enforce these restrictions generally, linear typing must ensure that the structural rules of \textit{contraction} and \textit{weakening} are not permissible. \cite{pierce2002types} Their impermissibility establish a remarkable connection between linear programs, linear maps, and linear neurons. However, not just any linear programming language can compile to linear neurons. Beyond maintaining a connection to linear maps, a linear programming language must ensure its connection is to linear maps between vector spaces over real numbers, and that these linear maps are effectively computable\footnote{There shouldn't be exponential scaling hiding somewhere.}. These are the central tensions in balance when designing $\cj$.
\newcommand{\ttvm}{\rd{\begin{bmatrix} 1 \\ 0\end{bmatrix}}}
\newcommand{\ffvm}{\rd{\begin{bmatrix} 0 \\ 1\end{bmatrix}}}
\newcommand{\bvm}[2]{\rd{\begin{bmatrix} #1 \\ #2\end{bmatrix}}}

\section{\texorpdfstring{Cajal{\small$(\nb{\Bool},\nb{\lto})$}}{Cajal}}

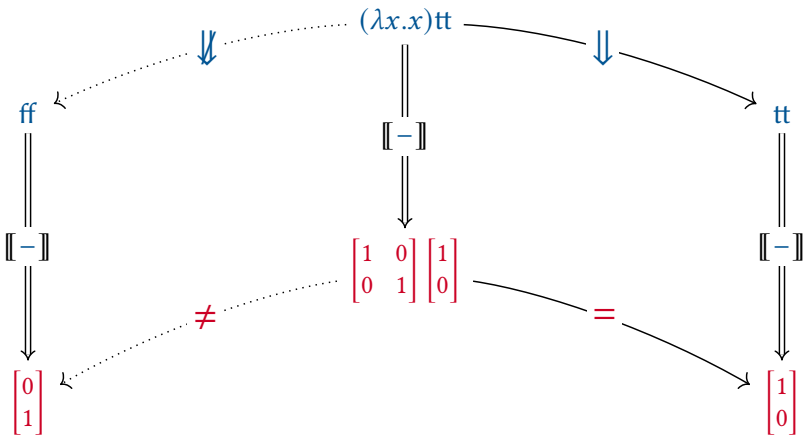
\begin{figure}[h]
\[\begin{tikzcd}
	&&&& {\text{\large \nb{$(\lambda x.x)\mathsf{tt}$}}} \\
	{\text{\large \nb{$\mathsf{ff}$}}} &&&&&&&& {\text{\large \nb{$\mathsf{tt}$}}} \\
	\\
	&&&& \begin{array}{c} \rd{\begin{bmatrix}1 \!\!\!\!&\!\!\!\! 0\\ 0 \!\!\!\!&\!\!\!\! 1\end{bmatrix}\begin{bmatrix}1\\0\end{bmatrix}} \end{array} \\
	\begin{array}{c} \rd{\begin{bmatrix}0\\1\end{bmatrix}} \end{array} &&&&&&&& \begin{array}{c} \rd{\begin{bmatrix}1\\0\end{bmatrix}} \end{array}
	\arrow["{\scalebox{2.0}{\nb{$\not\bstep$}}}"{description}, curve={height=10pt}, shorten >=2pt, dotted, from=1-5, to=2-1]
	\arrow["{\scalebox{2.0}{\nb{$\bstep$}}}"{description},shorten >=2pt, curve={height=-10pt}, from=1-5, to=2-9]
	\arrow["{\scalebox{1.4}{{$\c{\nb{-}}$}}}"{description}, Rightarrow, from=1-5, to=4-5]
	\arrow["{\scalebox{1.4}{{$\c{\nb{-}}$}}}"{description}, Rightarrow, from=2-1, to=5-1]
	\arrow["{\scalebox{1.4}{{$\c{\nb{-}}$}}}"{description}, Rightarrow, from=2-9, to=5-9]
	\arrow["{\scalebox{2.0}{\rd{$\neq$}}}"{description}, shorten >=-5pt, shorten <= -5pt, curve={height=10pt}, dotted, from=4-5, to=5-1]
	\arrow["{\scalebox{2.0}{\rd{$=$}}}"{description}, shorten >=-5pt, shorten <= -5pt,curve={height=-10pt}, from=4-5, to=5-9]
\end{tikzcd}\]

\caption{Predicting the behavior of compiled neurons}
\label{fig:ccc1}
\end{figure}

\cj\ programs compile correctly to linear neurons. In this section we clarify what that means and how to prove it. Fig. \ref{fig:ccc1} details what we want. If we know how \cj\ programs will $(\nb{\bstep})$ and won't $(\nb{\not\bstep})$ evaluate, we should be able to predict how their compiled neurons will $(\rd{=})$ and won't $(\rd{\neq})$ evaluate. If the compiler $(\c{\nb{-}})$ permits these predictions, then we say it is correct.
\setlist[itemize]{topsep=4pt, itemsep=2pt, parsep=2pt, leftmargin=1.5em, rightmargin=1.5em}

\subsection{Defining}
\label{sec:defining}

\begin{figure}[h]
  \begin{mdframed}[topline=false,innertopmargin=-1.07ex,innerleftmargin=-0.1ex,innerrightmargin=0ex,linewidth=.8pt,skipbelow=2ex]
    \rulediv{18pt}{\large \textsf{Syntax}}
    \vspace{.1em}
    \[  
    \begin{tabular}{c c}
        \scalebox{.8}{
        $\begin{aligned}
          &\;\;\;\;\;\;\;\;\;\nb{e} \in \nb{\t{Expression}}\\[3mm]
          \nb{e} \,\,\,\coloneqq 
	       \ & \ \ \ \nb{x} \in \t{\nb{Var}} &&\t{variable}\\
	       &\mid \nb{\tt} &&\t{true}\\
	       &\mid \nb{\ff} &&\t{false}\\
	       &\mid \nb{\ite{e_1}{e_2}{e_3}} &&\t{conditional}\\
	       &\mid \nb{\lam{x}{e}} &&\t{linear map}\\
	       &\mid \nb{e_1e_2} &&\t{linear application}\\
        \end{aligned}$}
    &
        \scalebox{.8}{
        $\begin{aligned}
          &\;\;\;\;\;\;\;\;\;\nb{\tau} \in \nb{\t{Type}}\\[3mm]
          \nb{\tau} \,\,\,\coloneqq
	       &\ \ \ \nb{\mathbb{2}} &&\t{boolean}\\
	       &\mid \nb{\tau_1 \lto \tau_2} &&\t{linear map}\\
        \end{aligned}$}
    \\[19mm]
        \scalebox{.8}{
        $\begin{aligned}
          &\;\;\;\;\;\;\;\;\;\nb{v} \in \nb{\t{Value}}\\[3mm]
          \nb{v} \,\,\,\coloneqq 
	       &\ \ \ \nb{\tt} && \t{true}\\
	       &\mid \nb{\ff} && \t{false}\\
	       &\mid \nb{\lam{x}{e}} && \t{linear map}
        \end{aligned}$}
    &
        \scalebox{.8}{
        $\begin{aligned} 
          &\;\;\;\;\;\;\;\;\;\nb{\D} \in \nb{\t{Context}}\\[3mm]
          \nb{\D} \,\,\,\coloneqq
	       &\ \ \ \nb{\emp} &&\t{empty}\\
	       &\mid \nb{\D,\bind{x}{\ty}} &&\t{binder}\\
        \end{aligned}$}
    \\[10mm]
    \end{tabular}
    \]
  \end{mdframed}
  \vspace{.8em}
  \caption{Syntax of $\cj${\scriptsize $(\nb{\Bool},\nb{\lto})$}}
  \label{fig:syntax}
\end{figure}

\subsubsection{Syntax} \textcolor{bmidnight}{\t{Cajal}} is a minimal, typed, higher-order and linear programming language. Fig. \ref{fig:syntax} details its syntax, equivalent to a simply-typed lambda calculus with booleans $(\nb{\Bool})$.
We want \cj\ to be easy to understand and iterate from. Hence we focus on a minimal language, the simplest functional programming language where you can directly program neural networks. In Section \ref{sec:discussion} we discuss simple extensions. They are absent here because they obscure the theory for little benefit. 

\subsubsection{Evaluating} The big-step evaluation relation in Fig. \ref{fig:evaluating} defines how \cj\ expressions evaluate. We chose a big-step relation because it simplifies proofs of type soundness and compiler correctness. The choice of call-by-value evaluation strategy was arbitrary. Nothing we discuss in this section hinges on that choice. The notation $\nb{\{x \map v\}(e)}$ describes a substitution, where $\nb{x}$ is a variable replaced by the value $\nb{v}$ in the expression $\nb{e}$. To simplify the proofs we assume every variable name is unique, eliding details concerning variable capture \cite{barendregt1984lambda}. 

\begin{figure}[h]
  \begin{mdframed}[topline=false,innertopmargin=-1.15ex,innerleftmargin=-0.1ex,innerrightmargin=0ex,linewidth=.8pt,skipbelow=2ex]
    \rulediv{25pt}{\large \textsf{Evaluating}}
    \vspace{.1em}
    \[  
    \begin{tabular}{c}
        \scalebox{.88}{
        $\step {e} {v} \iff \t{expression }\nb{e} \t{ evaluates to }\nb{v}$}
    \\[5mm]
        \scalebox{.8}{
        \def\extraVskip{3pt}
        \def\defaultHypSeparation{\hskip .05in}
        \bottomAlignProof
        \AxiomC{}
        \UnaryInfC{$\step{v}{v}$}
        \DisplayProof}
    \quad
        \scalebox{.8}{
        \def\extraVskip{3pt}
        \def\defaultHypSeparation{\hskip .05in}
        \bottomAlignProof
        \AxiomC{$\step{e_1}{\tt}$}
        \AxiomC{$\step {e_2} {v_2}$}
        \BinaryInfC{$\step  {\ite{e_1}{e_2}{e_3}}  {v_2}$}
        \DisplayProof}
    \quad
        \scalebox{.8}{
        \def\extraVskip{3pt}
        \def\defaultHypSeparation{\hskip .05in}
        \bottomAlignProof
        \AxiomC{$\step{e_1}{\ff}$}
        \AxiomC{$\step {e_3} {v_3}$}
        \BinaryInfC{$\step  {\ite{e_1}{e_2}{e_3}}  {v_3}$}
        \DisplayProof}
    \\[5mm]
        \scalebox{.8}{
        \def\extraVskip{3pt}
        \def\defaultHypSeparation{\hskip .05in}
        \bottomAlignProof
        \AxiomC{$\step {e_1} {\lam{x}{e}}$}
        \AxiomC{$\step {e_2} {v_2}$}
        \AxiomC{$\step {\{x \map v_2\}(e)} {v}$}
        \TrinaryInfC{$\step {e_1e_2} {v}$}
        \DisplayProof}
    \\[5mm]
    \end{tabular}
    \]
  \end{mdframed}
  \vspace{.8em}
  \caption{Evaluating $\cj${\scriptsize $(\nb{\Bool},\nb{\lto})$} programs}
  \label{fig:evaluating}
\end{figure}

\begin{figure}[h]
  \begin{mdframed}[topline=false,innertopmargin=-1.08ex,innerleftmargin=-0.1ex,innerrightmargin=0ex,linewidth=.8pt,skipbelow=2ex]
    \rulediv{18pt}{\large \textsf{Typing}}
    \vspace{0.1em}
    \[  
    \begin{tabular}{c}
        \scalebox{.88}{
        $\p{\D} {e} {\ty} \iff \t{program }\nb{e} \t{ has type }\nb{\ty}$}
    \\[5mm]
        \scalebox{.8}{
        \def\extraVskip{3pt}
        \def\defaultHypSeparation{\hskip .05in}
        \bottomAlignProof
        \Axiom$\fCenter$
        \UnaryInf$\fCenter \p{\bind{x}{\ty}} {x} {\ty}$
        \DisplayProof}
    \quad
        \scalebox{.8}{
        \def\extraVskip{3pt}
        \def\defaultHypSeparation{\hskip .05in}
        \bottomAlignProof
        \Axiom$\fCenter$
        \UnaryInf$\fCenter \p{\emp} {\tt} {\Bool}$
        \DisplayProof}
    \quad
        \scalebox{.8}{
        \def\extraVskip{3pt}
        \def\defaultHypSeparation{\hskip .05in}
        \bottomAlignProof
        \Axiom$\fCenter$
        \UnaryInf$\fCenter \p{\emp} {\ff} {\Bool}$
        \DisplayProof}
    \quad
        \scalebox{.8}{
        \def\extraVskip{3pt}
        \def\defaultHypSeparation{\hskip .05in}
        \bottomAlignProof
        \AxiomC{$\p{ \D,\bind{x}{\ty_1}} {e} {\ty_2}$}
        \UnaryInfC{$\p{\D} {\lam{x}{e}} {\ty_1 \lto \ty_2}$}
        \DisplayProof}
    \\[5mm]
        \scalebox{.8}{
        \def\extraVskip{3pt}
        \def\defaultHypSeparation{\hskip .05in}
        \bottomAlignProof
        \AxiomC{$\p{\D_1} {e_1} {\ty_1 \lto \ty_2}$}
        \AxiomC{$\p{\D_2} {e_2} {\ty_1}$}
        \BinaryInfC{$\p{\D_1\o\D_2} {e_1e_2} {\ty_2}$}
        \DisplayProof}
    \quad
        \scalebox{.8}{
        \def\extraVskip{3pt}
        \def\defaultHypSeparation{\hskip .05in}
        \bottomAlignProof
        \AxiomC{$\p{\D_1} {e_1} {\Bool}$}
        \AxiomC{$\p{\D_2} {e_2} {\ty}$}
        \AxiomC{$\p{\D_2} {e_3} {\ty}$}
        \TrinaryInfC{$\p{\D_1\o\D_2} {\ite{e_1}{e_2}{e_3}} {\ty}$}
        \DisplayProof}
    \\[5mm]
    \end{tabular}
    \]
  \end{mdframed}
  \vspace{.8em}
  \caption{Typing $\cj${\scriptsize $(\nb{\Bool},\nb{\lto})$} programs}
  \label{fig:typing}
\end{figure}

\subsubsection{Typing} The typing relation in Fig. \ref{fig:typing} defines what a \cj\ program is. Throughout this paper, programs always refer to a typing derivation $\p{\D}{e}{\ty}$. And these programs are similar to those from a simply-typed lambda calculus. However, the key difference is the presence of a context splitting relation $\nb{\o}$ which takes two contexts and ensures their domains are disjoint. For example, $\nb{(\bind{x}{\Bool}, \bind{y}{\Bool}) \o (\bind{z}{\Bool})}$ because $\{\nb{x},\nb{y}\}$ and $\{\nb{y}\}$ are disjoint sets. However, $\nb{(\bind{x}{\Bool}, \bind{y}{\Bool}) \o (\bind{y}{\Bool})}$ is not in the context splitting relation because $\{\nb{x},\nb{y}\}$ and $\{\nb{y}\}$ both contain $\nb{y}$ (see Appendix A.1 for definition of $\nb{\D_1 \o \D_2}$). The context splitting relation allows us to ensure contraction (variable duplication) is not permissible in our programs (Section \ref{sec:lin-prog}). 

With the typing relation we can identify the expressions with behavior (can evaluate to values). For example, $\p{\emp}{(\lam{x}{x})\tt}{\Bool}$ is a program. It evaluates to $\tt$ because $\step{(\lam{x}{x})\tt}{\tt}$. Even expressions with free variables can evaluate, so long as we provide a source environment $\nb{\sigma}$ that maps free variables to values. For example, $\nb{\sigma(x)} \bstep \tt$ if the source environment $\nb{\sigma}$ is $\nb{\{x \map \tt\}}$ (see Appendix A.2 for definition of $\nb{\sigma}$). This is important because the values programs evaluate to can be used to generate predictions for what compiled neurons should do. It's essential to our notion of compiler correctness. The following theorem ensures every program has values that can be used as a basis for this prediction. Note that the relation $\nb{\D} \- \nb{\sigma}$ ensures that source environments map variables to values of the type specified in the context (also defined in Appendix A.2). For example, $\nb{\bind{x}{\Bool}} \- \nb{\{x \map \tt\}}$ is in the relation. However, $\nb{\bind{x}{\Bool}} \- \nb{\{x \map \lam{y}{y}\}}$ is not because $\nb{\lam{y}{y}}$ is not type $\nb{\Bool}$.

\vspace{.5em}
\begin{bthm}{\textsf{(Programs evaluate to values)}}{wef}
\vspace{.25em}
\t{If }$\p{\D}{e}{\ty}$ \t{ and } $\nb{\D} \vdash \nb{\sigma}$ \t{, then } $\exists \nb{v}, \step{\sigma(e)}{v}$  
\end{bthm}

\vspace{.3em}

\noindent\ul{\textsf{Outline of Proof.}}

The proof is nearly identical to the logical relations proof of termination for simply-typed lambda calculus \cite{pierce2002types}. It differs only due to context splitting $\nb{\D_1 \o \D_2}$. A detailed proof is in Appendix B.5.

\subsection{Compiling}
\label{sec:compiling}

Compiling to linear neurons requires us to map programs to linear maps. Before presenting our solution, it's helpful to first consider an incorrect approach. This allows us to sketch out the important choices en route to our compiler correctness theorem.

\subsubsection{An incorrect approach} 

The first key choice our compiler makes is to map each type to a vector space of \textit{finite} dimension. The vector space's dimension relates to the number of neurons needed to implement values of that type. Therefore, vector spaces with infinite dimension are a non-starter, they would need an infinite number of neurons. Additionally, their dimension, even if finite, must scale effectively as a function of types. The number of neurons needed to implement a value of a type should not grow exponentially as types grow more complex. Otherwise to implement simple higher-order functions may quickly require too many neurons. With this in mind, let's try the following. We map all types to vector spaces whose dimension is 1.

\[
\begin{aligned}
\c{\nb{\Bool}} &= \rd{\R} &&=\{\rd{1}, \rd{4}, \rd{\dots}\}\\
\c{\nb{\ty_1 \lto \ty_2}} &= \rd{\R^{1\times1}} \!\!\!\!\!\!\!&&= \{\rd{\begin{bmatrix}3\end{bmatrix}},\rd{\begin{bmatrix}8\end{bmatrix}},\rd{\dots}\}
\end{aligned}
\]
\vspace{.1em}

The second key choice our compiler makes is to map each value to a vector in the vector space determined by its type. Correctness hinges on how we map $\nb{\tt}$ and $\nb{\ff}$. They must\footnote{Later, we discuss how to relax this requirement (Section \ref{sec:discussion}).} injectively map to a \textit{basis}. Otherwise it won't be possible to find linear maps that correspond to functions over booleans. For example, consider this attempt.

\vspace{.3em}
\setlist[itemize]{topsep=4pt, itemsep=4pt, parsep=8pt, leftmargin=1.5em, rightmargin=1.5em}
\begin{itemize}[label=$\triangleright$]
    \item For $\nb{\tt}$ we pick arbitrarily from $\rd{\R}$, 
    $$\c{\nb{\tt}} = \rd{2}$$
    \item For $\nb{\ff}$ we pick arbitrarily from $\rd{\R}$, 
    $$\c{\nb{\ff}} = \rd{5}$$
    \item For $\nb{\lam{x}{x}}$ we need a matrix $\rd{W}$ where $\rd{W}\c{\tt} = \c{\tt}$ and $\rd{W}\c{\ff} = \c{\ff}$, 
    \vspace{.4em}
    $$\c{\nb{\lam{x}{x}}}=\rd{\begin{bmatrix}1\end{bmatrix}}$$
    \item For $\nb{\lam{x}{\ite{\ul{x}}{\ul{\tt}}{\ul{\tt}}}}$ we need a matrix $\rd{W}$ where $\rd{W}\c{\tt} = \c{\tt}$ and $\rd{W}\c{\ff} = \c{\tt}$. But this $\rd{W}$ \textit{cannot} exist because it would violate both the preservation of vector addition and scalar multiplication by linear maps.
\end{itemize}

These issues arise because we didn't injectively map $\nb{\tt}$ and $\nb{\ff}$ to a basis; $\{\rd{2},\rd{5}\}$ is not a basis for $\rd{\R}$. Worse, a basis for $\rd{\R}$ contains only one vector. Therefore no such injective map exists. But we can correct these issues, so long as we compile types to vector spaces whose dimension match the number of \textit{distinct}\footnote{Distinct values aren't contextually equivalent.} values of a type. As a result, we can always find an injective map from values to a basis. In turn, there will be linear maps whose behavior mirrors \cj\ programs. Let's try again.

\subsubsection{A correct approach}
Recall the first key choice our compiler makes is to map each type to a vector space of finite dimension. The type $\nb{\Bool}$ has two distinct values $\nb{\tt}$ and $\nb{\ff}$. Therefore we need a vector space whose basis contains at least two elements---$\rd{\R^2}$ is a natural choice. The type $\nb{\ty_1 \lto \ty_2}$, if it really defines linear maps, contains $\rd{m\.n}$ distinct values, where $\rd{\t{dim}(\nb{\ty_2})}=\rd{m}$ is the number of distinct values at input type $\nb{\ty_1}$ and $\rd{\t{dim}(\nb{\ty_1})}=\rd{n}$ is the number of distinct values at input type $\nb{\ty_1}$. At this type $\rd{\R^{m\times n}}$ is a natural choice---a vector space of matrices with $\rd{m}$ rows and $\rd{n}$ columns. The next few examples demonstrate this.
\[
\begin{aligned}
\c{\nb{\Bool}} &= \rd{\R^{2}}\\
\c{\nb{\ty_1 \lto \ty_2}} &= \rd{\R^{\rd{\t{dim}}(\nb{\ty_2})\,\rd{\x}\,\rd{\t{dim}}(\nb{\ty_1})}}\\
\end{aligned}
\quad\quad\quad\quad
\begin{aligned}
\rd{\t{dim}}(\nb{\Bool}) &= \rd{2}\\
\rd{\t{dim}}(\nb{\ty_1 \lto \ty_2}) &= \rd{\t{dim}}(\nb{\ty_1})\,\rd{\.}\,\rd{\t{dim}}(\nb{\ty_2})\\
\end{aligned}
\]
\vspace{.1em}

Recall the second key choice our compiler makes is to map each value to a vector in the vector space determined by its type. By an informed choice on compiling types, we elide issues from the incorrect approach.

\vspace{.3em}
\setlist[itemize]{topsep=4pt, itemsep=4pt, parsep=8pt, leftmargin=1.5em, rightmargin=1.5em}
\begin{itemize}[label=$\triangleright$]
    \item For $\nb{\tt}$ we pick injectively from a basis\footnote{Throughout this paper we use orthonormal bases \cite{axler2015linear}, which simplify the process of extracting the coefficients of matrices.} of $\rd{\R^2}$, 
    $$\c{\nb{\tt}} = \ttvm$$
    \item For $\nb{\ff}$ we pick injectively from a basis of $\rd{\R^2}$, 
    $$\c{\nb{\ff}} = \ffvm$$
    \item For $\nb{\lam{x}{x}}$ we need a matrix $\rd{W}$ where $\rd{W}\c{\tt} = \c{\tt}$ and $\rd{W}\c{\ff} = \c{\ff}$, 
    \vspace{.4em}
    $$\c{\nb{\lam{x}{x}}}=\rd{\begin{bmatrix}1 \!\!&\! 0\\ 0 \!\!&\! 1\end{bmatrix}}$$
    \item For $\nb{\lam{x}{\ite{\ul{x}}{\ul{\tt}}{\ul{\tt}}}}$ we need a matrix $\rd{W}$ where $\rd{W}\c{\tt} = \c{\tt}$ and $\rd{W}\c{\ff} = \c{\tt}$.
    \vspace{.4em}
    $$\c{\nb{\lam{x}{\ite{\ul{x}}{\ul{\tt}}{\ul{\tt}}}}}=\rd{\begin{bmatrix}1 \!\!&\! 1 \\ 0 \!\!&\! 0\end{bmatrix}}$$
\end{itemize}

\subsubsection{A formal approach}
\label{sec:formal-approach}

The compiler in Fig. \ref{fig:compiling} turns these intuitions into an algorithm for compiling \cj\ programs to linear neurons. It also addresses the last key choice our compiler makes, how to compile expressions. The compiler maps a program $\p{\D}{e}{\ty}$ and a target environment $\sv{}$ to neurons $\et{}$.\footnote{Technically, the compiler is a relation. But we refer to it as mapping programs throughout the paper, consistent with how an implementation would behave.} These target environments map variables $\nb{x}$ to vectors in the vector space denoted by its type. For example, $\nb{\bind{x}{\Bool}} \- \{\nb{x} \map\! \bv{4}{8}$\} is a target environment mapping $\nb{x}$ to $\bv{4}{8} \in \c{\nb{\Bool}}$ (see Appendix A.3 for definition of $\nb{\D} \- \sv{}$). The compiler is only well-behaved on target environments satisfying this relation. We will demonstrate the compiler's behavior with several examples, and then a theorem justifying that it is correct. 

\begin{figure}[h]
  \begin{mdframed}[topline=false,innertopmargin=-1.15ex,innerleftmargin=-0.1ex,innerrightmargin=0ex,linewidth=.8pt,skipbelow=2ex]
    \rulediv{25pt}{\large \textsf{Compiling}}
    \vspace{0.1em}
    \[  
    \begin{tabular}{c}
    \scalebox{.88}{
    $\limp{\p{\D} {e} {\ty}}{\sv{}}{\et{}} \iff \t{program }\nb{e} \t{ compiles to neurons }\rd{\et{}}$}
    \\[5mm]
        \scalebox{.8}{
        \def\extraVskip{4pt}
        \def\defaultHypSeparation{\hskip .05in}
        \bottomAlignProof
        \Axiom$\fCenter$
        \UnaryInf$\fCenter \limp{\p{\bind{x}{\ty}}  {x}  {\ty}} {\{\nb{x}\map \vt{}\}} {\vt{}}$
        \DisplayProof}
    \quad
        \scalebox{.8}{
        \def\extraVskip{4pt}
        \def\defaultHypSeparation{\hskip .05in}
        \bottomAlignProof
        \Axiom$\fCenter$
        \UnaryInf$\fCenter \limp{\p{\emp} {\tt} \Bool}{\emp}{\begin{bmatrix} 1 &\!\!\! 0\end{bmatrix}^\tr}$
        \DisplayProof}
    \quad
        \scalebox{.8}{
        \def\extraVskip{4pt}
        \def\defaultHypSeparation{\hskip .05in}
        \bottomAlignProof
        \Axiom$\fCenter$
        \UnaryInf$\fCenter \limp{\p{\emp} {\ff} \Bool}{\emp}{\begin{bmatrix} 0 &\!\!\! 1\end{bmatrix}^\tr}$
        \DisplayProof}
    \\[5mm]
        \scalebox{.8}{
        \def\defaultHypSeparation{\hskip .05in}
        \def\extraVskip{1pt}
        \bottomAlignProof
        \Axiom$\fCenter \forall \,\bin{j} \in \rd{\t{basis}}\c{\nb{\ty_1}},\;\;\;\;\;\; \bout{i} \in \rd{\t{basis}}\c{\nb{\ty_2}}$
        \noLine
        \UnaryInf$\fCenter \limp{\p{\D,\bind{x}{\ty_1}} {e}  {\ty_2}}{\sv{}\{\nb{x} \map \bin{j}\}}{\sum_i\alpha_{ij}\.\bout{i}}$
        \def\extraVskip{4pt}
        \UnaryInf$\fCenter \limp{\p{\D} {\lam{x}{e}} {\ty_1 \lto \ty_2}}{\sv{}}{\Mat}$
        \DisplayProof}
    \quad
        \scalebox{.8}{
        \def\defaultHypSeparation{\hskip .05in}
        \def\extraVskip{1.5pt}
        \bottomAlignProof
        \Axiom$\fCenter \limp{\p{\D_1} {e_1} {\ty_1 \lto \ty_2}}{\sv1}{\et1}$
        \noLine
        \UnaryInf$\fCenter \limp{ \p{\D_2} {e_2} {\ty_1}}{\sv2}{\et2}$
        \def\extraVskip{4pt}
        \UnaryInf$\fCenter \limp{\p{\D_1 \o \D_2} {e_1e_2} {\ty_2}}{\sv1 \o \sv2}{\et{1}(\et2)_\nb{\ty_2}}$
        \DisplayProof}
    \\[5mm]
        \scalebox{.8}{
        \def\defaultHypSeparation{\hskip .05in}
        \def\extraVskip{1.5pt}
        \bottomAlignProof
        \Axiom$\fCenter \limp{\p{\D_1} {e_1} {\Bool}}{\sv1}{ \begin{bmatrix}\alpha_1 &\!\!\!\alpha_2\end{bmatrix}^\tr }$
        \noLine
        \UnaryInf$\fCenter \limp{\p{\D_2} {e_2} {\ty}}{\sv2}{\et2}$
        \noLine
        \UnaryInf$\fCenter \limp{\p{\D_2} {e_3} {\ty}}{\sv2}{\et3}$
        \def\extraVskip{4pt}
        \UnaryInf$\fCenter \limp{\p{\D_1 \o \D_2} {\ite{e_1}{e_2}{e_3}} {\ty}}{\sv1 \o \sv2}{\alpha_1\.\et{2}+\alpha_2\.\et{3}}$
        \DisplayProof}
    \\[5mm]
    \end{tabular}
    \]
  \end{mdframed}
  \vspace{.8em}
  \caption{Compiling $\cj${\scriptsize $(\nb{\Bool},\nb{\lto})$} programs to linear neurons}
  \label{fig:compiling}
\end{figure}

Compiling variables and constants is straight-forward. Variables map to whatever the environment $\sv{}$ specifies, and the constants $\nb{\tt}$ and $\nb{\ff}$ are injectively mapped to a basis of $\rd{\R^2}$. But functions are more subtle. Consider how we compile \scalebox{1.}{$\p{\emp}{\lam{x}{x}}{\Bool \lto \Bool}$} below. Functions compile to matrices, and the challenge is determining the coefficients of the matrix $\rd{\a{ij}}$. It's similar to determining the matrix of a linear map. We must observe how the body of a function compiles when under an environment that maps $\nb{x}$ to each basis element $\rd{b_j^{\t{in}}} \in \rd{\t{basis}}\c{\nb{\Bool}}=\{\ttv, \ffv\}$. The function $\rd{\t{basis}}\c{\nb{\ty}}$ just generates the \textit{standard}\footnote{See \cite{axler2015linear} for more details on the \textit{standard} basis.} basis for the vector space denoted by a type (see Appendix A.9 for definition). After decomposing the output of the compiler in terms of $\rd{\t{basis}}\c{\nb{\Bool}}$, we can extract the coefficients of the matrix.

\[
        \scalebox{.8}{
        \def\extraVskip{4pt}
        \def\defaultHypSeparation{\hskip .35in}
        \bottomAlignProof
        \AxiomC{}
        \UnaryInfC{$\limp{\p{\bind{x}{\Bool}}{x}{\Bool}}{\{\nb{x}\map
        \ttvm\}}{\ttvm}=\rd{\ul{\textbf{1}} \. \ttvm + \ul{\textbf{0}} \. \ffvm}$}
        \AxiomC{}
        \UnaryInfC{$\limp{\p{\bind{x}{\Bool}}{x}{\Bool}}{\{\nb{x}\map
        \ffvm\}}{\ffvm}=\rd{\ul{\textbf{0}} \. \ttvm + \ul{\textbf{1}} \. \ffvm}$}
        \BinaryInfC{$\limp{\p{\emp}{\lam{x}{x}}{\Bool}}{\emp}{\begin{bmatrix}  
            \textbf{1}  \!&\!  \textbf{0} \\
            \textbf{0}  \!&\!  \textbf{1} \\
        \end{bmatrix}
        }$}
        \DisplayProof}
\]

\vspace{.8em}
Compiling application is simpler. The function compiles to a matrix, the argument compiles to a vector, and the application compiles into a matrix-vector multiplication \cite{axler2015linear}. The subscript on $\et1(\et2)_\nb{\ty}$ determines the vector space of the output, which is needed to ensure matrix multiplication behaves correctly. For example, a $4 \times 2$ matrix might be describing the linear map $\rd{f}: \rd{\R^2 \lto \R^4}$ or $\rd{f} : \rd{\R^2 \lto \R^{2 \times 2}}$. To correctly model application of the linear map, the matrix multiplication must know the intended range. But when it is clear from context, we omit the subscript $\nb{\ty}$ on matrix multiplication (see Appendix A.10 for definition).

The next example demonstrates how applications are compiled for the program \scalebox{1.}{$\p{\emp}{(\lam{x}{x})\tt}{\Bool}$}. To be correct, it must preserve the behavior of the original program; we should be able to predict what compiled neurons will and won't do. To generate our predictions we just run the original program and observe what it does and doesn't evaluate to. Since \scalebox{1.}{$\step{(\lam{x}{x})\tt}{\tt}$} and \scalebox{1.}{$\nb{(\lam{x}{x})\tt} \not\bstep \ff$}, we must check that \scalebox{1.}{$\c{\nb{(\lam{x}{x})\tt}}=\c{\tt}$} and \scalebox{1.}{$\c{\nb{(\lam{x}{x})\tt}}\neq\c{\ff}$}. Both hold, so the example is correct.
\vspace{0.2em}

\[
        \scalebox{.8}{
        \def\extraVskip{4pt}
        \def\defaultHypSeparation{\hskip .35in}
        \bottomAlignProof
        \AxiomC{$\cdots$}
        \UnaryInfC{$\limp{\p{\emp}{\lam{x}{x}}{\Bool \lto \Bool}}{\emp}{\begin{bmatrix}  
            1  \!&\!  0 \\
            0  \!&\!  1 \\
        \end{bmatrix}
        }$}
        \AxiomC{}
        \UnaryInfC{$\limp{\p{\emp}{\tt}{\Bool}}{\emp}{\ttvm}$}
        \BinaryInfC{$\limp{\p{\emp}{(\lam{x}{x})\tt}{\Bool}}{\emp}{\begin{bmatrix}  
            1  \!&\!  0 \\
            0  \!&\!  1 \\
        \end{bmatrix}
        \begin{bmatrix}
        1\\0
        \end{bmatrix}_\nb{\Bool}
        }$}
        \DisplayProof}
\]
\vspace{.3em}

Compiling conditionals is interesting because there appear to be reasonable alternatives. Preserving the behavior of the original program is important, but it turns out a correct compiler to neurons must do more. Let's consider an alternative for compiling conditionals. 

\vspace{0.2em}

\[
        \scalebox{.8}{
        \def\defaultHypSeparation{\hskip .05in}
        \def\extraVskip{1.5pt}
        \bottomAlignProof
        \Axiom$\fCenter \limp{\p{\D_1} {e_1} {\Bool}}{\sv1}{ \begin{bmatrix}\alpha_1 &\!\!\!\alpha_2\end{bmatrix}^\tr }$
        \noLine
        \UnaryInf$\fCenter \limp{\p{\D_2} {e_2} {\ty}}{\sv2}{\et2}$
        \noLine
        \UnaryInf$\fCenter \limp{\p{\D_2} {e_3} {\ty}}{\sv2}{\et3}$
        \def\extraVskip{4pt}
        \LeftLabel{\textcolor{red}{\huge \xmark}\hspace{.5em}}
        \UnaryInf$\fCenter \limp{\p{\D_1 \o \D_2} {\ite{e_1}{e_2}{e_3}} {\ty}}{\sv1 \o \sv2}{
        \begin{cases} 
        \et2  & \t{if } \alpha_1 = 1 \\ 
        \et3  & \t{otherwise}
        \end{cases}}$
        \DisplayProof}
\]
\vspace{.3em}

Essentially all programming languages for defining neural networks, if they support conditionals, implement a rule like this---a hard branching conditional \cite{jax2018github, paszke2017automatic}. If $\nb{e_1}$ evaluates to $\tt$, you compile the true branch $\nb{e_2}$, otherwise you compile the false branch $\nb{e_3}$. In a sense it's not incorrect, it preserves the behavior of programs. For example, \scalebox{1.}{$\step{\ite{\ul{\tt}}{\ul{\tt}}{\ul{\tt}}}{\tt}$} and \scalebox{1.}{$\nb{\ite{\ul{\tt}}{\ul{\tt}}{\ul{\tt}}} \not\bstep \nb{\ff}$}, and using the hard branching conditional ensures \scalebox{1.}{$\c{\nb{\ite{\ul{\tt}}{\ul{\tt}}{\ul{\tt}}}}=\c{\tt}$} and \scalebox{1.}{$\c{\nb{\ite{\ul{\tt}}{\ul{\tt}}{\ul{\tt}}}}\neq\c{\ff}$}. This is useful if, once a neural network is trained, you want to switch between using different neural networks depending on some condition. However, before training, a hard branch breaks differentiation. As a result, you can't jointly train a neural network that produces the condition \scalebox{1.}{$\rd{\bv{\a1}{\a2}}$} and the outcomes of each branch $\rd{\et2}$ and $\rd{\et3}$; it can't be used to solve the conditional transform task with the data from Fig. \ref{fig:task}. The central issue is that the output, outside the boundary condition \scalebox{1.}{$\rd{\a1=1}$}, does not depend on \scalebox{1.}{$\rd{\bv{\a1}{\a2}}$}. Therefore the gradient with respect to \scalebox{1.}{$\rd{\bv{\a1}{\a2}}$} is zero almost everywhere.  As a result, gradient-based learning techniques can't train a neural network that determines \scalebox{1.}{$\rd{\bv{\a1}{\a2}}$}. This is why we opt for the following soft branching conditional.

\vspace{0.2em}

\[
        \scalebox{.8}{
        \def\extraVskip{4pt}
        \def\defaultHypSeparation{\hskip .35in}
        \bottomAlignProof
        \AxiomC{}
        \UnaryInfC{$\limp{\p{\bind{x}{\Bool}}{x}{\Bool}}{\{\nb{x} \map \vt{}\}}{\begin{bmatrix}\a1 \\ \a2\end{bmatrix}}$}
        \AxiomC{}
        \UnaryInfC{$\limp{\p{\emp}{\tt}{\Bool}}{\emp}{\ttvm}$}
        \AxiomC{}
        \UnaryInfC{$\limp{\p{\emp}{\ff}{\Bool}}{\emp}{\ffvm}$}
        \LeftLabel{\textcolor{ForestGreen}{\huge \cmark}\hspace{.5em}}
        \TrinaryInfC{$\limp{\p{\bind{x}{\Bool}}{\ite{\underline{x}}{\underline{\tt}}{\underline{\tt}}}{\Bool}}{\{\nb{x} \map \vt{}\}}{\alpha_1\.\ttvm+ \alpha_2 \. \ffvm}$}
        \DisplayProof}
\]
\vspace{.8em}

The soft branching conditional weights the output according to the compiled condition \scalebox{1.}{$\rd{\bv{\a1}{\a2}}$}. It's correct in the sense that it preserves the behavior of programs. But it also won't break learning. We use it to train a neural network like Fig. \ref{fig:direct-circuit} to solve the conditional transform task from Fig. \ref{fig:task} (Section \ref{sec:experiments}).

Our last example before discussing the compiler correctness theorem will be higher-order: a linearly typed variant of a Church encoded \scalebox{1.}{$\ff$}. These Church encodings provide ways of representing discrete data using higher-order functions \cite{pierce2002types}. For our compiler,  they allow us to embed discrete data (in this case, booleans) into vector spaces of varying dimension, and in non-obvious ways. They also compile to linear \textit{hypernetworks} (Section \ref{sec:lin-neuron}) because its matrix denotes a linear map $\rd{f} : \rd{\R^2 \lto \R^{2 \times 2}}$. To simplify things, we present the derivation in two parts. First, we compile a subprogram \scalebox{1.}{$\p{\bind{x}{\Bool}}{\lam{y}{\ite{x}{y}{y}}}{\Bool\lto\Bool}$}, and then the full encoding \scalebox{1.}{$\p{\emp}{\lam{x}{\lam{y}{\ite{\underline{y}}{\underline{x}}{\underline{x}}}}}{\Bool\lto\Bool\lto\Bool}$}.

\vspace{0.2em}

\[
        \scalebox{.8}{
        \def\extraVskip{4pt}
        \def\defaultHypSeparation{\hskip .35in}
        \bottomAlignProof
        \AxiomC{}
        \UnaryInfC{$\limp{\p{\bind{x}{\Bool}}{x}{\Bool}}{\{\nb{x}\map\vt{}\}}{\bvm{\alpha_1}{\alpha_2}}$}
        \AxiomC{}
        \UnaryInfC{$\limp{\p{\bind{y}{\Bool}}{y}{\Bool}}{\{\nb{y}\map\ttvm\}}{\ttvm}$}
        \AxiomC{}
        \UnaryInfC{$\limp{\p{\bind{y}{\Bool}}{y}{\Bool}}{\{\nb{y}\map\ttvm\}}{\ttvm}$}
        \TrinaryInfC{$\limp{\p{\bind{x}{\Bool},\bind{y}{\Bool}}{\ite{x}{y}{y}}{\Bool}}{\{\nb{x}\map
\vt{}, \nb{y}\map\ttvm\}}{\alpha_1\.\ttvm+\alpha_2\.\ttvm}=\rd{(\alpha_1+\alpha_2)\.\ttvm+0\.\ffvm}$}
        \DisplayProof}
\]

\[
        \scalebox{.8}{
        \def\extraVskip{4pt}
        \def\defaultHypSeparation{\hskip .35in}
        \bottomAlignProof
        \AxiomC{}
        \UnaryInfC{$\limp{\p{\bind{x}{\Bool}}{x}{\Bool}}{\{\nb{x}\map\vt{}\}}{\bvm{\alpha_1}{\alpha_2}}$}
        \AxiomC{}
        \UnaryInfC{$\limp{\p{\bind{y}{\Bool}}{y}{\Bool}}{\{\nb{y}\map\ffvm\}}{\ffvm}$}
        \AxiomC{}
        \UnaryInfC{$\limp{\p{\bind{y}{\Bool}}{y}{\Bool}}{\{\nb{y}\map\ffvm\}}{\ffvm}$}
        \TrinaryInfC{$\limp{\p{\bind{x}{\Bool},\bind{y}{\Bool}}{\ite{x}{y}{y}}{\Bool}}{\{\nb{x}\map
\vt{}, \nb{y}\map\ffvm\}}{\alpha_1\.\ffvm+\alpha_2\.\ffvm}=\rd{0\.\ttvm+(\alpha_1+\alpha_2)\.\ffvm}$}
        \DisplayProof}
\]
\vspace{.8em}
\[
        \scalebox{.8}{
        \def\extraVskip{4pt}
        \def\defaultHypSeparation{\hskip .35in}
        \bottomAlignProof
        \AxiomC{}
        \LeftLabel{($\nb{\t{id}_{x:\Bool}}$)\hspace{.5em}}
        \UnaryInfC{$\limp{\p{\bind{x}{\Bool}}{\lam{y}{\ite{x}{y}{y}}}{\Bool\lto\Bool}}{\{\nb{x} \map \vt{}\}}{\begin{bmatrix}  
(\alpha_1+\alpha_2)  \!&\!  0 \\
0  \!&\!  (\alpha_1+\alpha_2) \\
\end{bmatrix}}$}
        \DisplayProof}
\]
\vspace{.8em}

Compiling the subprogram $(\nb{\t{id}_{x:\Bool}})$ combines the function and conditional rules. It first uses the function rule to observe how the compiler acts on the body of the function, extending the environment by binding $\nb{y}$ to each basis element \scalebox{1.}{$\rd{b_j^{\t{in}}} \in \rd{\t{basis}}\c{\nb{\Bool}}=\{\ttv, \ffv\}$}. Under each environment the compiler then uses the conditional rule to soft branch on the compiled condition \scalebox{1.}{$\rd{\bv{\a1}{\a2}}$}. The output of the compiler is then rewritten to easily extract the coefficients of the matrix. With the subprogram $(\nb{\t{id}_{x:\Bool}})$  compiled, we can more easily compile the Church encoded \scalebox{1.}{\ff}.

\vspace{.8em}
\[
        \scalebox{.8}{
        \def\extraVskip{4pt}
        \def\defaultHypSeparation{\hskip .35in}
        \bottomAlignProof
        \AxiomC{}
        \LeftLabel{($\nb{\t{id}_{x:\Bool}}$)\hspace{.5em}}
        \UnaryInfC{$\limp{\p{\bind{x}{\Bool}}{\lam{y}{\ite{x}{y}{y}}}{\Bool\lto\Bool}}{\{\nb{x} \map \ttvm\}}{\begin{bmatrix}  
1  \!&\!  0 \\
0  \!&\!  1 \\
\end{bmatrix} =\rd{
1 \. 
\begin{bmatrix}
1  \!&\!  0 \\
0  \!&\!  0 \\
\end{bmatrix}
+
0 \.
\begin{bmatrix}
0  \!&\!  0 \\
1  \!&\!  0 \\
\end{bmatrix}
+
0 \.
\begin{bmatrix}
0  \!&\!  1 \\
0  \!&\!  0 \\
\end{bmatrix}
+
1 \.
\begin{bmatrix}
0  \!&\!  0 \\
0  \!&\!  1 \\
\end{bmatrix}
}}$}
        \DisplayProof}
\]
\vspace{.8em}
\[
        \scalebox{.8}{
        \def\extraVskip{4pt}
        \def\defaultHypSeparation{\hskip .35in}
        \bottomAlignProof
        \AxiomC{}
        \LeftLabel{($\nb{\t{id}_{x:\Bool}}$)\hspace{.5em}}
        \UnaryInfC{$\limp{\p{\bind{x}{\Bool}}{\lam{y}{\ite{x}{y}{y}}}{\Bool\lto\Bool}}{\{\nb{x} \map \ffvm\}}{\begin{bmatrix}  
1  \!&\!  0 \\
0  \!&\!  1 \\
\end{bmatrix}} =\rd{
1 \. 
\begin{bmatrix}
1  \!&\!  0 \\
0  \!&\!  0 \\
\end{bmatrix}
+
0 \.
\begin{bmatrix}
0  \!&\!  0 \\
1  \!&\!  0 \\
\end{bmatrix}
+
0 \.
\begin{bmatrix}
0  \!&\!  1 \\
0  \!&\!  0 \\
\end{bmatrix}
+
1 \.
\begin{bmatrix}
0  \!&\!  0 \\
0  \!&\!  1 \\
\end{bmatrix}
}$}
        \DisplayProof}
\]
\vspace{.8em}
\[
        \scalebox{.8}{
        \def\extraVskip{4pt}
        \def\defaultHypSeparation{\hskip .35in}
        \bottomAlignProof
        \AxiomC{}
        \UnaryInfC{$\limp{\p{\emp}{\lam{x}{\lam{y}{\ite{\underline{x}}{\underline{y}}{\underline{y}}}}}{\Bool\lto\Bool\lto\Bool}}{\emp}{\begin{bmatrix}
1  \!&\!  1 \\
0  \!&\!  0 \\
0  \!&\!  0 \\
1  \!&\!  1 \\ 
\end{bmatrix}}$}
        \DisplayProof}
\]
\vspace{.8em}

\subsubsection{Proving the compiler is correct} For closed programs, our intuition for what makes a compiler correct is based on Fig. \ref{fig:ccc1}. If the compiler permits predictions about what compiled neurons will and won't do, we say it is correct. These intuitions are captured in the following theorem.

\begin{bthm}{(Compiler preserves program behavior)}{}
\vspace{.25em}
\t{If }$\limp{\p{\emp}{e}{\Bool}}{\emp}{\et{}}$ \t{ and } $\limp{\p{\emp}{v}{\Bool}}{\emp}{\vt{}}$, \t{ then }
\vspace{.5em}
\begin{flalign*}
&\begin{aligned}
(a) \; & \step{e}{v}  \implies \et{}=\vt{} \\
(b) \; & \nb{e} \;\cancel{\bstep}\; \nb{v} \implies \et{}\neq\vt{}\\
\end{aligned}&&
\end{flalign*}    
\end{bthm}
\vspace{.3em}

\noindent\ul{\textsf{Outline of Proof.}}

To preserve \textit{closed} program behavior, this theorem requires that we can predict what compiled neurons $\rd{\et{}}$ will and won't do. These predictions come from what the original program $\nb{e}$ will and won't evaluate to.  The bulk of the difficulty is proving what compiled neurons $\rd{\et{}}$ will do, see $(a)$. The argument is by induction on evaluation $\step{e}{v}$ for the following lemma, which generalizes the type of programs from $\nb{\Bool}$ to $\nb{\ty}$.\footnote{Note that the compiler correctness theorem is \textit{not} true for function types because of condition $(b)$, $\nb{\lam{x}{x}} \not\bstep \nb{\lam{y}{}y}$ but both compile to the identity matrix.}

\vspace{.3em}
\begin{rlem}{(Compiler preserves what programs will do)}{}
\vspace{.25em}
\t{If $\limp{\p{\emp}{e}{\ty}}{\emp}{\et{}}$ and $\limp{\p{\emp}{v}{\ty}}{\emp}{\vt{}}$,}
\vspace{.45em}
\[
\step{e}{v} \implies \et{}=\vt{} \\
\] 
\end{rlem}
\vspace{.3em}
The argument is by induction on evaluation \scalebox{1.}{$\step{e}{v}$}. The value case $\step{v}{v}$ is immediate. The conditional case $\step{\ite{e_1}{e_2}{e_3}}{v}$ follows immediately after using the induction hypotheses. However, the application case is subtle \scalebox{1.}{$\step{e_1e_2}{v}$}. The argument requires a key lemma.

\vspace{.5em}
\begin{rlem}{(Compiler maps programs to linear maps)}{}
\vspace{.25em}
\t{If }$\limp{\p{\D}{e}{\ty}}{\sv{}\{\nb{x}\map\vt{1}\}}{\et{1}}$ \t{ and } $\limp{\p{\D}{e}{\ty}}{\sv{}\{\nb{x}\map\vt{2}\}}{\et{2}}$, \t{ then }
\vspace{.45em}
\[
\limp{\p{\D}{e}{\ty}}{\sv{}\{\nb{x} \map\alpha_1\.\vt1+\alpha_2\.\vt2\}}{\alpha_1\.\et1+\alpha_2\.\et{2}} \\
\] 
\end{rlem}
\vspace{.7em}

Previously, we discussed the compiler as a function that maps a program \scalebox{1.}{$\p{\D}{e}{\ty}$} and an environment \scalebox{1.}{$\sv{}$} to some neurons \scalebox{1.}{$\et{}$}. If we partially apply the compiler only to a program, then the compiler maps programs to \textit{linear} maps between the environment $\rd{\sv{}}$ and neurons $\et{}$. That's what this lemma ensures. As a consequence, if we observe how the compiler acts under the environments \scalebox{1.}{$\rd{\sv{}\{\nb{x} \map \vt1\}}$} and \scalebox{1.}{$\rd{\sv{}\{\nb{x} \map \vt2\}}$}, then we know how the compiler acts on combinations of these environments. This relieves a core difficulty in proving that the compiler preserves the behavior of \scalebox{1.}{$\step{e_1e_2}{v}$}. By induction it is easy to conclude that the compiler preserves the behavior of \scalebox{1.}{$\step{e_1}{\lam{x}{e}}$} and \scalebox{1.}{$\step{e_2}{v_2}$}. However, it is \textit{not} easy
to use induction to conclude the compiler preserves the behavior of \scalebox{1.}{$\step{\{x \map v_2\}(e)}{v}$}, which is the \textit{only} way we can determine anything about the compiled neurons \scalebox{1.}{$\rd{\vt{}}$}. This lemma is necessary for the induction to go through. Beyond the proof, it establishes a correspondence between linear types and linear maps. It is the most important lemma in this paper. Detailed proofs of the compiler correctness theorem and these lemmas are in Appendix B.1, B.3, and B.9.

A consequence of the compiler correctness theorem is that we can use \cj\ to directly program linear neurons. However, our original goal also requires us to link these neurons against potentially nonlinear neurons. This would allow us to directly program a neural network to solve the conditional transform task from Fig. \ref{fig:task}.

\subsection{Linking}
\label{sec:linking}
There are two ways to link linear neurons against nonlinear neurons.
\setlist[itemize]{topsep=4pt, itemsep=4pt, parsep=0pt, leftmargin=1.5em, rightmargin=1.5em}
\begin{itemize}[label=$\triangleright$]
    \item Linear neurons can receive inputs from nonlinear neurons\footnote{These inputs can be specified in the environment when compiling.}
    \item Linear neurons can send outputs to nonlinear neurons\footnote{Since our programs compile to mathematical functions, feeding their output to another network is function composition.}
\end{itemize}

Traditionally, a correct compiler that supports linking allows us to predict the behavior of linked programs, captured by compositional compiler correctness theorems \cite{patterson2019next}. Consider the following compositional compiler correctness theorem, whose proof is detailed in Appendix B.2.

\vspace{.2em}
\begin{bthm}{(Compiler preserves open program behavior)}{}
\vspace{.25em}
\t{If $\limp{\p{\D}{e}{\Bool}}{\sv{}}{\et{}}$, $\limp{\p{\emp}{v}{\Bool}}{\emp}{\vt{}}$ and $\c{\nb{\D} \vdash \nb{\sigma}} = \sv{}$, then
\vspace{.5em}
\begin{flalign*}
&\begin{aligned}
(a) \; & \step{\sigma(e)}{v}  \implies \et{}=\vt{} \\
(b) \; & \nb{\sigma(e)} \;\cancel{\bstep}\; \nb{v} \implies \et{}\neq\vt{}\\
\end{aligned}&&
\end{flalign*} }   
\end{bthm}
\vspace{.3em}

This theorem says we can predict what compiled neurons will and won't do, when linked against environments \textit{produced by our compiler}, $\c{\nb{\D} \vdash \nb{\sigma}}=\rd{\vec{\sigma}}$. This provides some assurance. However, there are many environments we may want to link against which aren't produced by our compiler. For example, nonlinear neurons may provide $\bv{1.3}{4.7}$ as input to linear neurons, which isn't an environment covered by the theorem. It could be possible to expand the environments we consider by weakening the exactness of our predictions. However, there is a more pressing concern.

Are these even the relevant predictions for our motivating example in Fig. \ref{fig:task}? We want to directly program a neural network to perform a conditional image transform. To directly program a neural network, \cj\ must compile to linear neurons that specify the conditional while linking against neurons that can convert images of any number into an image of a one. The way we use the compiler should determine how we justify its correctness, which we summarize in the following way.

\vspace{.4em}
\begin{quote}
\begin{center}
\textit{The compiler is correct if the neurons effectively learn the task.}
\end{center}
\end{quote}
\vspace{.4em}

\noindent Therefore, a practical theory of compiling to neurons must involve predictions about learning. But no one knows how to \textit{exactly} predict the dynamics of learning in nonlinear neural networks \cite{saxe2015deep}. Their dynamics are governed by nonlinear differential equations that are difficult to solve.\footnote{Resolving this is a central problem in the theory of neural networks \cite{roberts2022principles}.} This imposes an important limit on the predictions we can expect to make in a compiler correctness theorem. As is common in the study of nonlinear dynamical systems, \textit{experiments} now play an essential role in addressing the correctness of our compiler. Experiments help us characterize the learning dynamics of compiled neurons, to address whether the compiler correctness theorems we prove yield any benefit to learning tasks like those in Fig. \ref{fig:task}. In the following section, we develop this empirical theory of linking.

\section{Experiments}
\label{sec:experiments}

Our experiments are designed to test whether we see any benefits to learning when directly programming neural networks with \cj\ (Section \ref{sec:exp1}), and whether the compiler correctness theorem we proved is actually driving this benefit (Section \ref{sec:exp2}). We are interested in the relationship between the strength of a compiler correctness theorem and its impact on learning dynamics when linking against nonlinear neurons. There is also a brief exploration on how directly programming neural networks facilitates debugging their learning dynamics (Section \ref{sec:debugging}). 

To briefly summarize our findings, we find that directly programming neural networks helps neural networks learn faster, and that the strength of this impact relates to the strength of the compiler correctness theorem satisfied by the compiler producing the neural network.

\subsection{Experiment 1: Conditional image transformation}
\label{sec:exp1}

\subsubsection{Task}

Our first experiment is the conditional transform task in Fig. \ref{fig:task}. We need a function which conditionally transforms an input image $x$:
\setlist[itemize]{topsep=4pt, itemsep=1pt, parsep=2pt, leftmargin=1.5em, rightmargin=1.5em}
\begin{itemize}[label=$\triangleright$]
\item If $x$ is a 1, return $x$
\item If $x$ isn't a 1, transform $x$ to a 1
\end{itemize}


\subsubsection{Models}

The two models we test are the indirectly and directly programmed neural networks in Fig. \ref{fig:circuits}. Both models are implemented in PyTorch, a domain-specific language in Python for building neural networks. The indirect model uses native PyTorch whereas the direct model uses our implementation of \cj: an embedded domain-specific language within PyTorch. It can be helpful to think of each model as a parameterized differentiable function $f_\theta(x)$, whose parameters $\theta$ can be learned through gradient-based learning techniques to perform a computation described by a dataset of input-output examples.

The indirect model (Fig. \ref{fig:indirect-circuit}) is a 3-layer neural network. Its input layer has 784 neurons, mirroring the number of pixels in an input image. These connect to a hidden layer with 400 neurons and a $\t{relu}$ nonlinearity. These hidden layer neurons connect to an output layer of 784 neurons, mirroring the number of pixels in an output image. The strength of the connection between neurons (their weights) determines the network's overall function. In total, there are $627,200$ weights to learn.

The direct model (Fig. \ref{fig:direct-circuit}) is three 3-layer neural networks in parallel, which are wired together according to the conditional compiling rule in Fig. \ref{fig:compiling}. Its input layer also has 784 neurons. These then split off into three paths, each similar to but smaller than the indirect model. The middle path is a 3-layer neural network with 100 hidden neurons and 2 output neurons, meant to check whether an input image is a one or not. The top path is a 2-layer linear neural network with 784 input and output neurons. Its weights are determined prior to learning to compute identity and do not change during learning. This path is meant to return the original image if the middle path determines that the input image is a one. Finally, the bottom path is a 3-layer neural network with 300 hidden neurons and 784 output neurons, meant to transform the input image into an image of a one if the middle path determines that the input image is not a one. In short, the middle path checks the condition, the top path returns the original image, and the bottom path transforms the image. The end of the network just wires them together, either suppressing or enhancing the outputs of the top or bottom path depending on whether the input is classified as a one. Similar to the indirect model, there are $627,200$ weights to learn.

\begin{figure}[t]
  \includegraphics[scale=.44]{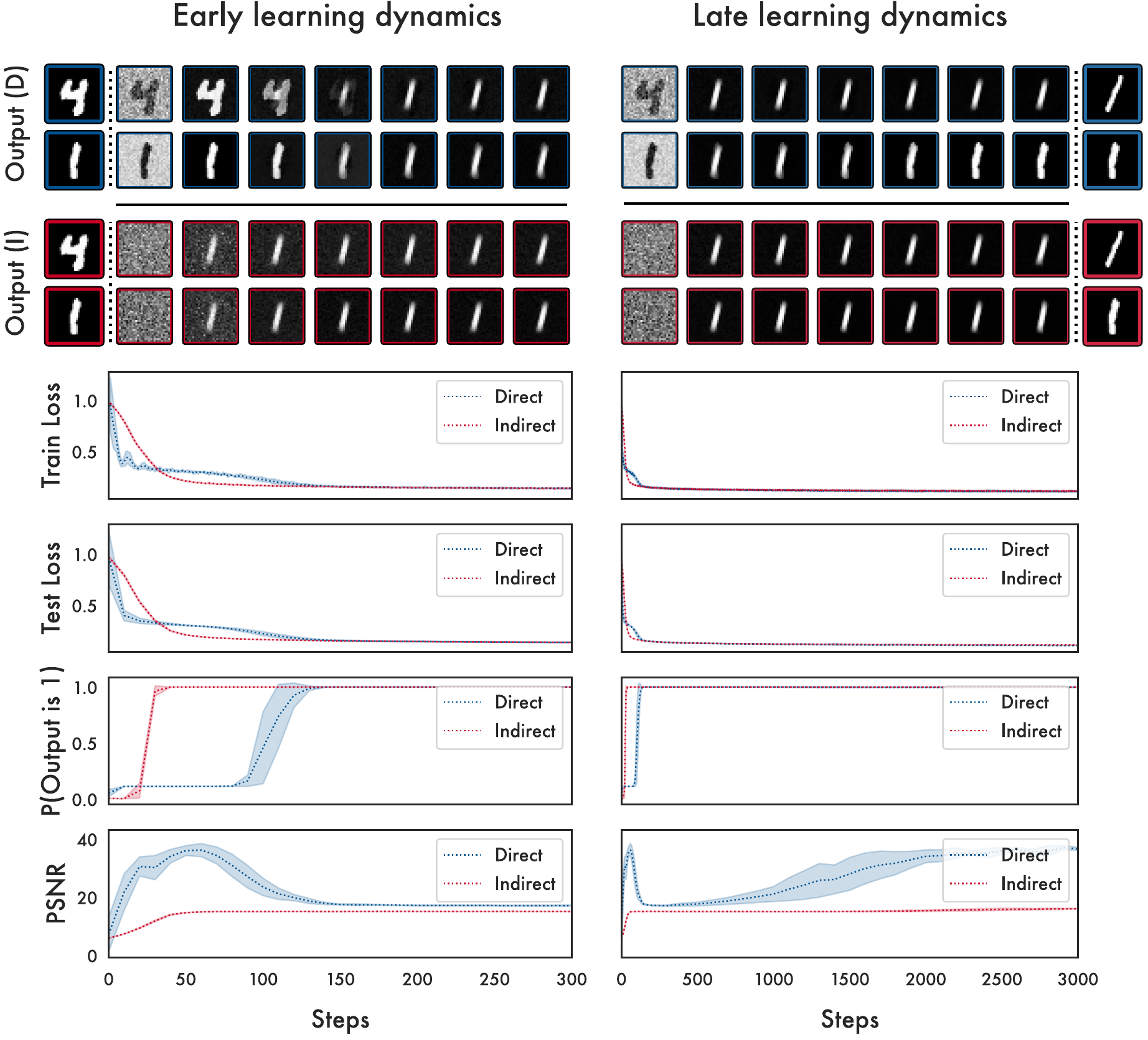}
  \centering

  \vspace{1.6em}
  \caption{Prototypical learning dynamics}
  \label{fig:local-dynamics}
\end{figure}

\subsubsection{Measures}

Before discussing the results, we briefly explain the different measures we use to compare learning dynamics for the indirectly and directly programmed neural networks.

The first measure of learning dynamics is a neural network's loss on the training and test dataset. Neural networks learn by iterative adjustments of their weights in order to minimize a loss function. These loss functions measure how well the network can approximate the input-output transformations specified in the training and test data. The loss is an important baseline measure. However, two networks can exhibit distinct behaviors while their training and test loss converge. Therefore, we need additional measures of learning dynamics that allow us to distinguish neural networks which behave differently.

The second measure of learning dynamics is the probability that the neural network's output is classified as a one. Before training, these networks behave wildly, and their outputs are seemingly random arrays of pixels. This measure captures when the network has learned a key part of the task---that it must produce images of a one. The probability is determined by a separate neural network trained to classify digits with near-perfect accuracy.\footnote{It is around 99\% accurate on the MNIST classification task \cite{lecun2010mnist}.}

The third measure of learning dynamics is the peak signal-to-noise ratio (PSNR) between the input and output image. It is commonly used as measure of image reconstruction error. This measure captures when the network has learned the hardest part of the task---that it must return the original image when the input image is of a one. The PSNR measures how close the output image is to the original.

\begin{figure}[t]
  \includegraphics[scale=.4]{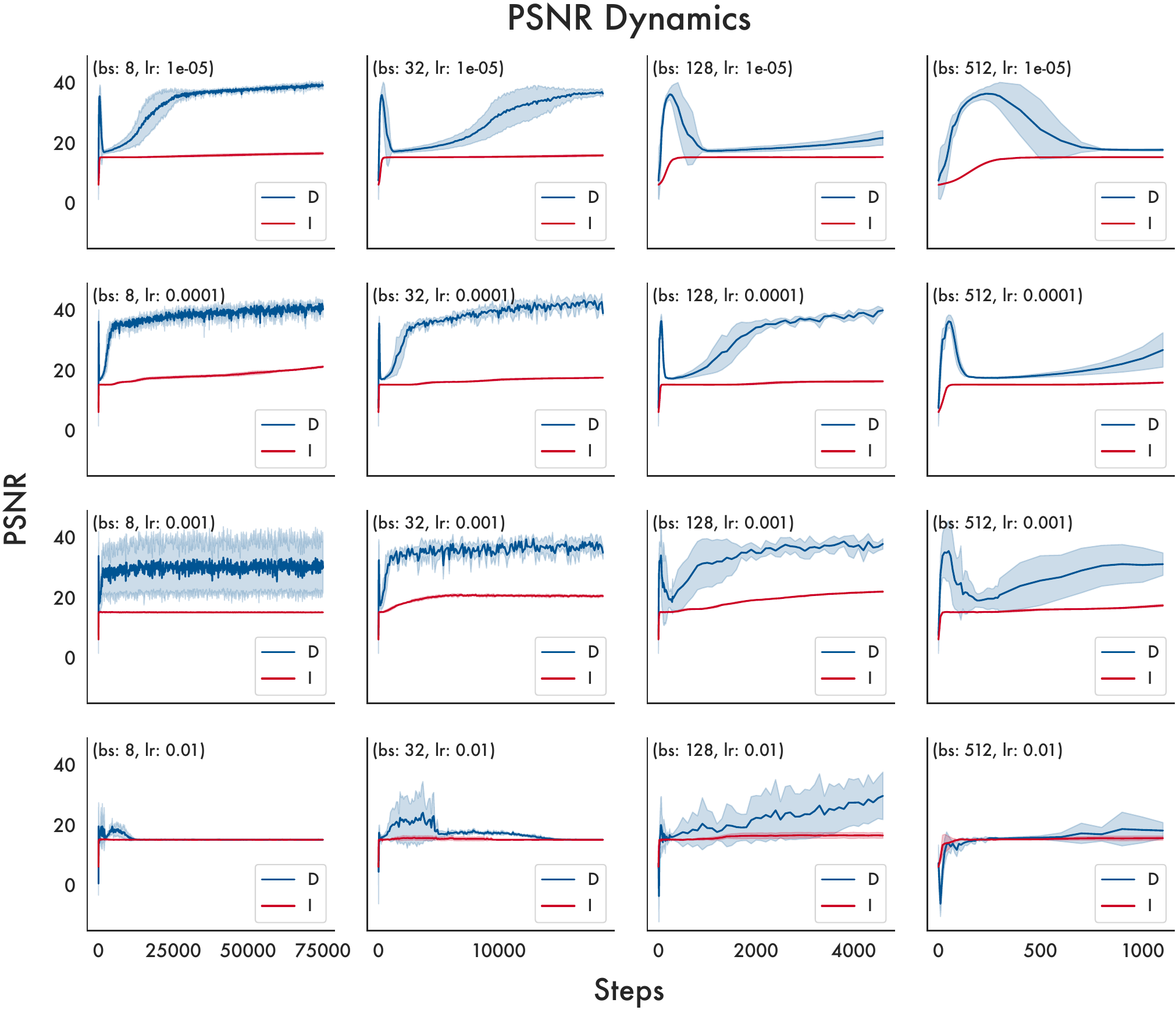}
  \centering
  \vspace{1.6em}
    
  \caption{Summary of dynamics across configurations}
  \vspace{.0em}
  \label{fig:global-dynamics}
\end{figure}

\subsubsection{Results}

To collect our data, we trained both models to approximate the function described by a training dataset of 60,000 pairs of images, and tested both models on a distinct dataset of 10,000 pairs of images. Both models used the Adam optimizer to compute weight updates to the neural network, using an L1 loss to measure model performance during training. Throughout training we recorded the previously described measures in order to understand the model's learning dynamics. We did this for 5 random seeds, 4 learning rates, and 4 training batch sizes---all can (and did) vary the learning dynamics.

Fig. \ref{fig:local-dynamics} summarizes the prototypical learning dynamics across training configurations. On the $x$ axis, each step indicates a weight update to the models. The early dynamics are the first 300 updates; the late dynamics are the first 3000 updates. On the $y$ axis are the measures previously discussed. Additionally, at the top are model outputs on two input images, for both the directly programmed neural network $(\t{D})$ and the indirectly programmed neural network $(\t{I})$. The first image on each row is the input to the model. The last image on each row indicates what the correct output should look like.\footnote{According to the test dataset.} The measures shown are across 5 random seeds with a learning rate of 0.0001 and training batch size of $128$ samples. The shaded ribbons denote the standard deviation across random seeds.

The indirect model struggles to conditionally transform the input image, instead learning to produce a constant image of a one on all inputs. However, the direct model is able to conditionally transform the input image. If the input is a one, by the end of training it returns a near identical copy. If the input isn't a one, it produces a constant image of a one like the indirect model. The indirect model converges quickly to its solution and stays there throughout the early and late learning dynamics. The direct model instead has a more interesting trajectory. In the first 100 steps, the direct model overuses the identity path in the network on all inputs. It then overcorrects and doesn't use the identity path on any input after 150 steps. For the remainder of training it slowly learns to reintegrate the identity path on inputs that are ones. To verify, observe the intermediate model outputs and PSNR dynamics.

Fig. \ref{fig:global-dynamics} depicts the PSNR learning dynamics across every training configuration. There are 16 subplots, each indicating a unique training configuration. The legend indicates the training configuration: across 5 random seeds we vary learning rate $(\t{lr})$ and training batch size $(\t{bs})$ for both the direct $(\t{D})$ and indirect $(\t{I})$ model. The shaded ribbons denote the standard deviation across random seeds. Only PSNR is shown because it was the measure that best distinguished model performance. On the $x$ axis, each step indicates a weight update to the models. The $x$ axis differs per column because we constrained each configuration to train for 10 iterations through the training dataset---a fixed data budget across configurations. This results in more weight updates for smaller batch sizes. On the $y$ axis is PSNR. Values between 30 and 40 indicate near perfect reconstruction when the input image is of a one. Values near 20 indicate the configuration learned to produce a constant one across any input. 

We can use Fig. \ref{fig:global-dynamics} to determine whether the direct model learns faster and with greater data-efficiency. To verify that the direct model learns faster, we must compare the best performing direct and indirect model across configurations with fixed training steps (within columns). The direct model consistently outperforms.  To verify that the direct model learns more efficiently, we must compare the best performing direct and indirect model across configurations with fixed data budget (all configurations). The direct model consistently outperforms. This does not mean there can't exist an indirect model that solves this task. There likely is. But our indirect model is reasonable, and yet is having trouble.\footnote{Similar indirect models can be trained to solve other tasks (Appendix C). Why then does it struggle with this problem?}

\subsubsection{Debugging}
\label{sec:debugging}

\begin{figure}[h]
  \includegraphics[scale=.43]{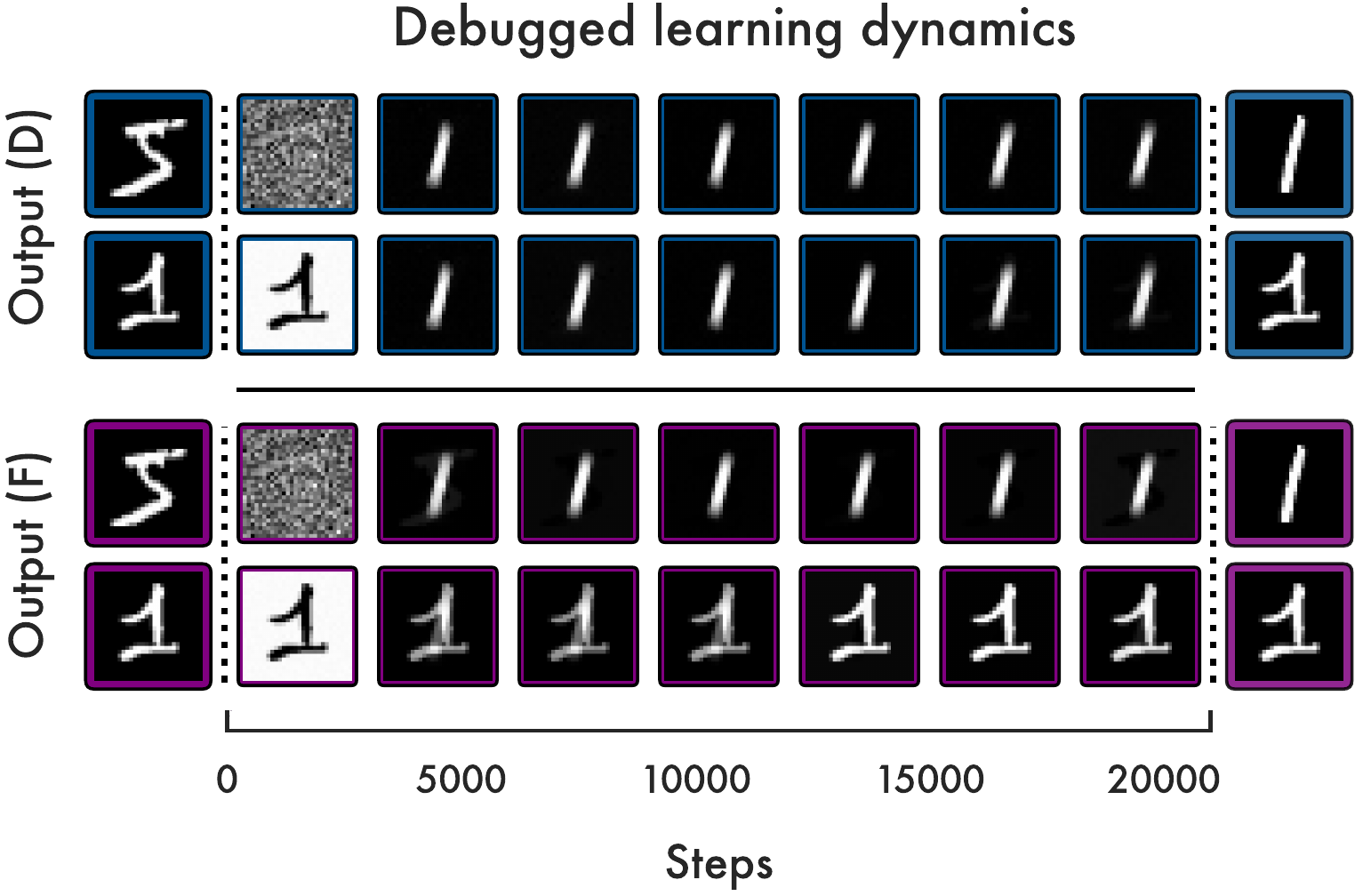}
  \centering
  \vspace{1em}
  \caption{Augmenting learning to debug learning dynamics}
  \label{fig:debugged}
\end{figure}

Beyond learning speed and efficiency, directly programming neural networks allows you to debug neural networks in new ways. For example, despite the direct model performing well on the task, there appear to be challenging examples that the direct model gets wrong. For example, in Fig. \ref{fig:debugged} the direct model $(\t{D})$ behaves well when the input is a five; it returns a constant one. But it behaves poorly when the input is a one; it doesn't return the original image. However, an important observation is that this particular image of a one could be mistaken for a two. Therefore, we were curious if the error was due to the middle path in Fig. \ref{fig:direct-circuit} misclassifying the input as a two. When we probed the output of the middle path we found the direct model was treating the problem input as if it wasn't a one. Moreover, if we perturbed the output of the middle path to make it consistent with how other images of ones were treated, we could recover the correct behavior---it would return the original image. Based on this diagnosis we augmented training with an added penalty for misclassifying inputs at the middle path. In the bottom-half of Fig. \ref{fig:debugged} we show the augmented learning dynamics with this penalty. As a result, we were able to correct the bug on this example. This debugging strategy was possible because we knew the neural network was directly programmed using a program like Fig. \ref{fig:direct-program}.
\newpage
\subsection{Experiment 2: Conditional image classification}
\label{sec:exp2}

\subsubsection{Task}

\begin{wrapfigure}{h}{0.4\textwidth}
    \centering
    \vspace{-1.4em}
    \includegraphics[width=.38\textwidth]{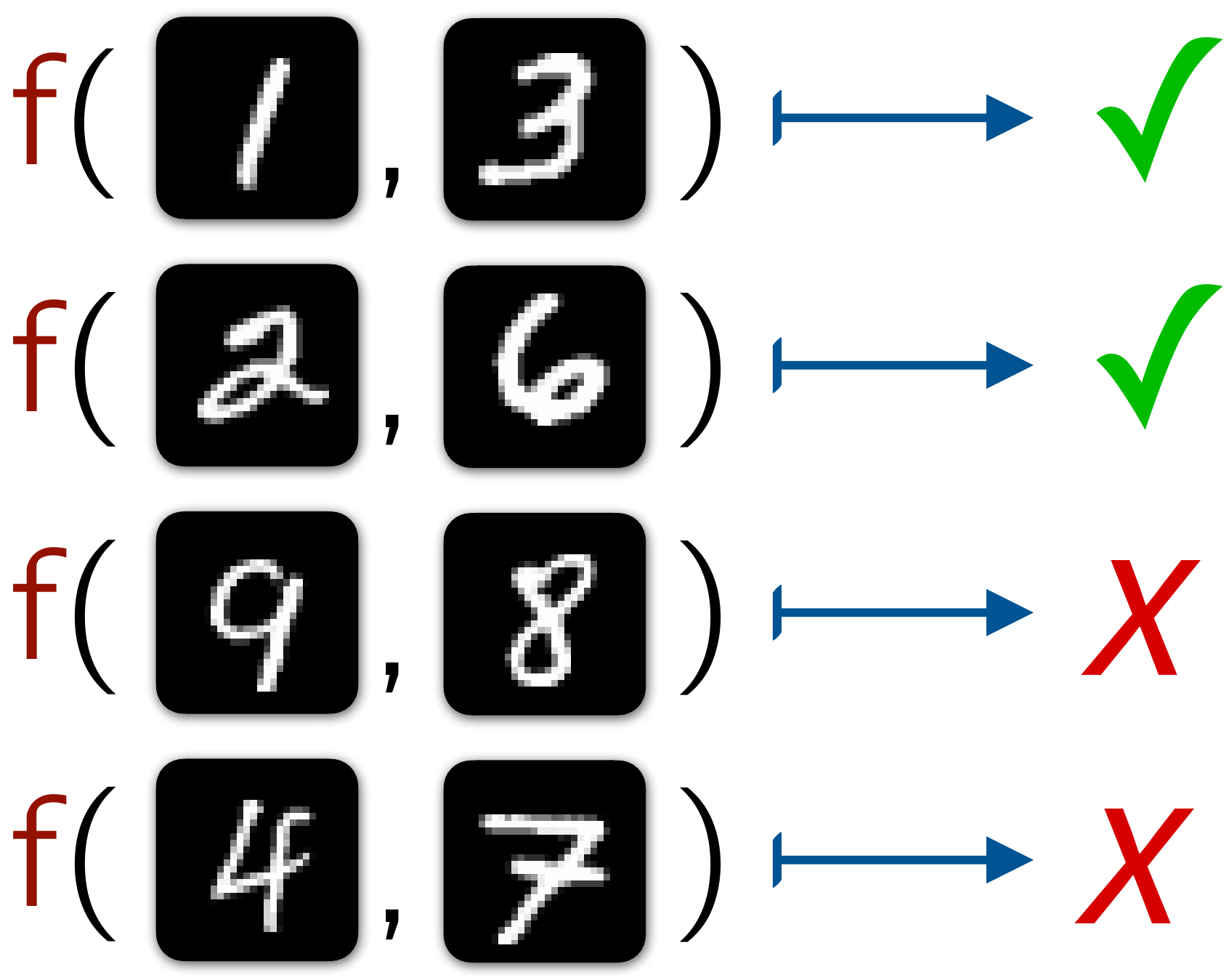}
    \vspace{.6em}
  \caption{Conditional classification}
  \label{fig:task2}
  \vspace{-2.5em}
\end{wrapfigure}

Our second experiment is the conditional classification task in Fig. \ref{fig:task2}. We need a function which conditionally classifies two input image $x_1$ and $x_2$:
\setlist[itemize]{topsep=4pt, itemsep=1pt, parsep=2pt, leftmargin=1.5em, rightmargin=1.5em}
\begin{itemize}[label=$\triangleright$]
\item If $x_1$ and $x_2$ are even numbers, return \textcolor{ForestGreen}{\cmark}
\item If $x_1$ and $x_2$ are odd numbers, return \textcolor{ForestGreen}{\cmark}
\item Otherwise, return \textcolor{red}{\xmark}
\end{itemize}

\subsubsection{Models}

We test four models. Like before, there is an indirect and direct model. But there is now a model which explores directly programming with Church encodings. These encodings compile to linear hypernetworks. An additional model explores directly programming with a type-preserving compiler \textit{refined} (keeping type translation) from our correct compiler. This model lets us test whether preserving behavior provides any benefit to learning.

Importantly, these models can be organized on a spectrum based on the strength of the compiler correctness theorem they satisfy.
$$\t{I} < \t{T} < \t{D}=\t{C}$$
Models $\t{D}$ and $\t{C}$ are on the strongest end, derived from a correct compiler. Model $\t{T}$ is a type-preserving but not behavior preserving compiler refined from the correct compiler. Model $\t{I}$ was not derived from any compiler correctness theorem. 

The indirect model $\t{(I)}$ is a 3-layer neural network. Its input layer has $784\.2$ neurons, mirroring the number of pixels in two input images. These connect to a hidden layer with 400 neurons and a $\t{relu}$ nonlinearity. These hidden layer neurons connect to an output layer of 2 neurons, where each neuron represents a distinct classification (\textcolor{ForestGreen}{\cmark} or \textcolor{red}{\xmark}). In total, there are $628,000$ weights to learn.

The direct model $\t{(D)}$ is derived from compiling the following program, which checks if both inputs are $\tt$ or $\ff$.
\vspace{-.3em}
\[
\p{\bind{x}{\Bool}, \bind{y}{\Bool}}{\ite{\ul{x}}{(\ite{\ul{y}}{\ul{\tt}}{\ul{\ff}})}{(\ite{\ul{y}}{\ul{\ff}}{\ul{\tt}})}}{\Bool}
\]
\noindent The compiled program will be linked against neural networks that implement $\nb{x}$ and $\nb{y}$. If these networks can learn to recognize whether an input is even or odd, they can use the compiled program to solve the task. The neural networks that we link with are both 3-layer neural networks similar to the indirect model---each has 784 input neurons, 400 hidden neurons, and 2 output neurons. In total, there are $628,800$ weights to learn.

The Church encoded model $\t{(C)}$ is derived from compiling a Church encoding $\nb{\t{eq}}$ of the previous program. For brevity, we omit the full encoding.
\[
\p{\bind{x}{\Bool \lto (\Bool \lto \Bool)}, \bind{y}{\Bool \lto (\Bool \lto \Bool)}}{(\t{eq}\,x)\,y}{\Bool}
\]
\noindent The compiled program produces a linear hypernetwork (implementing $\t{eq})$ which will link against neural networks that implement $\nb{x}$ and $\nb{y}$, which now represent Church encoded booleans. If these networks can learn to recognize whether an input is even or odd, and to produce a Church encoded boolean to represent this, then they can use the compiled $\nb{\t{eq}}$ to solve the task. The neural networks that we link with are both 3-layer neural networks similar to the indirect model---each has 784 input neurons, 400 hidden neurons, and 8 output neurons. In total, there are $633,600$ weights to learn.

The type preserving model $\t{(T)}$ is similar to the direct model. It keeps the type translation from the correct compiler, but permits identifying programs with randomly\footnote{The exact initialization is He initialization, which help \t{relu} networks  learn effectively \cite{he2015delving}.} initialized matrices between the denoted spaces. For example, the rule for compiling conditionals is swapped with a type-preserving, but not behavior preserving variant in this way. Instead of implementing soft-branching, it takes the tensor product of outputs from the neural networks meant for detecting even or odd digits, and randomly maps them to a final output. This random linear map can change during learning, and allows the network to potentially find its own solution to soft-branching without as many constraints imposed by the compiler. In total, there are $628,808$ weights to learn.

\subsubsection{Measures}

For this experiment, the measures are simpler. Our only measure of learning dynamics is the neural network's accuracy on the task. To take this measure we provide two input images from the test dataset to each model, which will then return $\bv{y_1}{y_2} \in \rd{\R^2}$. If $\rd{y_1} \geq \rd{y_2}$ then we say the model returns \textcolor{ForestGreen}{\cmark}. Otherwise it returns \textcolor{red}{\xmark}. The accuracy we report is the accuracy across the entire test dataset when soliciting model responses this way.

\subsubsection{Results}

\begin{figure}[t]
  \includegraphics[scale=.4]{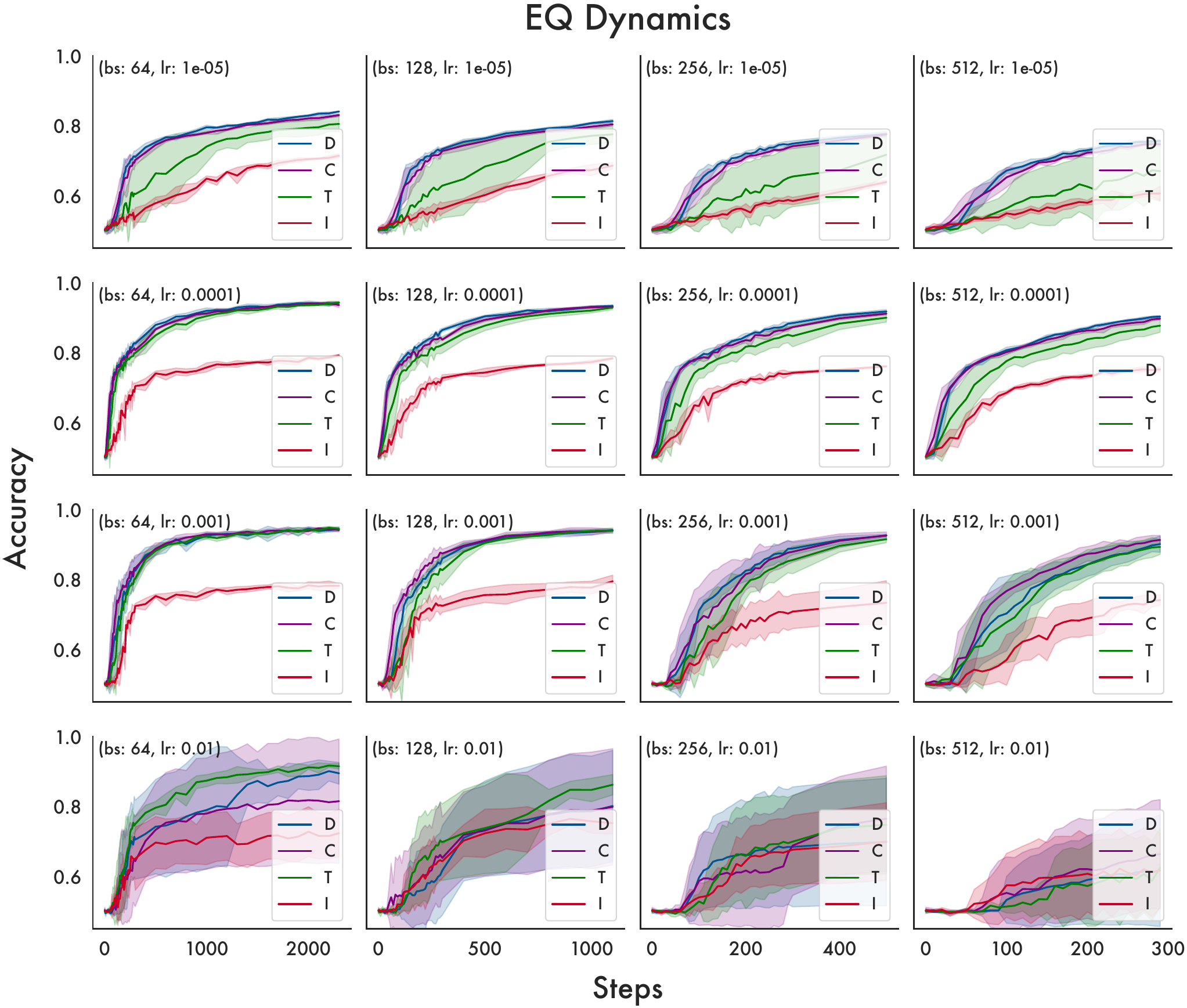}
  \centering
  \vspace{.1em}
  \caption{Summary of dynamics across configurations}
  \label{fig:global-dynamics2}

  \vspace{-.8em}
\end{figure}

To collect our data, we trained all models to approximate the function described by a training dataset of 30,000 pairs of images and the correct classification (\textcolor{ForestGreen}{\cmark} or \textcolor{red}{\xmark}). They were tested on a distinct dataset of 5,000 pairs of images and the correct classification. Both models used the Adam optimizer to compute weight updates to the neural network, using a cross entropy loss to measure model performance during training. Throughout training we recorded the test accuracy in order to understand the model's learning dynamics. We did this for 5 random seeds, 4 learning rates, and 4 training batch sizes---all can (and did) vary the learning dynamics.

Fig. \ref{fig:global-dynamics2} summarizes the learning dynamics across configurations, similar in format to Fig. \ref{fig:global-dynamics} from the first experiment. Again, we see that directly programming with a behavior preserving compiler leads to networks that learn faster and with greater data-efficiency. Interestingly, models $\t{D}$ and $\t{C}$ both perform comparably well. However, we suspect there are experiments that could distinguish these models, as hypernetworks tend to exhibit different learning dynamics to ordinary networks in general \cite{ha2016hypernetworks}. The type-preserving compiler $(\t{T})$ performs slightly worse overall than models $\t{D}$ and $\t{C}$ but better than the indirect model $\t{I}$. This shows that just getting the type translation correct can net you benefits for learning. The indirect model $(\t{I})$ generally struggles compared to those models directly programmed with a behavior preserving compiler or its refined type preserving variant. Three additional experiments are shown in Appendix C, showing similar trends on variants of this task.
\section{Related Work}
\label{sec:related}

\cj\ was deeply inspired by research in neurosymbolic programming, categorical semantics, compiler correctness, and differentiable programming. We will highlight the key insights stemming from these research areas while also taking the opportunity to identify where our work differs. 

\subsection{Neurosymbolic programming}
Neurosymbolic programs combine neural networks with discrete components. There are many approaches, but we focus on those which allow you to directly program an aspect of a neural network's function. 

\t{Scallop}, \t{TerpreT}, and \t{Differentiable Forth} are examples of programming languages where you can directly program a neural network \cite{li2023scallop, gaunt2017differentiable, bovsnjak2017programming}. They differ in programming paradigm, across logic, imperative, and probabilistic programming. But they all support a story similar to what we present with \cj. You can write programs that blend discrete and neural components, and these programs ultimately compile into neural networks with added structure that aids learning. These languages established that directly programming neural networks is possible across a rich array of problems. Our first intuitions on directly programming neural networks came from these languages. 

However, they lack a compiler correctness theory.\footnote{No compiler correctness theorems or proofs are available for these languages.} As a result, some principles underlying their design are difficult to understand or fix. For example, like \cj\ these languages lack infinite data like natural numbers or trees. This limitation stems from how types compile to vector spaces. Recall for finite data with $n$ distinct values, you compile to an $n$-dimensional vector space. With our current approach (consistent with \t{Scallop}, \t{TerpreT}, and \t{Differentiable Forth}), you would compile infinite data to an infinite-dimensional vector space; you need an infinite number of neurons to implement infinite data. \cj\ offers a simple setting to study and resolve (Section 7) these fundamental issues, and frames them for the first time as compiler correctness issues. This framing lets us leverage a rich theory on the categorical semantics of linear programs.
\subsection{Categorical semantics}

Linear types originate from linear logic, whose categorical semantics exposed a connection between linear programs and linear maps that has been studied for decades \cite{mellies2009categorical}. Of special interest are categorical semantics using linear maps between finite-dimensional vector spaces. We find their insight easier to use in a compiler.

The categorical semantics which most inspired \cj\ concerns models of lambda calculi in the category of finite vector spaces by Valiron and Zdancewic \cite{valiron2014modeling}. They translate typing judgments to morphisms in the category \t{FinVec}, whose objects are finite-dimensional vector spaces over finite fields and whose morphisms are linear maps between these vector spaces. They prove adequacy of the translation and explore language designs for which there exist fully abstract translations. Notably, they manually construct the matrices of linear maps denoted by Church numerals---an unorthodox description of discrete data as matrices. Similar descriptions of discrete programs as matrices are found in the categorical semantics of quantum programming languages, where typing judgments translate to morphisms in the category \t{FHilb} \cite{abramsky2004categorical}. The objects of $\t{FHilb}$ are finite-dimensional Hilbert spaces and its morphisms are linear maps between these Hilbert spaces. These descriptions of discrete data as matrices made us wonder whether they could be applied to directly program neural networks.

But these matrices are of the wrong kind for directly programming neural networks. The entries of matrices from $\t{FinVec}$ are from finite fields like $\mathbb{Z}_3=\{0,1,2\}$, the integers modulo 3. The entries of matrices from $\t{FHilb}$ are complex numbers like 1 + 2i. To directly program neural networks the entries of our matrix should be real numbers. There are semantics of linear programs as linear maps between real-valued vector spaces, however these don't consider discrete data like booleans \cite{mellies2009categorical}. Beyond this, we are unaware of any semantics which discusses the relation between higher-order linear programs and linear hypernetworks, or which considers the difficulty in proving practical theorems when linking against nonlinear neurons.

\vspace{-.1em}
\subsection{Compiler correctness}

Throughout the paper our intuition for a correct compiler was one that allows useful predictions about what compiled programs will and won't do. This perspective was inspired by the following observation.

\epigraph{%
  \textsf{``For a compiler correctness theorem to assure complete trust, the theorem must reflect the reality of how the
compiler will be used.''}%
}{\textsf{D. Patterson and A. Ahmed,} \par\textsf{The Next 700 Compiler Correctness Theorems \cite{patterson2019next}}}

\noindent In their paper, Patterson and Ahmed point highlight ways in which compiler correctness theorems can be mathematically correct, and yet unusable in practice. For example, a compiler correctness theorem only concerning closed programs says nothing about the behavior of programs under linking. This insight, highlighted but not unique to Patterson and Ahmed, underlies many works on \textit{compositional} compiler correctness. Collectively, these efforts motivating their theorems acted as a compass when designing the appropriate compiler correctness theorem in \cj.

To our knowledge, there are \textit{zero} papers on proving the correctness of a compiler for directly programming neural networks with discrete data. Our work shows that proving correctness requires a unique balance of proof and experiment. The behavior of programs linked against nonlinear neurons are difficult, if impossible to predict \cite{saxe2015deep}. Therefore experiment plays a key role in the theory of linking---which aims to ensure that directly programming neural networks improves various measures of performance. This relationship between theory and experiment is atypical in the compiler correctness literature. However, it is common in the study of dynamical systems.

\noindent  

%
\subsection{Differentiable programming}

Differentiable programming languages like $\t{JAX}$ and $\t{PyTorch}$ are popular for building neural networks \cite{jax2018github, paszke2017automatic}. Their key feature is automatic differentiation, which frees programmer from hand-designing the derivatives necessary to implement learning algorithms. They also greatly simplify and accelerate training by making it easy to train neural networks on specialized hardware like Graphics Processing Units (GPU). These features informed our decision to implement \cj\ as an embedded language within \t{PyTorch}, allowing us to embed a direct-style of programming neural networks inside an efficient and expressive programming language. There has been a lot of interest in proving the correctness of automatic differentiation in these languages, and more recently in how to translate extremely large neural networks to specialized hardware \cite{partir_ml}.

However, these efforts don't concern themselves with discrete data because they aren't immediately differentiable. In instances where conditionals are discussed, the conditionals are implemented in such a way that prevents directly programming neural networks (Section \ref{sec:formal-approach}). The study of \cj\ and its extensions provide a path for languages like \t{JAX} and \t{PyTorch} to support discrete data as first-class entities, enabling a rich interplay between learning and the discrete structures of ordinary programming.
\section{Discussion}
\label{sec:discussion}

\cj\ is a minimal language for studying how to directly program neural networks. This paper takes initial steps in developing its programming language theory. We identify theorems worth proving, and also the limits to what we can expect of our compiler correctness theorems when linking against nonlinear neurons. To supplement these limits we emphasize the role of experiments in verifying the correctness of compilers for directly programming neural networks. Our data are consistent with a broader empirical literature in machine learning---directly programming neural networks can make neural networks learn more efficiently, and in ways that are easier to interpret. The data also suggest that the strength of this impact relates to the strength of the compiler correctness theorem satisfied by the compiler producing the neural network. However, many questions remain. The limitations on expressive power and efficiency are paramount.

On expressive power, \cj\ is missing many types. There are simple extensions---unit, sums, and products. They can be added without significantly impacting the theory in this paper. No additional lemmas would be needed, just additional cases in the compiler and proofs to consider. To permit affine functions involves trickier changes to the theory. Several typing rules, and hence several rules in the compiler would change. The key lemma identifying linear programs with linear maps would change to instead identify affine programs with affine maps. However, the frontier concerns infinite data. No language for directly programming neural networks supports infinite data. We think the answer involves compiling to \textit{recurrent} neural networks. In separate work we are developing \Cjn, a variant with natural numbers. This involves substantial changes to the theory, identifying linear programs instead with linear dynamical systems.

On efficiency, a chief concern is how quickly the number of neurons needed to implement a program grow as a function of its type. The number of neurons needed to implement a program relates to the dimension of the underlying vector space its type compiles to. \cj\ exhibits bilinear scaling because $\rd{\t{dim}}(\nb{\ty_1 \lto \ty_2}) = \rd{\t{dim}}(\nb{\ty_1})\,\rd{\.}\,\rd{\t{dim}}(\nb{\ty_2})$. For sufficiently higher-order programs this is insufficient. This scaling stems from a demand that programs are \textit{exactly} implemented by linear maps (Section 4.2.2). If we relax this demand there may be a route to improved scaling via dimensionality reduction, which can approximately embed vectors into spaces of smaller dimension \cite{van2009dimensionality}. These are areas of future work. 

\cj\ is a foundation from which to explore a direct style of programming neural networks. As a minimal vehicle, its theory is tractable, and the questions posed in this work can be studied with recourse to a deep bench of existing techniques in programming language theory. Ultimately, these studies may uncover a rich interplay between learning and the discrete structures of ordinary programming.

\section*{Acknowledgments}

We thank the anonymous reviewers for detailed feedback that improved the clarity and quality of this work.

\bibliographystyle{ACM-Reference-Format}
\bibliography{bibliography.bib}
\clearpage 

\appendix
\newtheorem{definitionx}{Definition}[section] 

\newenvironment{ndef}[1]{%
  \begin{definitionx}[#1]%
}{%
  \end{definitionx}%
}

\section{Definitions}

\subsection{Context Splitting}
\label{app:split}

\begin{center}
\bottomAlignProof
\AxiomC{$\nb{\D} = \nb{\D_1} \,\cup\, \nb{\D_2}$}
\AxiomC{$\nb{\emp} = \nb{\t{dom}}{(\nb{\D_1})} \,\cap\, \nb{\t{dom}}{(\nb{\D_2})}$}
\BinaryInfC{$\nb{\D} = \nb{\D_1} \o \nb{\D_2}$}
\DisplayProof
\end{center}
\vspace{1em}

\subsection{Source environment}
\label{app:src-env}

\begin{center}
\bottomAlignProof
\AxiomC{}
\UnaryInfC{$\nb{\emp} \- \nb{\emp}$}
\DisplayProof
\quad\quad
\bottomAlignProof
\AxiomC{$\nb{\D}\-\nb{\d}$}
\AxiomC{$\p{\emp}{v}{\ty} $}
\BinaryInfC{$\nb{\D,\bind{x}{\ty}} \- \nb{\d\{x \map v\}}$}
\DisplayProof
\end{center}
\vspace{1em}

\subsection{Target environment}
\label{app:trg-env}

\begin{center}
\bottomAlignProof
\AxiomC{}
\UnaryInfC{$\nb{\emp} \- \rd{\emp}$}
\DisplayProof
\quad\quad
\bottomAlignProof
\AxiomC{$\nb{\D} \- \sv{}$}
\AxiomC{$\vt{}\in \c{\nb{\ty}}$}
\BinaryInfC{$\nb{\D,\bind{x}{\ty}} \- \sv{}\rd{\{\nb{x} \map \vt{}\}}$}
\DisplayProof
\end{center}
\vspace{1em}

\subsection{Compiling types}

$$\c{\nb{\_}}: \nb{\t{Type}} \to \rd{\t{Vector Space}}$$\\[-1em]
$$
\begin{aligned}
\c{\nb{\Bool}} &= \rd{\R^{2}}\\
\c{\nb{\ty_1 \lto \ty_2}} &= \rd{\R^{\rd{\t{dim}}(\nb{\ty_2})\,\rd{\x}\,\rd{\t{dim}}(\nb{\ty_1})}}\\
\end{aligned}
$$
\\

\subsection{Compiling environments}
\begin{center}
\bottomAlignProof
\AxiomC{}
\UnaryInfC{$\c{\nb{\emp \-\emp}} = \rd{\emp}$}
\DisplayProof
\quad\quad
\bottomAlignProof
\AxiomC{$\c{\nb{\D\-\d}}=\rd{\sv{}}$}
\AxiomC{$\limp{\p{\emp}{v}{\ty}}{\emp}{\vt{}}$}
\BinaryInfC{$\c{\nb{\D,\bind{x}{\ty} \-\d\{x \map v\}}} = \rd{\sv{}\{\nb{x} \map \vt{}\}}$}
\DisplayProof
\end{center}
\vspace{1em}

\subsection{Matrix}

$$
\Mat = \rd{\begin{bmatrix}  
\alpha_{11}  & \!\!\!\!\cdots\!\!\!\! &  \alpha_{1n} \\
\vdots       & \!\!\!\!\ddots\!\!\!\! &  \vdots \\
\alpha_{1m}  & \!\!\!\!\cdots\!\!\!\! &  \alpha_{mn}
\end{bmatrix}}
$$
\vspace{1em}

\subsection{Reshaping a matrix to a vector}
\label{app:vec}
\t{For $\rd{i}=\rd{1..m}$ and $\rd{j}=\rd{1..n}$,}
$$\rd{\t{vec}(\Mat)}=\rd{\t{[}\beta_{m\.n}\t{]}} \;\textsf{ where }\; \rd{\beta_{i+m(j-1)}}=\rd{\alpha_{ij}}$$
\t{This definition takes a matrix and vertically concatenates all column vectors from left to right.}
\vspace{1em}

\subsection{Reshaping a vector to a matrix}
\label{app:reshape}
\t{For $\rd{i}=\rd{1..m}$ and $\rd{j}=\rd{1..n}$, where $\rd{\t{dim}}(\nb{\ty_2})=\rd{m}$ and $\rd{\t{dim}}(\nb{\ty_1})=\rd{n}$,
$$\rd{\t{[}\beta_{m\.n}\t{]}_{\nb{\ty_1\lto\ty_2}}}=\Mat \;\text{ where }\; \rd{\alpha_{ij}}=\rd{\beta_{i+m(j-1)}}$$
For $\nb{\ty}\neq \nb{\ty_1\lto\ty_2}$,$$\rd{\t{[}\beta\t{]}_\nb{\ty}}=\rd{\beta}$$}
\vspace{1em}

\subsection{Basis of a type}
\label{app:basis}
\t{Where $\rd{\t{dim}}(\nb{\ty})=\rd{n}$,}
$$\begin{aligned}
\rd{\t{basis}}\c{\nb{\ty}}&=\overbrace{\left(\rd{\begin{bmatrix}1 \\ 0 \\ \vdots \\ 0\end{bmatrix}_\nb{\!\ty}},\rd{\begin{bmatrix}0 \\ 1 \\ \vdots \\ 0\end{bmatrix}_\nb{\!\ty}},\;\cdots\;,\rd{\begin{bmatrix}0 \\ 0 \\ \vdots \\ 1\end{bmatrix}_\nb{\!\ty}}\right)}^\rd{n}
\end{aligned}$$
\vspace{1em}

\subsection{Matrix multiplication}
\label{app:matmul}
\t{Where $\bout{i}\in\rd{\t{basis}}\c{\nb{\ty}}$,}
$$\rd{\et1(\et2)_\nb{\ty}}=\rd{\et1(\t{vec}(\et2))_\nb{\ty}}=\rd{\Mat(\beta_j)_\nb{\ty}} = \rd{\sum_i\sum_j\alpha_{ij}\.\beta_j\.\bout{i}}
$$
\vspace{1em}

\subsection{Logical Typing}
$$\lp{\D}{e}{\ty} \;\iff\; \forall \nb{\d} \in \Nv{\D},\, \nb{\d(e)}\in \E{\ty}$$
\vspace{1em}

\subsection{Logical Relation}
$$\E{\_}: \nb{\t{Type}} \to \Big\{\nb{\t{Expression}}\Big\}$$\\[-1em]
\[
\begin{aligned}
\V{\Bool}&=\Big\{\begin{array}{l}
\nb{\tt},\nb{\ff}
\end{array}\Big\}\\
\V{\ty_1 \lto \ty_2} &=
\Big\{\begin{array}{l}
\nb{\lam{x}{e}} \mid \forall\nb{v}\in \V{\ty_1},\nb{(\lam{x}{e})v} \in \E{\ty_2}\\
\end{array}
\Big\}\\\\
\E{\ty} &= \left\{
\nb{e} \;\middle|\;
\begin{array}{l}
\nb{e} \bstep \nb{v},\\
\nb{v}\in \V{\ty}\\
\end{array}
\right\}\\\\
\end{aligned}
\]

$$\Nv{\_}: \nb{\t{Context}} \to \Big\{\nb{\t{Environment}} \Big\}$$\\[-1em]
$$
\begin{aligned}
\Nv{\emp} &= \Big\{\nb{\emp} \Big\}\\
\Nv{\D,\bind{x}{\ty}} &= 
\Big\{\begin{array}{l}
 \nb{\d\{x \map v\}}\mid \nb{\d} \in \Nv{\D}, \nb{v} \in \V{\ty}
\end{array}
\Big\}
\end{aligned}
$$
\vspace{1em}

\clearpage
\section{Proofs}

\subsection{Theorem (Compiler preserves program behavior)}
\label{app:ccc}
\t{Where $\limp{\p{\emp}{e}{\Bool}}{\emp}{\et{}}$ and $\limp{\p{\emp}{v}{\Bool}}{\emp}{\vt{}}$,
\begin{flalign*}
&\begin{aligned}
(a) \; & \step{e}{v}  \implies \et{}=\vt{} \\
(b) \; & \nb{e} \;\cancel{\bstep}\; \nb{v} \implies \et{}\neq\vt{}\\
\end{aligned}&&
\end{flalign*} }

\begin{proof}\leavevmode\\
\t{
By assumption $\limp{\p{\emp}{e}{\Bool}}{\emp}{\et{}}$\\
By assumption $\limp{\p{\emp}{v}{\Bool}}{\emp}{\vt{}}$\\
Consider each condition,
\setlist[itemize]{topsep=4pt, itemsep=4pt, parsep=2pt, leftmargin=1.5em, rightmargin=1.5em}
\begin{itemize}[label=$\triangleright$]
\item \textbf{Condition} $(a)$,\\[.25em]
	Because compiler preserves what programs will do $\et{}=\vt{}$
\item \textbf{Condition} $(b)$,\\[.25em]
	Because compiler preserves what programs won't do $\et{} \neq \vt{}$
\end{itemize}}
\end{proof}
\vspace{1em}

\subsection{Corollary (Compiler preserves open program behavior)}
\t{Where $\limp{\p{\D}{e}{\Bool}}{\sv{}}{\et{}}$, $\limp{\p{\emp}{v}{\Bool}}{\emp}{\vt{}}$ and $\c{\nb{\D} \- \nb{\sigma}} = \sv{}$,
\begin{flalign*}
&\begin{aligned}
(a) \; & \step{\sigma(e)}{v}  \implies \et{}=\vt{} \\
(b) \; & \nb{\sigma(e)} \;\cancel{\bstep}\; \nb{v} \implies \et{}\neq\vt{}\\
\end{aligned}&&
\end{flalign*} }

\begin{proof}\leavevmode\\
\t{A corollary of compiler being invariant to closing environments and the compiler preserving closed program behavior.}
\end{proof}
\vspace{1em}

\subsection{Lemma (Compiler preserves what programs will do)}
\label{app:cclemma1}
\t{Where $\limp{\p{\emp}{e}{\ty}}{\emp}{\et{}}$ and $\limp{\p{\emp}{v}{\ty}}{\emp}{\vt{}}$,
$$\step{e}{v} \implies \et{}=\vt{}$$}

\begin{proof}\leavevmode\\
\t{Consider each inductive case of $\step{e}{v}$,
\setlist[itemize]{topsep=4pt, itemsep=4pt, parsep=2pt, leftmargin=1.5em, rightmargin=1.5em}
\begin{itemize}[label=$\triangleright$]
\item \textbf{Case} $\step{v}{v}$,\\[.25em]
	Because equality is reflexive $\vt{}=\vt{}$
\item \textbf{Case} $\step{\ite{e_1}{e_2}{e_3}}{v_2}$,\\[.25em]
	By inversion on evaluation $\step{e_1}{\tt}$ and $\step{e_2}{v_2}$\\
	By inversion on compiler and because it maps to neurons $\limp{\p{\emp}{e_1}{\ty}}{\emp}{\et1} \in\c{\nb{\Bool}}$\\
	By inversion on compiler and because it maps to neurons $\limp{\p{\emp}{e_2}{\ty}}{\emp}{\et2} \in\c{\nb{\ty}}$\\
	By inversion on compiler and because it maps to neurons $\limp{\p{\emp}{e_3}{\ty}}{\emp}{\et3} \in\c{\nb{\ty}}$\\
	By assumption and compiler mapping to neurons $\limp{\p{\emp}{v_2}{\ty}}{\emp}{\vt2}\in\c{\nb{\ty}}$\\
	By induction $\et2=\vt2$\\
	By compiler $\limp{\p{\emp}{\tt}{\Bool}}{\emp}{\begin{bmatrix} 1 &\!\!\! 0\end{bmatrix}^\tr}$  \\
	By induction $\et1=\rd{\begin{bmatrix} 1 &\!\!\! 0\end{bmatrix}^\tr}$\\
	By equality $\limp{\p{\emp}{\ite{e_1}{e_2}{e_3}}{\ty}}{\emp}{1\.\et2+0\.\et3}=\et2=\vt2$
\item \textbf{Case} $\step{\ite{e_1}{e_2}{e_3}}{v_3}$,\\[.25em]
	Similar to previous case but using induction hypothesis on other branch
\item \textbf{Case} $\step{e_1e_2}{v}$,\\[.25em]
	\textbf{Show} $\rd{\et1(\et2)}_{\nb{\ty_2}}=\vt{}$\\
	By inversion on evaluation $\step{e_1}{\lam{x}{e}}$ and $\step{e_2}{v_2}$ and $\step{\{x \map v_2\}(e)}{v}$\\
	By inversion on compiler and because it maps to neurons,\\ 
    $\limp{\p{\emp}{e_1}{\ty_1 \lto \ty_2}}{\emp}{\et1} \in \c{\nb{\ty_1 \lto \ty_2}}$\\
	By compiler and because it maps to neurons,\\ 
    $\limp{\p{\emp}{\lam{x}{e}}{\ty_1 \lto \ty_2}}{\emp}{\Mat} \in \c{\nb{\ty_1\lto\ty_2}}$ \\
	By induction $\rd{\et1}=\Mat$ \\
	\textbf{Show} $\rd{\Mat(\rd{\t{vec}}(\et2))_{\nb{\ty_2}}}=\vt{}$\\
	By inversion on compiler and because it maps to neurons,\\
    $\limp{\p{\emp}{e_2}{\ty_1}}{\emp}{\et2} \in \c{\nb{\ty_1}}$\\
	By compiler and because it maps to neurons,\\
    $\limp{\p{\emp}{\vs2}{\ty_1}}{\emp}{\vt2} \in \c{\nb{\ty_1}}$  \\
	By induction $\et2=\vt2$ \\
	Because equality is a congruence $\rd{\t{vec}}(\et2)=\rd{\t{vec}}(\vt2)=\rd{\begin{bmatrix}\vt{21} & \!\!\!\cdots\!\!\! & \vt{2n}\end{bmatrix}^\tr}$\\
	\textbf{Show} $\rd{\sum_i\sum_\rd{j} \alpha_{ij}\.\vt{2j}\.\bout{i}}=\vt{}$ \\
	By inversion on compiler and because it maps to neurons,\\
    $\limp{\p{\bind{x}{\ty_1}}{e}{\ty_2}}{\{\nb{x}\map  \bin{j}\}}{\sum_i\alpha_{ij}\.\bout{i}} \in \c{\nb{\ty_2}}$\\
	By compiler and the basis theorem $\c{\nb{\{x \map v_2\}}}=\rd{\{\nb{x} \map \vt2\}}=\rd{\{\nb{x} \map \sum_j\vt{2j}\.\bin{j}\}}$\\
	Because compiler is linear in its environment,\\
    $\limp{\p{\bind{x}{\ty_1}}{e}{\ty_2}}{\rd{\{\nb{x} \map \sum_j\vt{2j}\.\bin{j}\}}}{\sum_j \vt{2j}\.\sum_i\alpha_{ij}\.\bout{i}}$\\
	Because multiplication distributes over addition,\\
    $\limp{\p{\bind{x}{\ty_1}}{e}{\ty_2}}{\rd{\{\nb{x} \map \sum_j\vt{2j}\.\bin{j}\}}}{\sum_i\sum_\rd{j} \alpha_{ij}\.\vt{2j}\.\bout{i}}$\\
	Because the compiler is invariant to closing environments,\\
    $\limp{\p{\emp}{\{x \map v_2\}(e)}{\ty_2}}{\emp}{\sum_i\sum_\rd{j} \alpha_{ij}\.\vt{2j}\.\bout{i}}$\\
	By induction $\rd{\sum_i\sum_\rd{j} \alpha_{ij}\.\vt{2j}\.\bout{i}}=\vt{}$\\
\end{itemize}
    }
\end{proof}
\vspace{1em}

\subsection{Lemma (Compiler preserves what programs won't do)}
\t{Where $\limp{\p{\emp}{e}{\Bool}}{\emp}{\et{}}$ and $\limp{\p{\emp}{v}{\Bool}}{\emp}{\vt{}}$},
$$\es{} \not\bstep \vs{} \implies \et{}\neq\vt{}$$
\begin{proof}\leavevmode\\
\t{Because closed programs terminate $\es{} \bstep \vs1$ where $\vs1\neq\vs{}$ by assumption\\ 
Because of canonical forms $\vs1$ and $\vs{}$ could be either $\nb{\tt}$ or $\nb{\ff}$\\
Proceed by cases on $(\vs1,\vs{})$,
\setlist[itemize]{topsep=4pt, itemsep=4pt, parsep=2pt, leftmargin=1.5em, rightmargin=1.5em}
\begin{itemize}[label=$\triangleright$]
\item \textbf{Case} $(\nb{\tt},\nb{\ff})$\\[.25em]
	By negation $\et{}=\vt{}$\\
	Because the compiler preserves observable behavior $\et{}=\rd{\begin{bmatrix}1 &\!\!\! 0\end{bmatrix}^\tr}=\vt{}=\rd{\begin{bmatrix}0 &\!\!\! 1\end{bmatrix}^\tr}$\\
	But $\rd{\begin{bmatrix}1 &\!\!\! 0\end{bmatrix}^\tr}\neq\rd{\begin{bmatrix}0 &\!\!\! 1\end{bmatrix}^\tr}$ so we have a contradiction
\item \textbf{Case} $(\nb{\ff},\nb{\tt})$\\[.25em]
	Similar to previous case
\end{itemize}}
\end{proof}
\vspace{1em}

\subsection{Theorem (Programs evaluate to values)}
\label{proof:eval_values}
$$\p{\D}{e}{\ty} \implies \exists \nb{v},\step{e}{v}$$
\begin{proof}\leavevmode\\
\t{
Because programs can be logically typed $\lp{\D}{e}{\ty}$\\
By logical typing $\nb{\sigma(e)}\in\E{\ty}$\\
Because logically typed programs terminate $\exists \nb{v},
\step{\sigma(e)}{v}$
}
\end{proof}

\subsection{Lemma (Programs can be logically typed)}
$$\p{\D}{e}{\ty} \implies \lp{\D}{e}{\ty}$$

\begin{proof}\leavevmode\\
\t{Consider each inductive case of typing,
\setlist[itemize]{topsep=4pt, itemsep=4pt, parsep=2pt, leftmargin=1.5em, rightmargin=1.5em}
\begin{itemize}[label=$\triangleright$]
\item \textbf{Case} $\p{\emp}{\tt}{\Bool}$\\[.25em]
	\textbf{Show} $\nb{\tt} \in \E{\Bool}$\\
	Consider each condition,
    \begin{itemize}[label=$\triangleright$]
	\item \textbf{Condition} $\exists \nb{v},\step{\tt}{v}$\\[.25em]
		Let $\exists \nb{v}=\nb{\tt}$\\
		By evaluation $\nb{\tt} \bstep \nb{\tt}$
	\item \textbf{Condition} $\nb{\tt} \in \V{\Bool}$\\[.25em]
		By logical relation $\nb{\tt} \in \V{\Bool}$
    \end{itemize}
\item \textbf{Case} $\p{\emp}{\ff}{\Bool}$\\[.25em]
	Similar to previous case
\item \textbf{Case} $\p{\bind{x}{\ty}}{x}{\ty}$\\[.25em]
	\textbf{Show} $\nb{\{x \map v\}(x)}\in\E{\ty}$ where $\nb{\{x \map v\}}\in\Nv{\bind{x}{\ty}}$ \\
	Consider each condition,
    \begin{itemize}[label=$\triangleright$]
	\item \textbf{Condition} $\exists \nb{v_1}, \step{\{x \map v\}(x)}{v_1}$\\[.25em]
		By substitution $\nb{\{x \map v\}(x)}=\nb{v}$\\
		Let $\exists \nb{v_1}=\nb{v}$\\
		By evaluation $\step{v}{v}$
	\item \textbf{Condition} $\nb{v} \in \V{\ty}$\\[.25em]
		By logical relation $\nb{\{x \map v\}}\in\Nv{\bind{x}{\ty}}$ where $\nb{v} \in \V{\ty}$
    \end{itemize}
\item \textbf{Case} $\p{\D}{\lam{y}{e}}{\ty_1 \lto \ty_2}$\\[.25em]
	\textbf{Show} $\nb{\lam{y}{\d(e)}}\in\E{\ty_1 \lto \ty_2}$ where $\nb{\d}\in\Nv{\D}$\\
	Consider each condition,
    \begin{itemize}[label=$\triangleright$]
	\item \textbf{Condition} $\exists \nb{v}, \step{\lam{y}{\d(e)}}{v}$\\[.25em]
		Let $\exists \nb{v}=\nb{\lam{y}{\d(e)}}$\\
		By evaluation $\step{\lam{y}{\d(e)}}{\lam{y}{\d(e)}}$
	\item \textbf{Condition} $\nb{\lam{y}{\d(e)}}\in\V{\ty_1\lto\ty_2}$\\[.25em]
		\textbf{Show} $\nb{(\lam{y}{\d(e)})v_1} \in \E{\ty_2}$\\
		By assumption $\nb{v_1} \in \V{\ty_1}$\\
		Consider each condition,
        \begin{itemize}[label=$\triangleright$]
		\item \textbf{Condition} $\exists \nb{v}, \step{(\lam{y}{\d(e)})v_1}{v}$\\[.25em]
			By inversion on typing $\p{\D,\bind{y}{\ty_1}}{e}{\ty_2}$\\
			By induction $\lp{\D,\bind{y}{\ty_1}}{e}{\ty_2}$ i.e. $\nb{\d\{y \map v_1\}(e)} \in \E{\ty_2}$ \\
			By logical relation $\step{\d\{y \map v_1\}(e)}{v}$ where $\nb{v} \in \V{\ty_2}$ \\
			Because substitutions commute $\nb{\d\{y \map v_1\}(e)}=\nb{\{y \map v_1\}(\d(e))}$ \\
			Let $\exists \nb{v}=\nb{v}$ \\
			By evaluation $\step{(\lam{y}{\d(e)})v_1}{v}$ 
		\item \textbf{Condition} $\nb{v} \in \V{\ty_2}$\\[.25em]
			By logical relation $\step{\{y \map v_1\}(\d(e))}{v}$ where $\nb{v} \in \V{\ty_2}$ 
        \end{itemize}
    \end{itemize}
\item \textbf{Case} $\p{\D_1 \o \D_2}{e_1e_2}{\ty_2}$\\[.25em]
	\textbf{Show} $\nb{\d_1(e_1)\d_2(e_2)} \in \E{\ty_2}$\\
	Consider each condition,
    \begin{itemize}[label=$\triangleright$]
	\item \textbf{Condition} $\exists \nb{v},\step{\d_1(e_1)\d_2(e_2)}{v}$ \\[.25em]
		By inversion $\p{\D_1}{e_1}{\ty_1 \lto \ty_2}$\\
		By inversion $\p{\D_2}{e_2}{\ty_1}$ \\
		By induction $\lp{\D_1}{e_1}{\ty_1 \lto \ty_2}$\\
		By induction $\lp{\D_2}{e_2}{\ty_1}$ \\
		By logical relation $\step{\d_1(e_1)}{\lam{x}{e}}$ where $\nb{\lam{x}{e}}\in\V{\ty_1 \lto \ty_2}$\\
		By logical relation $\step{\d_2(e_2)}{v_2}$ where $\nb{v_2}\in\V{\ty_1}$ \\
		By logical relation $\step{(\lam{x}{e})v_2}{v}$ where $\nb{v}\in\V{\ty_2}$ \\
		Let $\exists \nb{v}=\nb{v}$ \\
		By evaluation $\step{\d_1(e_1)\d_2(e_2)}{v}$
	\item \textbf{Condition} $\nb{v} \in\V{\ty_2}$ \\[.25em]
		By logical relation $\step{(\lam{x}{e})v_2}{v}$ where $\nb{v}\in\V{\ty_2}$ 
    \end{itemize}
\item \textbf{Case} $\p{\D_1 \o \D_2}{\ite{e_1}{e_2}{e_3}}{\ty}$\\[.25em]
	\textbf{Show} $\nb{\ite{\d_1(e_1)}{\d_2(e_2)}{\d_2(e_3)}} \in \E{\ty}$\\ 
	By inversion on typing $\p{\D_1}{e_1}{\Bool}$ \\
	By induction $\lp{\D_1}{e_1}{\Bool}$ \\
	By logical relation $\step{\d_1(e_1)}{v_1}$ where $\nb{v_1}\in\V{\Bool}=\{\nb{\tt},\nb{\ff}\}$  \\
	Consider each case on $\nb{v_1}=\nb{\tt}$ or $\nb{v_1}=\nb{\ff}$,
    \begin{itemize}[label=$\triangleright$]
	\item \textbf{Case} $\nb{v_1}=\nb{\tt}$\\[.25em]
		Consider each condition,
        \begin{itemize}[label=$\triangleright$]
		\item \textbf{Condition} $\exists \nb{v}, \step{\ite{\d_1(e_1)}{\d_2(e_2)}{\d_2(e_3)}}{v}$\\[.25em]
			By inversion on typing $\p{\D_2}{e_2}{\ty}$ and $\p{\D_2}{e_3}{\ty}$ \\
			By induction $\lp{\D_2}{e_2}{\ty}$ and $\lp{\D_2}{e_3}{\ty}$ \\
			By logical relation $\step{\d_2(e_2)}{v_2}$ where $\nb{v_2}\in\V{\ty}$ \\
			Let $\exists \nb{v}=\nb{v_2}$\\
			By evaluation $\step{\ite{\d_1(e_1)}{\d_2(e_2)}{\d_2(e_3)}}{v_2}$
		\item \textbf{Condition} $\nb{v_2}\in\V{\ty}$ \\[.25em]
			By logical relation $\step{\d_2(e_2)}{v_2}$ where $\nb{v_2}\in\V{\ty}$ 
        \end{itemize}
	\item \textbf{Case} $\nb{v_1}=\nb{\ff}$ \\[.25em]
		Similar to previous case
    \end{itemize}
\end{itemize}}
\end{proof}
\vspace{1em}

\subsection{Lemma (Logically typed programs evaluate to values)}
\t{Where $\nb{\d}\in\Nv{\D}$,
$$\lp{\D}{e}{\ty} \implies \exists \nb{v},\step{\d(e)}{v}$$}
\begin{proof}\leavevmode\\
\t{By logical typing $\nb{\d(e)}\in\E{\ty}$ \\
By logical relation $\exists \nb{v},\step{\d(e)}{v}$}
\end{proof}
\vspace{1em}

\subsection{Lemma (Compilers maps programs to neurons)}
$$\limp{\p{\D}{e}{\ty}}{\sv{}}{\et{}} \implies \et{}\in \c{\nb{\ty}}$$
\begin{proof}\leavevmode\\
\t{Consider each inductive case of compiler,
\begin{itemize}[label=$\triangleright$]
\item \textbf{Case} $\limp{\p{\bind{x}{\ty}}  {x}  {\ty}} {\{\nb{x}\map \vt{}\}} {\vt{}}$\\[.25em]
	By inversion on target environments $\vt{}\in\c{\nb{\ty}}$ 
\item \textbf{Case} $\limp{\p{\emp} {\tt} \Bool}{\emp}{\begin{bmatrix} 1 &\!\!\! 0\end{bmatrix}^\tr}$\\[.25em]
	By compiling types $\c{\nb{\Bool}}=\rd{\R^2}$ \\
	By definition $\rd{\R^2}=\rd{\{\alpha_1\.\begin{bmatrix} 1 &\!\!\! 0\end{bmatrix}^\tr + \alpha_2\.\begin{bmatrix} 0 &\!\!\! 1\end{bmatrix}^\tr \,\mid \alpha_i \in \R\}}$ \\
	By equality $\rd{\begin{bmatrix} 1 &\!\!\! 0\end{bmatrix}^\tr}=\rd{1\.\begin{bmatrix} 1 &\!\!\! 0\end{bmatrix}^\tr+0\.\begin{bmatrix} 0 &\!\!\! 1\end{bmatrix}^\tr} \in \rd{\R^2}$ 
\item \textbf{Case} $\limp{\p{\emp} {\ff} \Bool}{\emp}{\begin{bmatrix} 0 &\!\!\! 1\end{bmatrix}^\tr}$\\[.25em]
	Similar to previous case
\item \textbf{Case} $\limp{\p{\D} {\lam{x}{e}} {\ty_1 \lto \ty_2}}{\sv{}}{ \Mat}$ \\[.25em]
	\textbf{Show} $\Mat \in \rd{\R^{\rd{\t{dim}}(\nb{\ty_2})\,\rd{\x}\,\rd{\t{dim}}(\nb{\ty_1})}}$\\
	By compiling types $\c{\nb{\ty_1\lto\ty_2}}=\c{\rd{\R^{\rd{\t{dim}}(\nb{\ty_2})\,\rd{\x}\,\rd{\t{dim}}(\nb{\ty_1})}}}$\\
	By inversion on compiler $\limp{ \p{\D,\bind{x}{\ty_1}} {e}  {\ty_2}}{\sv{}\{\nb{x} \map \bin{j}\}}{\sum_i\alpha_{ij}\.\bout{i}}$\\
	By induction $\rd{\sum_i\alpha_{ij}\.\bout{i}} \in \c{\nb{\ty_2}}$\\
	By inversion on compiler $\rd{i}=\rd{1..}\,\rd{\t{len}}(\rd{\t{basis}}\c{\nb{\ty_2}})=\rd{\t{dim}}(\nb{\ty_2})$ \\
	By inversion on compiler $\rd{j}=\rd{1..}\,\rd{\t{len}}(\rd{\t{basis}}\c{\nb{\ty_1}})=\rd{\t{dim}}(\nb{\ty_1})$ \\
	Therefore $\Mat \in \rd{\R^{\rd{\t{dim}}(\nb{\ty_2})\,\rd{\x}\,\rd{\t{dim}}(\nb{\ty_1})}}$
\item \textbf{Case} $\limp{\p{\D_1 \o \D_2} {e_1e_2} {\ty_2}}{\sv1 \o \sv2}{\et1(\et2)}$\\[.25em]
	\textbf{Show} $\rd{\rd{\et1}(\t{vec}(\et2))_\nb{\ty_2}}\in\c{\nb{\ty_2}}$ \\
	By inversion on compiler $\limp{\p{\D_1} {e_1} {\ty_1 \lto \ty_2}}{\sv1}{\et1}$ and $\limp{\p{\D_2} {e_2} {\ty_1}}{\sv2}{\et2}$  \\
	By induction $\et1\in\c{\nb{\ty_1\lto\ty_2}}=\rd{\R^{\t{dim}(\nb{\ty_2})\x\t{dim}(\nb{\ty_1})}}$ and $\et2\in\c{\nb{\ty_1}}$\\
	By vectorizing $\rd{\t{vec}(\et2)} \in \rd{\R^{\t{dim}(\nb{\ty_1})}}$ \\
	By matrix multiplication $\rd{\et1(\t{vec}(\et2))_\nb{\ty_2}} \in \c{\nb{\ty_2}}$
\end{itemize}}
\end{proof}
\vspace{1em}

\subsection{Lemma (Compiler maps programs to linear maps)}
\label{app:cclemma2}
\t{Where $\limp{\p{\D}{e}{\ty}}{\sv{}\{\nb{x}\map\vt{1}\}}{\et{1}}$ and $\limp{\p{\D}{e}{\ty}}{\sv{}\{\nb{x}\map\vt{2}\}}{\et{2}}$,
$$\limp{\p{\D}{e}{\ty}}{\sv{}\{\nb{x} \map\alpha_1\.\vt1+\alpha_2\.\vt2\}}{\alpha_1\.\et1+\alpha_2\.\et{2}}$$}
\begin{proof}\leavevmode\\
\t{Direct consequence of compiler being additive and homogenous in its environment.}
\end{proof}
\vspace{1em}

\subsection{Lemma (Compiler is additive in its environment)}
\t{Where $\limp{\p{\D}{e}{\ty}}{\sv{}\{\nb{x}\map\vt{1}\}}{\et{1}}$ and $\limp{\p{\D}{e}{\ty}}{\sv{}\{\nb{x}\map\vt{2}\}}{\et{2}}$,}
$$ \limp{\p{\D}{e}{\ty}}{\sv{}\{\nb{x}\map \vt1+\vt2\}}{\et1+\et{2}}$$
\begin{proof}\leavevmode\\
\t{Consider each inductive case of $\limp{\p{\D}{e}{\ty}}{\sv1}{\et{1}}$,
\begin{itemize}[label=$\triangleright$]
\item \textbf{Case} $\limp{\p{\bind{x}{\ty}}  {x}  {\ty}} {\{\nb{x}\map \vt{1}\}} {\vt{1}}$\\[.25em]
	Because compiler maps to neurons $\limp{\p{\bind{x}{\ty}}  {x}  {\ty}} {\{\nb{x}\map \vt{1}\}} {\vt{1}} \in \c{\nb{\ty}}$  \\
	Because compiler maps to neurons $\limp{\p{\bind{x}{\ty}}  {x}  {\ty}} {\{\nb{x}\map \vt{2}\}} {\vt{2}}\in\c{\nb{\ty}}$\\
	By compiler $\limp{\p{\bind{x}{\ty}}  {x}  {\ty}} {\{\nb{x}\map \vt1+\vt{2}\}} {\vt1 +\vt{2}}$
\item \textbf{Case} $\limp{\p{\emp} {\tt} \Bool}{\emp}{\begin{bmatrix} 1 &\!\!\! 0\end{bmatrix}^\tr}$\\[.25em]
	Vacuous because environment must be non-empty
\item \textbf{Case} $\limp{\p{\emp} {\ff} \Bool}{\emp}{\begin{bmatrix} 0 &\!\!\! 1\end{bmatrix}^\tr}$\\[.25em]
	Vacuous because environment must be non-empty
\item \textbf{Case} $\limp{\p{\D,\bind{x}{\ty_{11}}} {\lam{y}{e}} {\ty_1 \lto \ty_2}}{\sv{}\{\nb{x} \map \vt1\}}{ \Mat }$\\[.25em]
	Because compiler maps to neurons,\\
    $\limp{\p{\D,\bind{x}{\ty_{11}}} {\lam{y}{e}} {\ty_1 \lto \ty_2}}{\sv{}\{\nb{x} \map \vt1\}}{ \Mat } \in\c{\nb{\ty_1 \lto \ty_2}}$\\
	Because compiler maps to neurons,\\
    $\limp{\p{\D,\bind{x}{\ty_{11}}} {\lam{y}{e}} {\ty_1 \lto \ty_2}}{\sv{}\{\nb{x} \map \vt2\}}{ \t{[}\beta_{ij}\t{]}}\in\c{\nb{\ty_1\lto\ty_2}}$\\
	By inversion on compiler $\limp{\p{\D,\bind{x}{\ty_{11}},\bind{y}{\ty_1}}{e}{\ty_2}}{\sv{}\{\nb{x} \map \vt1, \nb{y} \map \bin{j}\}}{\sum_i\alpha_{ij}\.\bout{i}}$\\
	Because compiler maps to neurons $\rd{\sum_i\alpha_{ij}\.\bout{i}} \in \c{\nb{\ty_2}}$ and $\rd{\sum_i\beta_{ij}\.\bout{i}} \in \c{\nb{\ty_2}}$\\
	By inversion on compiler $\limp{\p{\D,\bind{x}{\ty_{11}},\bind{y}{\ty_1}}{e}{\ty_2}}{\sv{}\{\nb{x} \map \vt2, \nb{y} \map \bin{j}\}}{\sum_i\beta_{ij}\.\bout{i}}$\\
	Because contexts permit exchange $\limp{\p{\D,\bind{y}{\ty_1},\bind{x}{\ty_{11}}}{e}{\ty_2}}{\sv{}\{\nb{y} \map \bin{j}, \nb{x} \map \vt1\}}{\sum_i\alpha_{ij}\.\bout{i}}$\\
	Because contexts permit exchange $\limp{\p{\D,\bind{y}{\ty_1},\bind{x}{\ty_{11}}}{e}{\ty_2}}{\sv{}\{\nb{y} \map \bin{j}, \nb{x} \map \vt2\}}{\sum_i\beta_{ij}\.\bout{i}}$\\
	By induction, \\
    $\limp{\p{\D,\bind{y}{\ty_1},\bind{x}{\ty_{11}}}{e}{\ty_2}}{\sv{}\{\nb{y} \map \bin{j}, \nb{x} \map \vt1 + \vt2\}}{\sum_i\alpha_{ij}\.\bout{i}+\sum_i\beta_{ij}\.\bout{i}}=\rd{\sum_i(\alpha_{ij}+\beta_{ij})\.\bout{i}}$\\
	Because contexts permit exchange,\\
    $\limp{\p{\D,\bind{x}{\ty_{11}},\bind{y}{\ty_1}}{e}{\ty_2}}{\sv{}\{\nb{x} \map \vt1 + \vt2, \nb{y} \map \bin{j}\}}{\rd{\sum_i(\alpha_{ij}+\beta_{ij})\.\bout{i}}}$\\
	By compiler $\limp{\p{\D,\bind{x}{\ty_{11}}} {\lam{y}{e}} {\ty_1 \lto \ty_2}}{\sv{}\{\nb{x} \map \vt1+\vt2\}}{ \t{[}\alpha_{ij}+\beta_{ij}\t{]}} = \rd{\Mat +\t{[}\beta_{ij}\t{]}}$ 
\item \textbf{Case} $\limp{\p{\D_1 \o \D_2} {e_1e_2} {\ty_2}}{(\sv{1} \o \sv{2})\{\nb{x} \map \vt1\}}{\et{11}(\et{12})}=\rd{\t{[}\alpha^1_{ij}\t{]}(\beta^1_j)}$ \\[.25em]
	\textbf{Show} $\limp{\p{\D_1 \o \D_2} {e_1e_2} {\ty_2}}{(\sv{1} \o \sv{2})\{\nb{x} \map \vt1+\vt2\}}{\t{[}\alpha^1_{ij}\t{]}(\beta^1_j) + \t{[}\alpha^2_{ij}\t{]}(\beta^2_j)}$\\
	\textbf{Show} $\limp{\p{\D_1 \o \D_2} {e_1e_2} {\ty_2}}{(\sv{1} \o \sv{2})\{\nb{x} \map \vt1+\vt2\}}{\rd{\sum_i\sum_j(\alpha^1_{ij}\.\beta^1_j+\alpha^2_{ij}\.\beta_j^2)\.\bout{i}}}$\\
	Because compiler maps to neurons,\\ 
    $\limp{\p{\D_1 \o \D_2} {e_1e_2} {\ty_2}}{(\sv{1} \o \sv{2})\{\nb{x} \map \vt1\}}{\t{[}\alpha^1_{ij}\t{]}(\beta^1_j)}\in\c{\nb{\ty_2}}$\\
	Because compiler maps to neurons,\\
    $\limp{\p{\D_1 \o \D_2} {e_1e_2} {\ty_2}}{(\sv{1} \o \sv{2})\{\nb{x} \map \vt2\}}{\t{[}\alpha^2_{ij}\t{]}(\beta^2_j)}\in\c{\nb{\ty_2}}$\\
	Proceed by cases on whether $\nb{x}$ occurs free in $\nb{e_1}$ or $\nb{e_2}$
    \begin{itemize}[label=$\triangleright$]
	\item \textbf{Case} $\nb{x} \in \nb{\t{fv}}(\nb{e_1})$\\[.25em]
		By inversion on compiler and because it maps to neurons,\\
        $\limp{\p{\D_1}{e_1}{\ty_1\lto\ty_2}}{\sv{1}\{\nb{x} \map \vt1\}}{\t{[}\alpha^1_{ij}\t{]}}\in\c{\nb{\ty_1 \lto \ty_2}}$ \\
		By inversion on compiler and because it maps to neurons,\\ 
        $\limp{\p{\D_2}{e_2}{\ty_1}}{\sv{2}}{\beta^1_j}\in\c{\nb{\ty_1}}$\\
		By inversion on compiler and because it maps to neurons,\\ 
        $\limp{\p{\D_1}{e_1}{\ty_1\lto\ty_2}}{\sv{1}\{\nb{x} \map \vt2\}}{\t{[}\alpha^2_{ij}\t{]}}\in\c{\nb{\ty_1\lto\ty_2}}$ \\
		By inversion on compiler and because it maps to neurons,\\
        $\limp{\p{\D_2}{e_2}{\ty_1}}{\sv{2}}{\beta^2_j}=\rd{\beta_j^1}\in\c{\nb{\ty_1}}$ \\
		By induction,\\ 
        $\limp{\p{\D_1}{e_1}{\ty_1\lto\ty_2}}{\sv{1}\{\nb{x} \map \vt1+\vt2\}}{\t{[}\alpha^1_{ij}\t{]}+\t{[}\alpha^2_{ij}\t{]}}=\rd{\t{[}\alpha^1_{ij}+\alpha^2_{ij}\t{]}}$ \\
		By compiler,\\ 
        $\limp{\p{\D_1 \o \D_2} {e_1e_2} {\ty_2}}{(\sv{1} \o \sv{2})\{\nb{x} \map \vt1+\vt2\}}{\t{[}\alpha^1_{ij}+\alpha^2_{ij}\t{]}(\beta^1_{j})}=\rd{\sum_i\sum_j(\alpha^1_{ij}\.\beta_j^1+\alpha^2_{ij}\.\beta_j^1)\.\bout{i}}$
	\item \textbf{Case} $\nb{x} \in \nb{\t{fv}}(\nb{e_2})$\\[.25em]
		By inversion on compiler and because it maps to neurons,\\
        $\limp{\p{\D_1}{e_1}{\ty_1\lto\ty_2}}{\sv{1}}{\t{[}\alpha^1_{ij}\t{]}} \in \c{\nb{\ty_1\lto\ty_2}}$\\
		By inversion on compiler and because it maps to neurons,\\ $\limp{\p{\D_2}{e_2}{\ty_1}}{\sv{2}\{\nb{x} \map \vt1\}}{\beta^1_j}\in\c{\nb{\ty_1}}$ \\
		By inversion on compiler and because it maps to neurons,\\ $\limp{\p{\D_1}{e_1}{\ty_1\lto\ty_2}}{\sv{1}}{\t{[}\alpha^2_{ij}\t{]}}=\rd{\t{[}\alpha^1_{ij}\t{]}}\in\c{\nb{\ty_1\lto\ty_2}}$\\ 
		By inversion on compiler and because it maps to neurons,\\ 
        $\limp{\p{\D_2}{e_2}{\ty_1}}{\sv{2}\{\nb{x} \map \vt2\}}{\beta^2_j}\in\c{\nb{\ty_1}}$ \\
		By induction,\\
        $\limp{\p{\D_2}{e_2}{\ty_1}}{\sv{2}\{\nb{x} \map \vt1 + \vt2\}}{\beta^1_j+\beta^2_j}$ \\
		By compiler,\\
        $\limp{\p{\D_1 \o \D_2} {e_1e_2} {\ty_2}}{(\sv{1} \o \sv{2})\{\nb{x} \map \vt1+\vt2\}}{\t{[}\alpha^1_{ij}\t{]}(\beta^1_{j}+\beta^2_{j})}=\rd{\sum_i\sum_j(\alpha^1_{ij}\.\beta_j^1+\alpha^1_{ij}\.\beta_j^2)\.\bout{i}}$
    \end{itemize}
\item \textbf{Case} $\limp{\p{\D_1 \o \D_2} {\ite{e_1}{e_2}{e_3}} {\ty}}{(\sv1 \o \sv2)\{\nb{x} \map \vt1\}}{\alpha_1\.\et{2}+\alpha_2\.\et{3}}$\\[.25em] 
	Proceed by cases on whether $\nb{x} \in \nb{\t{fv}}(\nb{e_1})$ or ($\nb{x} \in \nb{\t{fv}}(\nb{e_2})$ and $\nb{x} \in \nb{\t{fv}}(\nb{e_3})$)
    \begin{itemize}[label=$\triangleright$]
    \item \textbf{Case} $\nb{x} \in \nb{\t{fv}}(\nb{e_1})$\\[.25em]
		  \textbf{Show},\\ $\limp{\p{\D_1 \o \D_2} {\ite{e_1}{e_2}{e_3}} {\ty}}{(\sv1 \o \sv2)\{\nb{x} \map \vt1+\vt2\}}{(\alpha_1+\beta_1)\.\et{21}+(\alpha_2+\beta_2)\.\et{31}}$ \\
		By inversion on compiler and because it maps to neurons,\\ 
        $\limp{\p{\D_1} {e_1} {\Bool}}{\sv1\{\nb{x} \map \vt1\}}{ \begin{bmatrix}\alpha_1 &\!\!\! \alpha_2\end{bmatrix}^\tr }\in\c{\nb{\Bool}}$ \\
		By inversion on compiler and because it maps to neurons,\\
        $\limp{\p{\D_1} {e_1} {\Bool}}{\sv1\{\nb{x} \map \vt2\}}{ \begin{bmatrix}\beta_1 &\!\!\! \beta_2\end{bmatrix}^\tr }\in\c{\nb{\Bool}}$ \\
		By induction $\limp{\p{\D_1} {e_1} {\Bool}}{\sv1\{\nb{x} \map \vt1+ \vt2\}}{ \begin{bmatrix}\alpha_1+\beta_1 &\!\!\! \alpha_2+\beta_2\end{bmatrix}^\tr }$  \\
		By inversion on compiler and because it maps to neurons,\\
        $\limp{\p{\D_2} {e_2} {\ty}}{\sv2}{ \et{21}}\in\c{\nb{\ty}}$ \\
		By inversion on compiler and because it maps to neurons,\\ 
        $\limp{\p{\D_2} {e_3} {\ty}}{\sv2}{ \et{31}}\in\c{\nb{\ty}}$ \\
		By compiler,\\ 
        $\limp{\p{\D_1 \o \D_2} {\ite{e_1}{e_2}{e_3}} {\ty}}{(\sv1 \o \sv2)\{\nb{x} \map \vt1+\vt2\}}{(\alpha_1+\beta_1)\.\et{21}+(\alpha_2+\beta_2)\.\et{31}}$
	\item \textbf{Case} $\nb{x} \in \nb{\t{fv}}(\nb{e_2})$ and $\nb{x} \in \nb{\t{fv}}(\nb{e_3})$\\[.25em]
		  \textbf{Show},\\ 
          $\limp{\p{\D_1 \o \D_2} {\ite{e_1}{e_2}{e_3}} {\ty}}{(\sv1 \o \sv2)\{\nb{x} \map \vt1+\vt2\}}{\alpha_1\.(\et{21}+\et{22})+\alpha_2\.(\et{31}+\et{32})}$\\
		By inversion on compiler and because it maps to neurons,\\
        $\limp{\p{\D_1} {e_1} {\Bool}}{\sv1}{ \begin{bmatrix}\alpha_1 &\!\!\! \alpha_2\end{bmatrix}^\tr }\in\c{\nb{\Bool}}$ \\
		By inversion on compiler and because it maps to neurons,\\
        $\limp{\p{\D_2} {e_2} {\ty}}{\sv2\{\nb{x}\map \vt1\}}{ \et{21}}\in\c{\nb{\ty}}$ \\
		By inversion on compiler and because it maps to neurons,\\
        $\limp{\p{\D_2} {e_2} {\ty}}{\sv2\{\nb{x}\map \vt2\}}{ \et{22}}\in\c{\nb{\ty}}$ \\
		By induction $\limp{\p{\D_2} {e_2} {\ty}}{\sv2\{\nb{x}\map \vt1+\vt2\}}{ \et{21}+\et{22}}$\\
		By inversion on compiler and because it maps to neurons,\\ $\limp{\p{\D_2} {e_3} {\ty}}{\sv2\{\nb{x}\map \vt1\}}{ \et{31}}\in\c{\nb{\ty}}$ \\
		By inversion on compiler and because it maps to neurons,\\ $\limp{\p{\D_2} {e_3} {\ty}}{\sv2\{\nb{x}\map \vt2\}}{ \et{32}}\in\c{\nb{\ty}}$ \\
		By induction $\limp{\p{\D_2} {e_3} {\ty}}{\sv2\{\nb{x}\map \vt1+\vt2\}}{ \et{31}+\et{32}}$	\\
		By compiler,\\ $\limp{\p{\D_1 \o \D_2} {\ite{e_1}{e_2}{e_3}} {\ty}}{(\sv1 \o \sv2)\{\nb{x} \map \vt1+\vt2\}}{\alpha_1\.(\et{21}+\et{22})+\alpha_2\.(\et{31}+\et{32})}$ 
    \end{itemize}
\end{itemize}}
\end{proof}
\vspace{1em}

\subsection{Lemma (Compiler is homogenous in its environment}
\t{Where $\limp{\p{\D}{e}{\ty}}{\sv{}\{\nb{x} \map \vt{}\}}{\et{}}$
$$\limp{\p{\D}{e}{\ty}}{\sv{}\{\nb{x} \map \alpha\.\vt{}\}}{\alpha\.\et{}}$$}
\begin{proof}\leavevmode\\
\t{Consider each inductive case of $\limp{\p{\D}{e}{\ty}}{\sv{}}{\et{}}$,
\begin{itemize}[label=$\triangleright$]
\item \textbf{Case} $\limp{\p{\bind{x}{\ty}}  {x}  {\ty}} {\{\nb{x}\map \vt{}\}} {\vt{}}$ \\[.25em]
	By compiler $\limp{\p{\bind{x}{\ty}}  {x}  {\ty}} {\{\nb{x}\map \alpha \.\vt{}\}} {\alpha\.\vt{}}$
\item \textbf{Case} $\limp{\p{\emp} {\tt} \Bool}{\emp}{\begin{bmatrix} 1 &\!\!\! 0\end{bmatrix}^\tr}$\\[.25em]
	Vacuous because environment must be non-empty
\item \textbf{Case} $\limp{\p{\emp} {\ff} \Bool}{\emp}{\begin{bmatrix} 0 &\!\!\! 1\end{bmatrix}^\tr}$\\[.25em]
	Vacuous because environment must be non-empty
\item \textbf{Case} $\limp{\p{\D,\bind{x}{\ty_{11}}} {\lam{y}{e}} {\ty_1 \lto \ty_2}}{\sv{}\{\nb{x} \map \vt{}\}}{ \Mat }$ \\[.25em]
	By inversion on compiler $\limp{\p{\D,\bind{x}{\ty_{11}},\bind{y}{\ty_1}}{e}{\ty_2}}{\sv{}\{\nb{x} \map \vt{}, \nb{y} \map \bin{j}\}}{\sum_i\alpha_{ij}\.\bout{i}}$\\
	Because contexts permit exchange $\limp{\p{\D,\bind{y}{\ty_1},\bind{x}{\ty_{11}}}{e}{\ty_2}}{\sv{}\{\nb{y} \map \bin{j}, \nb{x} \map \vt{}\}}{\sum_i\alpha_{ij}\.\bout{i}}$\\
	By induction $\limp{\p{\D,\bind{y}{\ty_1},\bind{x}{\ty_{11}}}{e}{\ty_2}}{\sv{}\{\nb{y} \map \bin{j}, \nb{x} \map \bar{\alpha}\.\vt{}\}}{\sum_i\bar{\alpha}\.\alpha_{ij}\.\bout{i}}$\\
	Because contexts permit exchange,\\ 
    $\limp{\p{\D,\bind{x}{\ty_{11}}, \bind{y}{\ty_1}}{e}{\ty_2}}{\sv{}\{\nb{x} \map \bar{\alpha}\.\vt{}, \nb{y} \map \bin{j}\}}{\sum_i\bar{\alpha}\.\alpha_{ij}\.\bout{i}}$\\
	By compiler $\limp{\p{\D,\bind{x}{\ty_{11}}} {\lam{y}{e}} {\ty_1 \lto \ty_2}}{\sv{}\{\nb{x} \map \bar{\alpha}\.\vt{}\}}{ \t{[}\bar{\alpha}\.\alpha_{ij}\t{]}}=\rd{\bar{\alpha}\.\Mat}$ 
\item \textbf{Case} $\limp{\p{\D_1 \o \D_2} {e_1e_2} {\ty_2}}{(\sv{1} \o \sv{2})\{\nb{x} \map \vt{}\}}{\et{11}(\et{12})}=\rd{\t{[}\alpha_{ij}\t{]}(\beta_j)}$\\[.25em]
	Proceed by cases on whether $\nb{x} \in \nb{\t{fv}}(\nb{e_1})$ or $\nb{x} \in \nb{\t{fv}}(\nb{e_2})$
    \begin{itemize}[label=$\triangleright$]
	\item \textbf{Case} $\nb{x} \in \nb{\t{fv}}(\nb{e_1})$\\[.25em]
		By inversion on compiler $\limp{\p{\D_1} {e_1} {\ty_1 \lto \ty_2}}{\sv1\{\nb{x} \map\vt{}\}}{\Mat}$ \\
		By inversion on compiler $\limp{ \p{\D_2} {e_2} {\ty_1}}{\sv2}{\beta_j}$\\
		By induction $\limp{\p{\D_1} {e_1} {\ty_1 \lto \ty_2}}{\sv1\{\nb{x} \map \bar{\alpha}\.\vt{}\}}{\bar{\alpha} \.\Mat}$ \\
		By compiler $\limp{\p{\D_1 \o \D_2} {e_1e_2} {\ty_2}}{(\sv{1} \o \sv{2})\{\nb{x} \map \vt{}\}}{(\bar{\alpha}\.\t{[}\alpha_{ij}\t{]})(\beta_j)}=\rd{\bar{\alpha}\.\Mat(\beta_j)}$
	\item \textbf{Case} $\nb{x} \in \nb{\t{fv}}(\nb{e_2})$ \\[.25em]
		By inversion on compiler $\limp{\p{\D_1} {e_1} {\ty_1 \lto \ty_2}}{\sv1}{\Mat}$ \\
		By inversion on compiler $\limp{ \p{\D_2} {e_2} {\ty_1}}{\sv2\{\nb{x} \map\vt{}\}}{\beta_j}$\\
		By induction $\limp{\p{\D_2} {e_2} {\ty_1}}{\sv2\{\nb{x} \map \bar{\alpha}\.\vt{}\}}{\bar{\alpha} \.\beta_j}$ \\
		By compiler $\limp{\p{\D_1 \o \D_2} {e_1e_2} {\ty_2}}{(\sv{1} \o \sv{2})\{\nb{x} \map \vt{}\}}{\t{[}\alpha_{ij}\t{]}(\bar{\alpha}\.\beta_j)}=\rd{\bar{\alpha}\.\Mat(\beta_j)}$
    \end{itemize}
\item \textbf{Case} $\limp{\p{\D_1\o\D_2}{\ite{e_1}{e_2}{e_3}}{\ty}}{\sv1\o\sv2}{\alpha_1\.\et2+\alpha_2\.\et3}$ \\[.25em]
	Proceed by cases on whether $\nb{x} \in \nb{\t{fv}}(\nb{e_1})$ or ($\nb{x} \in \nb{\t{fv}}(\nb{e_2})$ and $\nb{x} \in \nb{\t{fv}}(\nb{e_3})$)
    \begin{itemize}[label=$\triangleright$]
    \item \textbf{Case} $\nb{x}\in\nb{\t{fv}}(\nb{e_1})$\\[.25em]
		By inversion on compiler $\limp{\p{\D_1} {e_1} {\Bool}}{\sv1\{\nb{x} \map \vt1\}}{ \begin{bmatrix}\alpha_1 &\!\!\! \alpha_2\end{bmatrix}^\tr }$ \\
		By induction $\limp{\p{\D_1} {e_1} {\Bool}}{\sv1\{\nb{x} \map \bar{\alpha}\.\vt1\}}{\bar{\alpha}\. \begin{bmatrix}\alpha_1 &\!\!\! \alpha_2\end{bmatrix}^\tr }=\rd{\begin{bmatrix}\bar{\alpha}\.\alpha_1 &\!\!\! \bar{\alpha}\.\alpha_2\end{bmatrix}^\tr}$ \\
		By inversion on compiler $\limp{\p{\D_2} {e_2} {\ty}}{\sv2}{ \et{2}}$\\
		By inversion on compiler $\limp{\p{\D_2} {e_3} {\ty}}{\sv2}{ \et{3}}$\\
		By compiler $\limp{\p{\D_1\o\D_2}{\ite{e_1}{e_2}{e_3}}{\ty}}{\sv1\o\sv2}{\bar{\alpha}\.\alpha_1\.\et2+\bar{\alpha}\.\alpha_2\.\et3}$ 
	\item \textbf{Case} $\nb{x} \in \nb{\t{fv}}(\nb{e_2})$ and $\nb{x} \in \nb{\t{fv}}(\nb{e_3})$ \\[.25em]
		By inversion on compiler $\limp{\p{\D_1} {e_1} {\Bool}}{\sv1}{ \begin{bmatrix}\alpha_1 &\!\!\! \alpha_2\end{bmatrix}^\tr }$ \\
		By inversion on compiler $\limp{\p{\D_2} {e_2} {\ty}}{\sv2\{\nb{x} \map \vt2\}}{ \et2 }$ \\
		By inversion on compiler $\limp{\p{\D_2} {e_3} {\ty}}{\sv2\{\nb{x} \map \vt2\}}{ \et3 }$ 
    \end{itemize}
\end{itemize}}
\end{proof}
\vspace{1em}

\subsection{Lemma (Compiler is invariant to closing environments)}
\t{Where $\c{\nb{\D}\-\nb{\d}}=\sv{}$ and $\limp{\p{\D}{e}{\ty}}{\sv{}}{\et{}}$,}
$$\c{\p{\emp}{\d(e)}{\ty}}(\rd{\emp})=\et{}$$
\begin{proof}\leavevmode\\
\t{Consider each inductive case of $\c{\nb{\D}\-\nb{\d}}=\sv{}$,
\begin{itemize}[label=$\triangleright$]
\item \textbf{Case} $\c{\nb{\emp \-\emp}} = \rd{\emp}$\\[.25em]
	By assumption $\limp{\p{\D}{e}{\ty}}{\sv{}}{\et{}}$ \\
	By substitution $\limp{\p{\D}{\emp (e)}{\ty}}{\sv{}}{\et{}}$ 
\item \textbf{Case} $\c{\nb{\D,\bind{x}{\ty_1}}\- \nb{\d\{x \map v\}}}=\rd{\sv{}\{\nb{x} \map \vt{}\}}$ \\[.25em]
	\textbf{Show} $\limp{\p{\emp}{\d\{x \map v\}(e)}{\ty}}{\emp}{\et{}}$\\
	\textbf{Show} $\limp{\p{\emp}{\d(\{x \map v\}(e))}{\ty}}{\emp}{\et{}}$\\
	By assumption $\limp{\p{\D,\bind{x}{\ty_1}}{e}{\ty}}{\sv{}\{\nb{x} \map \vt{}\}}{\et{}}$ \\
	By inversion on compiling environments $\c{\nb{\D\-\d}}=\sv{}$\\ 
	Because the compiler is invariant to partially closing environments,\\
    $\limp{\p{\D}{\{x \map v\}(e)}{\ty}}{\sv{}}{\et{}}$\\
	By induction $\limp{\p{\emp}{\d(\{x \map v\}(e))}{\ty}}{\emp}{\et{}}$
\end{itemize}}
\end{proof}
\vspace{1em}

\subsection{Lemma (Compiler is invariant to partially closing environments)}
\t{Where $\c{\nb{\D,\bind{x}{\ty_1}\-\d\{x \map v\}}}=\rd{\sv{}\{\nb{x} \map \vt{}\}}$, and $\limp{\p{\D,\bind{x}{\ty_1}}{e}{\ty}}{\sv{}\{\nb{x} \map \vt{}\}}{\et{}}$,}
$$\limp{\p{\D}{\{x \map v\}(e)}{\ty}}{\sv{}}{\et{}}$$
\begin{proof}\leavevmode\\
\t{Consider each inductive case of compiler,
\begin{itemize}[label=$\triangleright$]
\item \textbf{Case} $\limp{\p{\bind{x}{\ty}}  {x}  {\ty}} {\{\nb{x}\map \vt{}\}} {\vt{}}$\\[.25em]
	By substitution $\nb{\{x \map v\}(x)}=\nb{v}$\\
	By inversion on compiling environments $\limp{\p{\emp}{v}{\ty}}{\emp}{\vt{}}$ \\
\item \textbf{Case} $\limp{\p{\emp} {\tt} \Bool}{\emp}{\begin{bmatrix} 1 &\!\!\! 0\end{bmatrix}^\tr}$ \\[.25em]
	Vacuous because context can't be empty
\item \textbf{Case} $\limp{\p{\emp} {\ff} \Bool}{\emp}{\begin{bmatrix} 0 &\!\!\! 1\end{bmatrix}^\tr}$\\[.25em]
	Vacuous because context can't be empty
\item \textbf{Case} $\limp{\p{\D,\bind{x}{\ty}} {\lam{y}{e}} {\ty_1 \lto \ty_2}}{\sv{}\{\nb{x} \map \vt{}\}}{ \t{[}\alpha_{ij}\t{]}}$\\[.25em]
    \textbf{Show} $\limp{\p{\D}{\{x \map v\}(\lam{y}{e})}{\ty_1 \lto \ty_2}}{\sv{}}{\Mat}$ \\
    \textbf{Show} $\limp{\p{\D}{\lam{y}{\{x \map v\}(e)}}{\ty_1 \lto\ty_2}}{\sv{}}{\Mat}$ \\
	By inversion on compiler $\limp{ \p{\D,\bind{x}{\ty},\bind{y}{\ty_1}} {e}  {\ty_2}}{\sv{}\{\nb{x} \map \vt{},\nb{y} \map \bin{j}\}}{\sum_i\alpha_{ij}\.\bout{i}}$\\
	Because contexts permit exchange $\limp{ \p{\D,\bind{y}{\ty_1}, \bind{x}{\ty}} {e}  {\ty_2}}{\sv{}\{\nb{y} \map \bin{j}, \nb{x} \map \vt{}\}}{\sum_i\alpha_{ij}\.\bout{i}}$ \\
	By induction $\limp{ \p{\D,\bind{y}{\ty_1}} {\{x \map v\}(e)}  {\ty_2}}{\sv{}\{\nb{y} \map \bin{j}\}}{\sum_i\alpha_{ij}\.\bout{i}}$\\
	By compiler $\limp{\p{\D}{\lam{y}{\{x \map v\}(e)}}{\ty}}{\sv{}}{\Mat}$ 
\item \textbf{Case} $\limp{\p{(\D_1 \o \D_2),\bind{x}{\ty}} {e_1e_2} {\ty_2}}{(\sv1 \o \sv2)\{\nb{x} \map \vt{}\}}{\et1(\et2)}$\\[.25em]
	Proceed by cases on whether $\nb{x} \in \nb{\t{fv}}(\nb{e_1})$ or $\nb{x} \in \nb{\t{fv}}(\nb{e_2})$ 
    \begin{itemize}[label=$\triangleright$]
    \item \textbf{Case} $\nb{x} \in \nb{\t{fv}}(\nb{e_1})$\\[.25em]
		\textbf{Show} $\limp{\p{\D_1 \o \D_2} {\{x \map v\}(e_1)e_2} {\ty_2}}{\sv1 \o \sv2}{\et1(\et2)}$ \\
		By inversion on compiler $\limp{\p{\D_1,\bind{x}{\ty}} {e_1} {\ty_1 \lto \ty_2}}{\sv1\{\nb{x} \map v\}}{\et1}$\\
		By inversion on compiler $\limp{ \p{\D_2} {e_2} {\ty_1}}{\sv2}{\et2}$ \\
		By induction $\limp{\p{\D_1} {\{x \map v\}(e_1)} {\ty_1 \lto \ty_2}}{\sv1}{\et1}$ \\
		By compiler $\limp{\p{\D_1 \o \D_2} {\{x \map v\}(e_1)e_2} {\ty_2}}{\sv1 \o \sv2}{\et1(\et2)}$
	\item \textbf{Case} $\nb{x} \in \nb{\t{fv}}(\nb{e_2})$\\[.25em]
		Similar to previous case but inducting on $\nb{e_2}$
    \end{itemize}
\item \textbf{Case} $\limp{\p{(\D_1 \o \D_2),\bind{x}{\ty_1}} {\ite{e_1}{e_2}{e_3}} {\ty}}{(\sv1 \o \sv2)\{\nb{x} \map \vt1\}}{\alpha_1\.\et{2}+\alpha_2\.\et{3}}$\\[.25em]
	Proceed by cases on whether $\nb{x} \in \nb{\t{fv}}(\nb{e_1})$ or ($\nb{x} \in \nb{\t{fv}}(\nb{e_2})$ and $\nb{x} \in \nb{\t{fv}}(\nb{e_3})$)
    \begin{itemize}[label=$\triangleright$]
    \item \textbf{Case} $\nb{x} \in \nb{\t{fv}}(\nb{e_1})$\\[.25em]
		By inversion on compiler $\limp{\p{\D_1,\bind{x}{\ty_1}} {e_1} {\Bool}}{\sv1\{\nb{x} \map \vt1\}}{ \begin{bmatrix}\alpha_1 &\!\!\! \alpha_2\end{bmatrix}^\tr }$ \\
		By inversion on compiler $\limp{\p{\D_2} {e_2} {\ty}}{\sv2}{ \et2}$ \\
		By inversion on compiler $\limp{\p{\D_2} {e_3} {\ty}}{\sv2}{ \et3}$ \\
		By induction $\limp{\p{\D_1} {\{x \map v_1\}(e_1)} {\Bool}}{\sv1}{ \begin{bmatrix}\alpha_1 &\!\!\! \alpha_2\end{bmatrix}^\tr }$\\
		By compiler $\limp{\p{\D_1 \o \D_2} {\ite{e_1}{e_2}{e_3}} {\ty}}{\sv1 \o \sv2}{\alpha_1\.\et{2}+\alpha_2\.\et{3}}$
	\item \textbf{Case} $\nb{x} \in \nb{\t{fv}}(\nb{e_2})$ and $\nb{x} \in \nb{\t{fv}}(\nb{e_3})$\\[.25em]
		Similar to previous case but inducting on $\nb{e_2}$ and $\nb{e_3}$ 
    \end{itemize}
\end{itemize}}
\end{proof}
\vspace{1em}




\clearpage
\section{Experiments}
\label{app:experiments}

The following experiments are variants of the conditional classification task (Section \ref{sec:exp2}). In this appendix we will briefly state their results, pointing out important differences but assuming you are familiar with the previously discussed conditional classification task.

\subsection{Conditional classification: XOR}

\subsubsection{Task}

\begin{wrapfigure}{ht}{0.35\textwidth}
    \centering
    \vspace{-1.4em}
    \includegraphics[width=.3\textwidth]{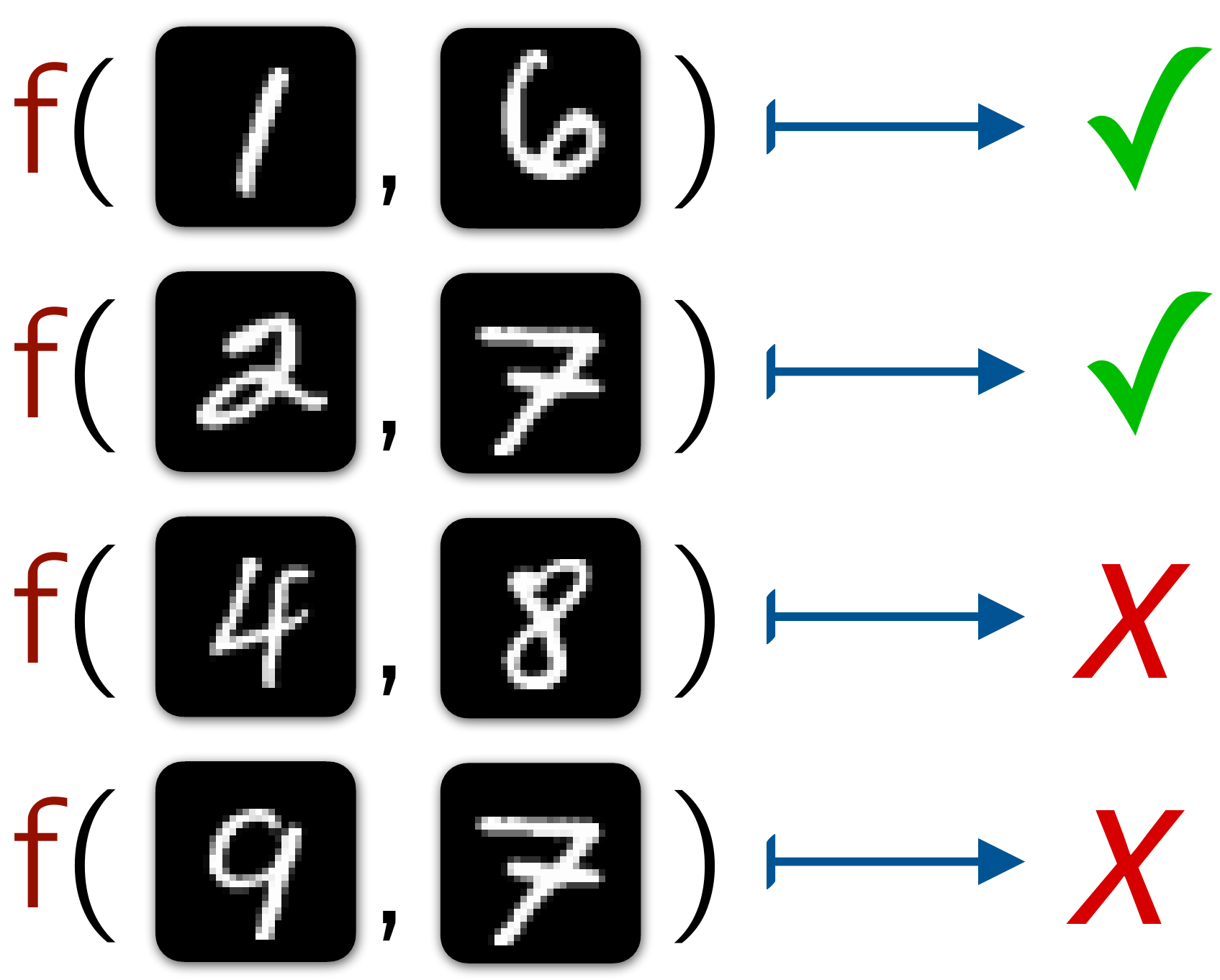}
    \vspace{.6em}
  \caption{XOR task}
  \label{fig:task2xor}
  \vspace{-2.5em}
\end{wrapfigure}

Our second experiment is the conditional classification task in Fig. \ref{fig:task2}. We need a function which conditionally classifies two input image $x_1$ and $x_2$:
\setlist[itemize]{topsep=4pt, itemsep=1pt, parsep=2pt, leftmargin=1.5em, rightmargin=1.5em}
\begin{itemize}[label=$\triangleright$]
\item If $x_1$ is even and $x_2$ is odd, return \textcolor{ForestGreen}{\cmark}
\item If $x_1$ is odd and $x_2$ is even, return \textcolor{ForestGreen}{\cmark}
\item Otherwise, return \textcolor{red}{\xmark}
\end{itemize}

\subsubsection{Models}
The direct model $\t{(D)}$ is derived from compiling the following program, which computes XOR of its inputs. The Church encoded model $\t{(C)}$ is the Church encoding of this program.
\vspace{-.3em}
\[
\p{\bind{x}{\Bool}, \bind{y}{\Bool}}{\ite{\ul{x}}{(\ite{\ul{y}}{\ul{\ff}}{\ul{\tt}})}{(\ite{\ul{y}}{\ul{\tt}}{\ul{\ff}})}}{\Bool}
\]

\begin{figure}[b]
  \includegraphics[scale=.37]{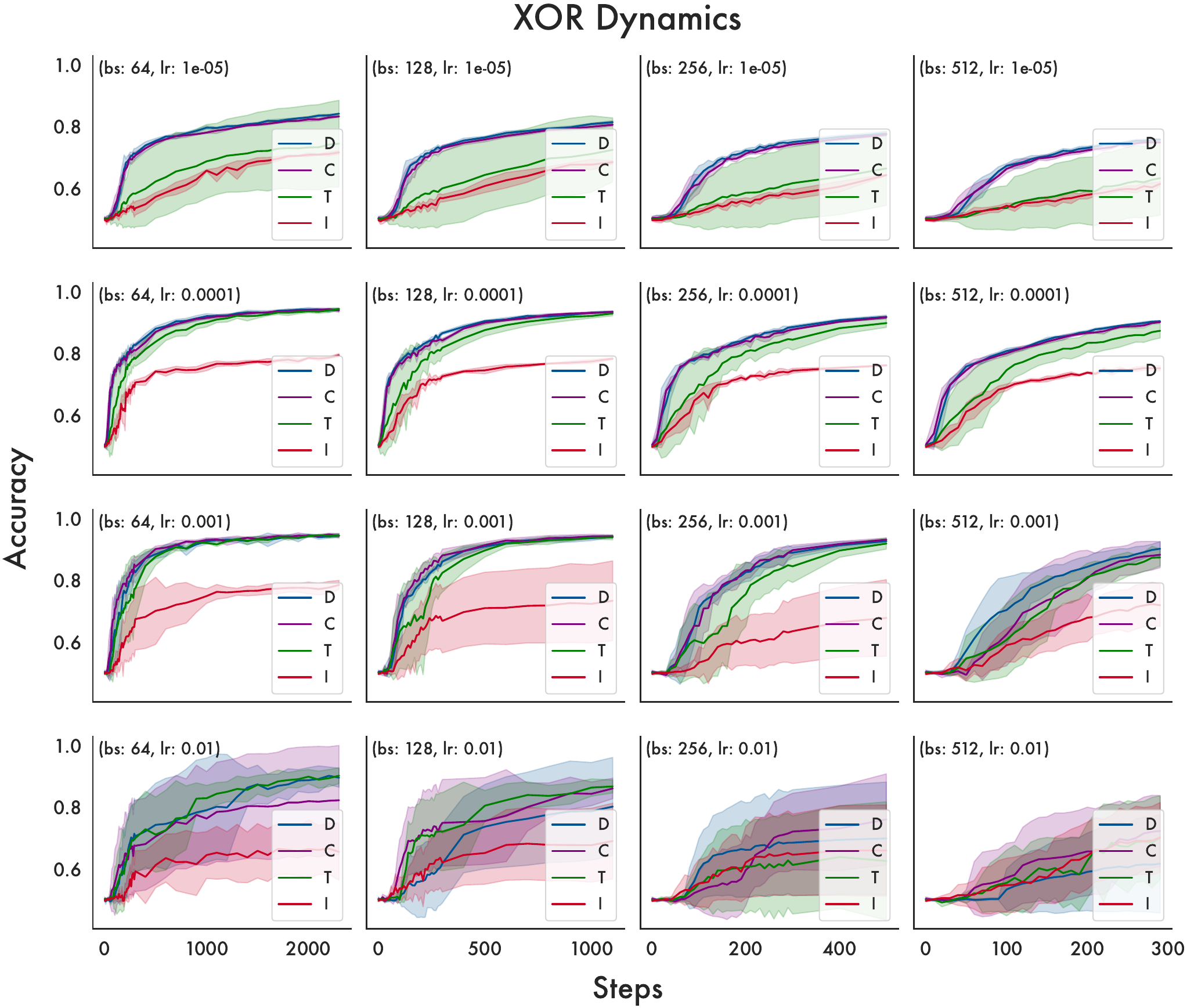}
  \centering
  \vspace{.6em}    
  \caption{Summary of dynamics across configurations}
\end{figure}

\clearpage

\subsection{Conditional classification: AND}

\subsubsection{Task}

\begin{wrapfigure}{ht}{0.35\textwidth}
    \centering
    \vspace{-1.4em}
    \includegraphics[width=.3\textwidth]{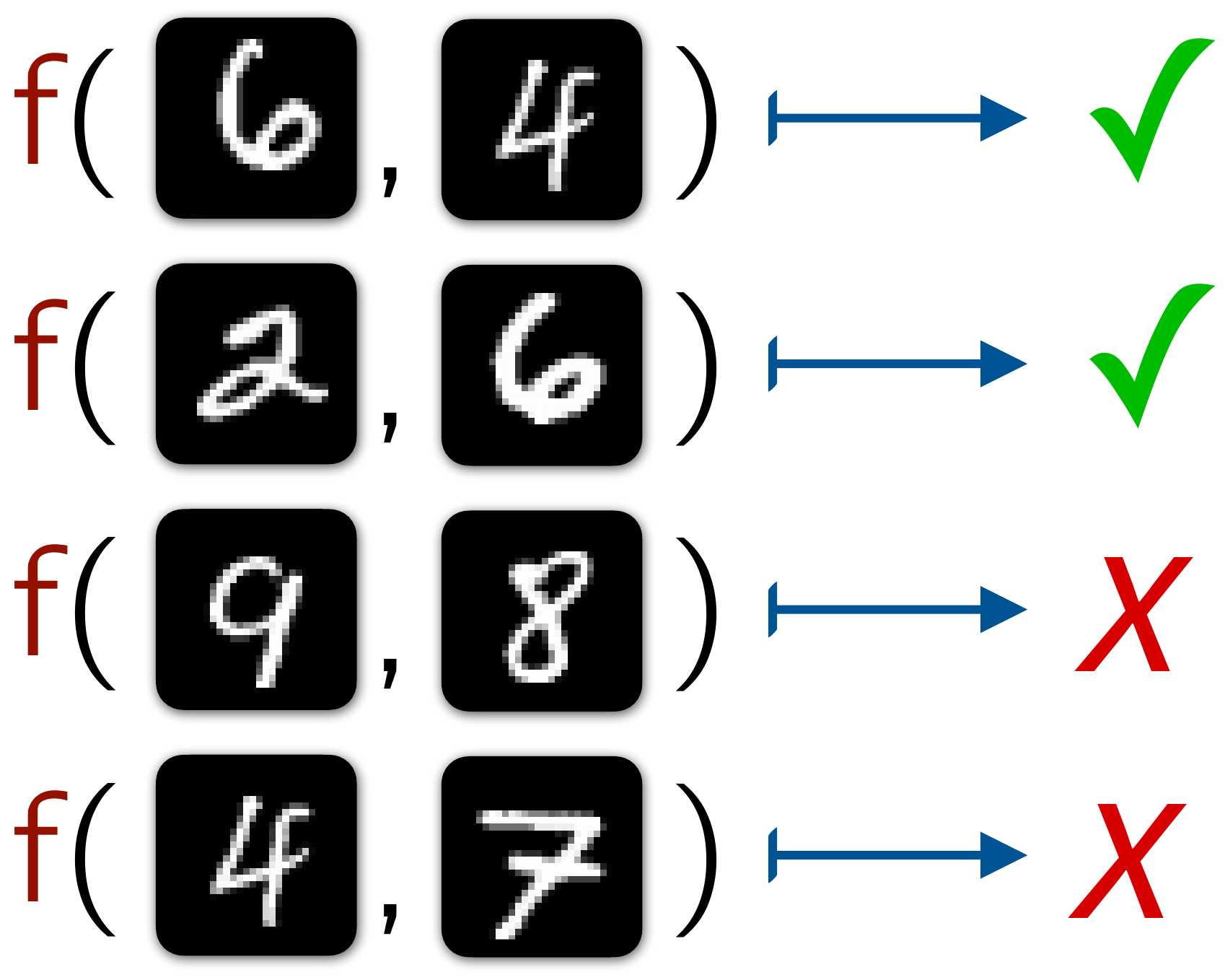}
    \vspace{.6em}
  \caption{AND task}
  \label{fig:task2and}
  \vspace{-2.5em}
\end{wrapfigure}

Our second experiment is the conditional classification task in Fig. \ref{fig:task2}. We need a function which conditionally classifies two input image $x_1$ and $x_2$:
\setlist[itemize]{topsep=4pt, itemsep=1pt, parsep=2pt, leftmargin=1.5em, rightmargin=1.5em}
\begin{itemize}[label=$\triangleright$]
\item If $x_1$ is even and $x_2$ is even, return \textcolor{ForestGreen}{\cmark}
\item Otherwise, return \textcolor{red}{\xmark}
\end{itemize}

\subsubsection{Models}
The direct model $\t{(D)}$ is derived from compiling the following program, which computes AND of its inputs. The Church encoded model $\t{(C)}$ is the Church encoding of this program.

\vspace{.5em}
\[
\p{\bind{x}{\Bool}, \bind{y}{\Bool}}{\ite{\ul{x}}{(\ite{\ul{y}}{\ul{\tt}}{\ul{\ff}})}{(\ite{\ul{y}}{\ul{\ff}}{\ul{\ff}})}}{\Bool}
\]
\begin{figure}[h]
  \includegraphics[scale=.37]{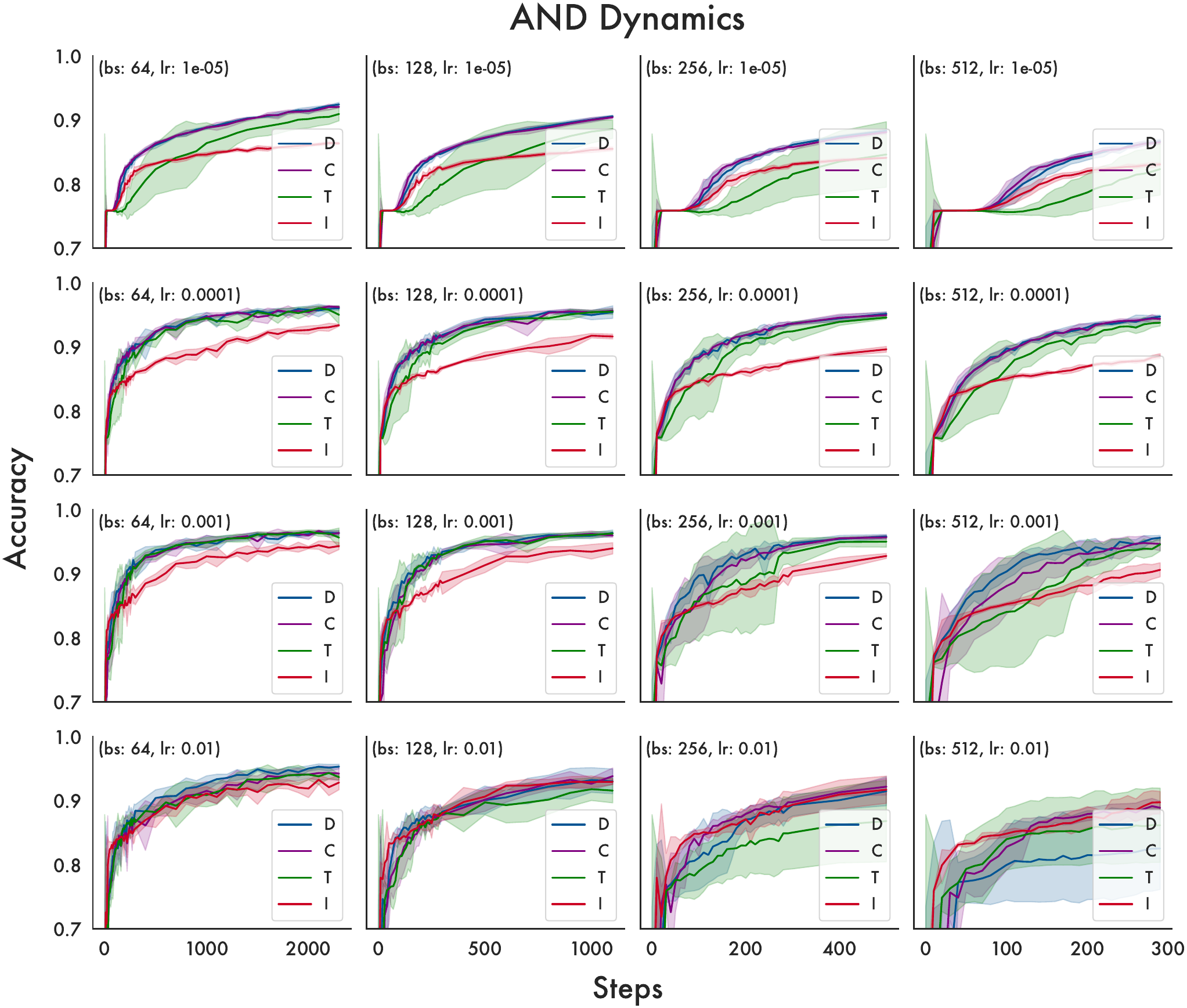}
  \centering
  \vspace{.6em}    
  \caption{Summary of dynamics across configurations}
\end{figure}

\clearpage

\subsection{Conditional classification: OR}

\subsubsection{Task}

\begin{wrapfigure}{ht}{0.35\textwidth}
    \centering
    \vspace{-1.4em}
    \includegraphics[width=.3\textwidth]{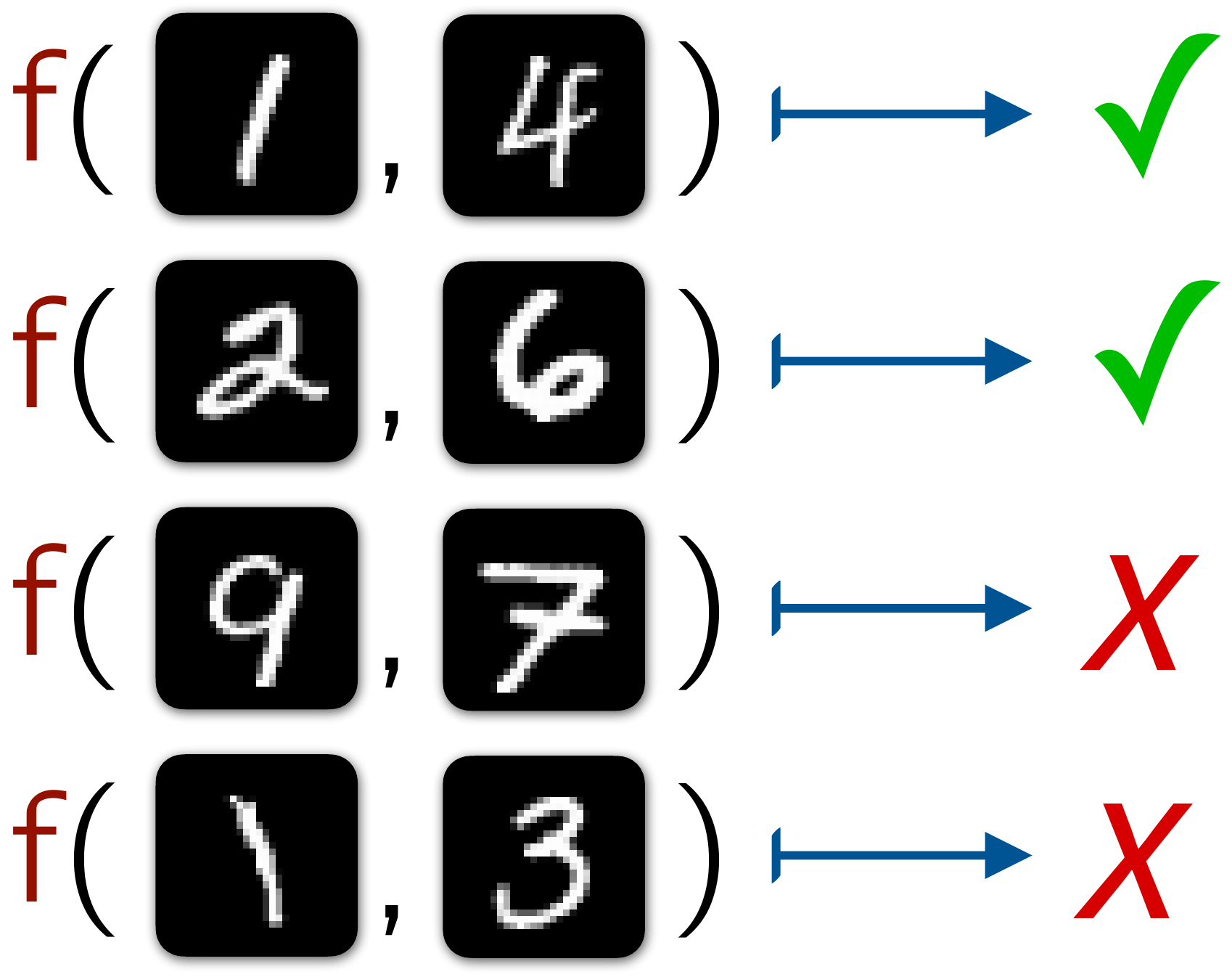}
    \vspace{.6em}
  \caption{OR Task}
  \label{fig:task2or}
  \vspace{-2.5em}
\end{wrapfigure}

Our second experiment is the conditional classification task in Fig. \ref{fig:task2}. We need a function which conditionally classifies two input image $x_1$ and $x_2$:
\setlist[itemize]{topsep=4pt, itemsep=1pt, parsep=2pt, leftmargin=1.5em, rightmargin=1.5em}
\begin{itemize}[label=$\triangleright$]
\item If $x_1$ is even or $x_2$ is odd, return \textcolor{ForestGreen}{\cmark}
\item Otherwise, return \textcolor{red}{\xmark}
\end{itemize}

\subsubsection{Models}
The direct model $\t{(D)}$ is derived from compiling the following program, which computes OR of its inputs. The Church encoded model $\t{(C)}$ is the Church encoding of this program.
\vspace{.5em}

\[
\p{\bind{x}{\Bool}, \bind{y}{\Bool}}{\ite{\ul{x}}{(\ite{\ul{y}}{\ul{\tt}}{\ul{\tt}})}{(\ite{\ul{y}}{\ul{\tt}}{\ul{\ff}})}}{\Bool}
\]

\begin{figure}[h]
  \includegraphics[scale=.37]{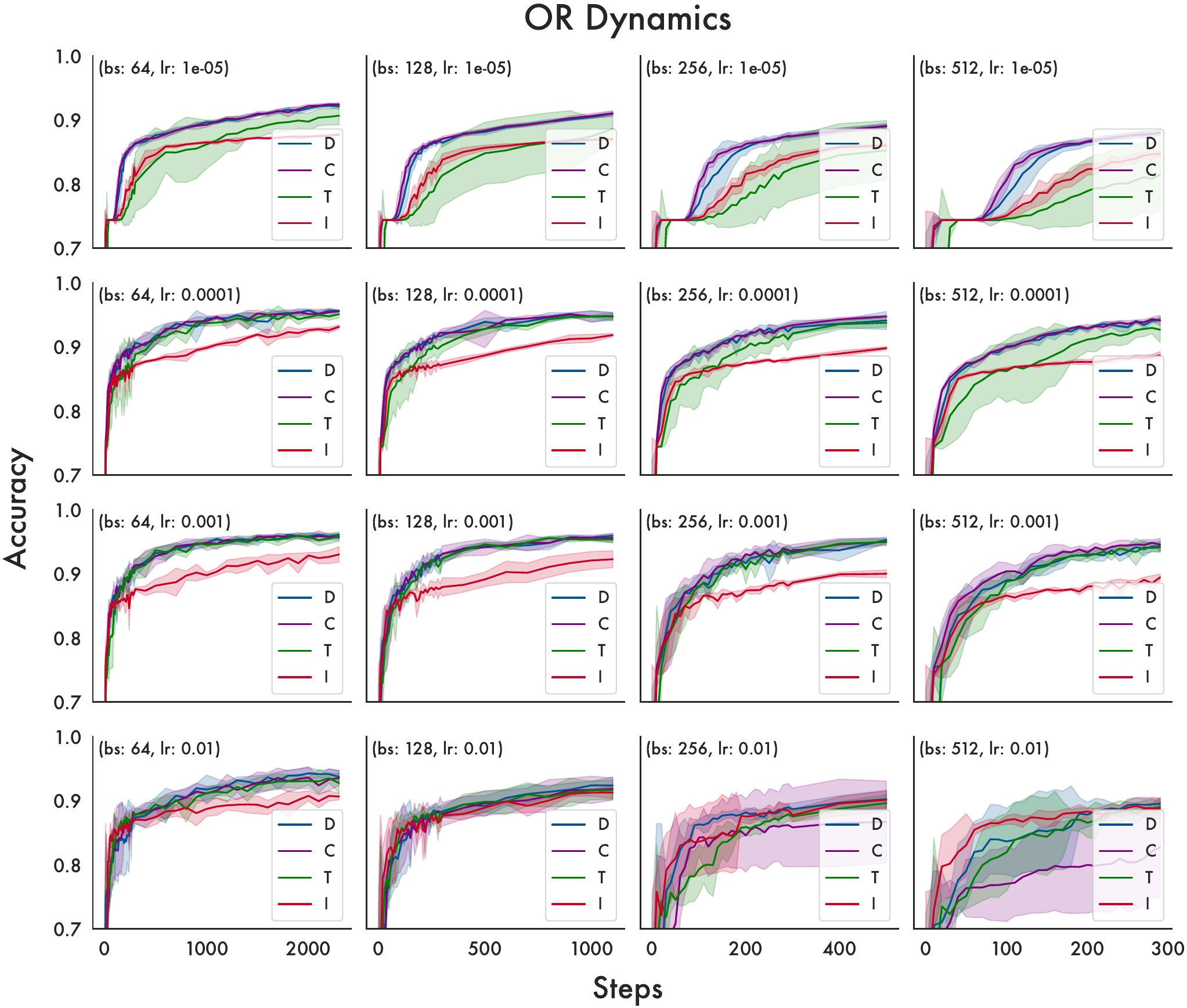}
  \centering
  \vspace{.6em}    
  \caption{Summary of dynamics across configurations}
\end{figure}

\clearpage

\end{document}